%% file: main.tex
\documentclass[10pt,twocolumn,letterpaper]{article}

\usepackage[pagenumbers]{cvpr} 

\input{preamble}
\definecolor{cvprblue}{rgb}{0.21,0.49,0.74}
\usepackage[pagebackref,breaklinks,colorlinks,allcolors=cvprblue]{hyperref}

\usepackage{tcolorbox}
\usepackage{graphicx}
\usepackage{caption}
\usepackage{multirow}
\usepackage[table]{xcolor}

\def\confName{CVPR}
\def\confYear{2027}

\title{Memorization Lives in a Few Attention Heads: Mitigating Memorization in Diffusion Models via Projection-Based Suppression}

\author{%
Puwei Lian$^{1}$\quad
Yujun Cai$^{1}$\quad
Songze Li$^{1,2,\dagger}$\quad
Yuezu Lv$^{3}$\quad
Guanghui Wen$^{1}$\\[0.3em]
$^{1}$Southeast University\\
$^{2}$Engineering Research Center of Blockchain Application,\\
Supervision and Management (Southeast University), Ministry of Education\\[0.3em]
$^{3}$Beijing Institute of Technology\\
{\tt\small 220255555@seu.edu.cn \quad yujuncai@seu.edu.cn \quad songzeli@seu.edu.cn}\\
{\small $^\dagger$Corresponding author}
}

\makeatletter
\renewcommand\paragraph{\@startsection{paragraph}{4}{\z@}%
  {0.9ex \@plus 0.2ex \@minus 0.1ex}%
  {-1em}%
  {\normalfont\normalsize\bfseries}}
\makeatother

\begin{document}
\maketitle
\input{sec/0_abstract}    
\input{sec/1_intro}
\input{sec/2_related}

\input{sec/3_pre}
\input{sec/4_method}

\input{sec/5_exper}
\input{sec/6_con}

{
    \small
    \bibliographystyle{ieeenat_fullname}
    \bibliography{main}
}

\input{sec/X_suppl}

\end{document}

%% file: sec/0_abstract.tex
\begin{abstract}
Diffusion models can memorize and reproduce training samples during inference, raising serious privacy and copyright concerns. Existing inference-time mitigation methods often suppress conditional information globally, which may reduce memorization but can also harm semantic alignment and image quality. In this paper, we revisit memorization from the perspective of cross-attention heads. Through systematic analysis and targeted ablation, we find that memorization is mainly driven by a small subset of shallow-layer attention heads with abnormally strong semantic control, rather than by the entire cross-attention mechanism. Motivated by this finding, we propose \textbf{Projection-based Top-$k$ Attention Head Suppression}, a targeted inference-time mitigation method. Our method first identifies the top-k attention heads most related to memorization using a semantic-control score, and then removes their condition-specific components through directional projection between conditional and unconditional representations. This design selectively weakens memorization-related features while preserving useful semantic information. Extensive experiments on different diffusion models and datasets show that our method effectively reduces memorization and achieves a better trade-off between memorization mitigation and image quality than existing approaches.
\end{abstract}

%% file: sec/1_intro.tex
\section{Introduction}
\label{sec:intro}

Diffusion models have demonstrated remarkable generative capabilities in image and video synthesis \cite{ho2020denoising, zhang2023adding, liu2023audioldm, yang2025cogvideox}, producing highly realistic and high-quality visual content. However, these models also raise serious privacy, copyright, and security concerns \cite{carlini2023extracting, somepalli2023diffusion, lian2025unveiling, zeng2025advi2i, huang2026beauty}. Recent studies have shown that diffusion models may memorize and reproduce training samples nearly verbatim during inference \cite{carlini2023extracting, somepalli2023diffusion, wen2024detecting}. Such memorization may leak sensitive or copyrighted data, while also degrading the model's generalization and generation diversity.

To mitigate memorization, existing studies have explored various inference-time strategies with relatively low computational overhead. These methods reduce memorization by perturbing input prompts or optimizing text embeddings \cite{somepalli2023understanding, wen2024detecting}, modifying the classifier-free guidance process \cite{ho2022classifier, jain2025classifier, chen2025enhancing}, adjusting the initial noise or latent representations \cite{han2025adjusting, jeon2025understanding}, and manipulating token-level cross-attention distributions \cite{ren2024unveiling}. However, these interventions lack fine-grained targeting and may inadvertently suppress useful semantic information, leading to semantic drift or image quality degradation. NeMo \cite{hintersdorf2024finding} takes a model-localization perspective and identifies prompt-specific memorization neurons in cross-attention value projections. However, it requires prompt-specific multi-seed probing, introducing additional overhead, while providing limited insight into how memorization-related signals are structurally organized within the attention mechanism.



\begin{figure}[t]
  \centering
  \includegraphics[width=\linewidth]{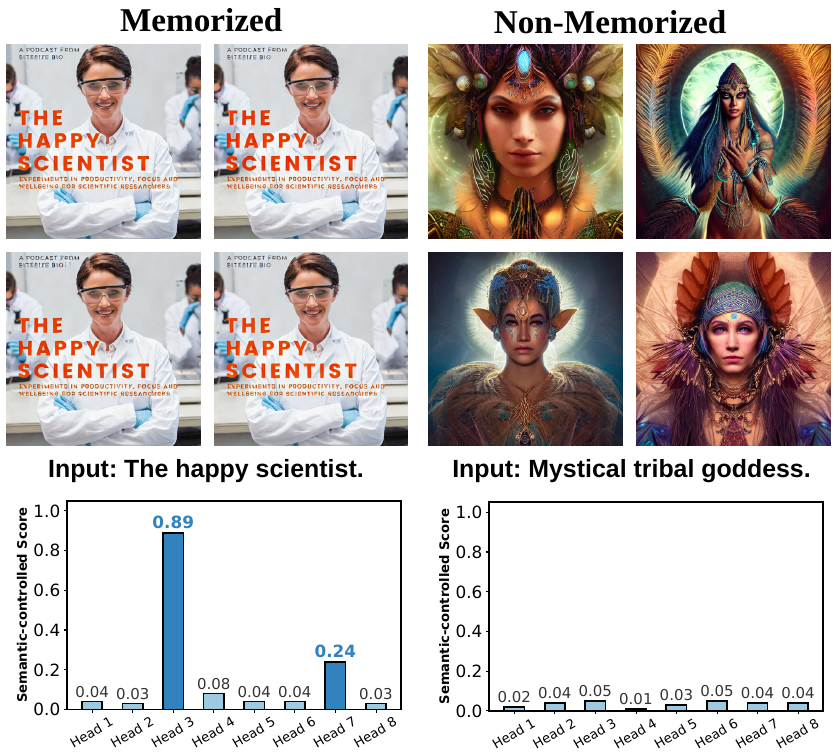}

  \caption{Examples of semantic-control scores across attention heads. Memorized images tend to activate a few distinctly prominent attention heads, whose semantic response strengths are substantially higher than those of other attention heads. In contrast, non-memorized images exhibit a much smoother and more uniform distribution of semantic strength across attention heads.}
  \label{Attentoin_show}
  \vspace{-10pt}
\end{figure}

Conditional generation in diffusion models relies on the coordinated behavior of attention heads across layers. This motivates us to investigate how memorization-related conditional influence is distributed among individual attention heads. By comparing memorized and non-memorized samples, we find that a small subset of shallow-layer attention heads exhibits abnormally strong semantic control in memorized samples, as illustrated in Fig.~\ref{Attentoin_show}. To investigate whether these heads contribute to memorization, we conduct targeted ablation experiments comparing heads with strong semantic control, weak semantic control, and randomly selected heads. Our results show that suppressing heads with strong semantic control significantly reduces memorization, whereas suppressing the other groups has little effect. These findings suggest that memorization is primarily driven by a small subset of shallow-layer attention heads with unusually strong semantic control.

Motivated by this finding, we propose \textbf{Projection-based Top-$k$ Attention Head Suppression}, an inference-time approach for targeted memorization mitigation. Specifically, we use a semantic-control score to identify the Top-$k$ shallow-layer attention heads associated with memorization. For each selected head, we use directional projection to isolate and suppress the memorization-related component of its conditional representation with a controllable strength. This targeted approach effectively reduces training data reproduction while better preserving semantic consistency and image quality, achieving a more favorable trade-off than existing methods. Our main contributions are summarized as follows:


\begin{itemize}[left=3pt]
    \item \textbf{Insight.} We reveal that memorization is associated with a small subset of shallow-layer attention heads that are subject to abnormally strong conditional semantic control, providing a structural understanding of how memorization emerges across attention heads and layers.
    
    \item \textbf{Method.} We propose a projection-based Top-$k$ attention head suppression that identifies shallow-layer attention heads most strongly associated with memorization and selectively suppresses their memorization-related components through directional projection, thereby avoiding the limitations of global conditional suppression.
    
    \item \textbf{Validation.} Experiments across different models and datasets demonstrate that our method consistently and effectively reduces memorization while preserving semantic consistency and image quality.
\end{itemize}

%% file: sec/2_related.tex
\section{Related Work}

\paragraph{Memorization in Diffusion Models.} Diffusion models have been shown to memorize exact training examples \cite{gu2023memorization, hintersdorf2024finding, buchanan2026edge}. \citet{somepalli2023diffusion} demonstrated that, in certain cases, these models can reproduce samples from their training data with near-perfect fidelity. This phenomenon has also been observed in large-scale pretrained models such as Stable Diffusion \cite{rombach2022high}. Moreover, \citet{carlini2023extracting, webster2023reproducible} extracted portions of the training data from Stable Diffusion, along with their corresponding prompts, through data extraction attacks. Based on these findings, prior work \cite{webster2023reproducible, ren2024unveiling} categorizes memorization in text-to-image diffusion models into two types: verbatim memorization and template memorization. Verbatim memorization refers to cases where a model generates outputs that are nearly identical to specific training samples when conditioned on the original prompts. Template memorization describes a weaker form of memorization, in which the generated images remain highly similar to training examples while containing minor non-semantic variations.

\paragraph{Inference-time Memorization Mitigation.} Inference-time mitigation of memorization in diffusion models has attracted increasing attention due to its practicality and efficiency. Several studies \cite{somepalli2023understanding, wen2024detecting} mitigated memorization through prompt optimization, while other methods \cite{chen2024towards, chen2025enhancing, jain2025classifier} addressed this issue by modifying the classifier-free guidance process. Building on classifier-free guidance modification, \citet{han2025adjusting} further improved mitigation performance by optimizing semantic information in the initial noise. In addition, \citet{jeon2025understanding} mitigated memorization by optimizing latent representations based on the sharpness of the log-probability density. Meanwhile, \citet{ren2024unveiling} reduced memorization by adjusting the model’s attention to the beginning token. Moreover, NeMo \cite{hintersdorf2024finding} identifies and prunes prompt-specific memorization neurons in the value projections of cross-attention layers. However, its neuron-level analysis provides limited insight into how memorization-related signals are organized within the attention mechanism. In contrast, we investigate memorization from an attention-head perspective and find that it is primarily associated with a small subset of attention heads. Based on this observation, our method dynamically identifies and selectively suppresses these heads during denoising, reducing memorized generation while better preserving semantic consistency and image quality.

%% file: sec/3_pre.tex
\begin{figure*}[h]
  \centering
  \includegraphics[width=\linewidth]{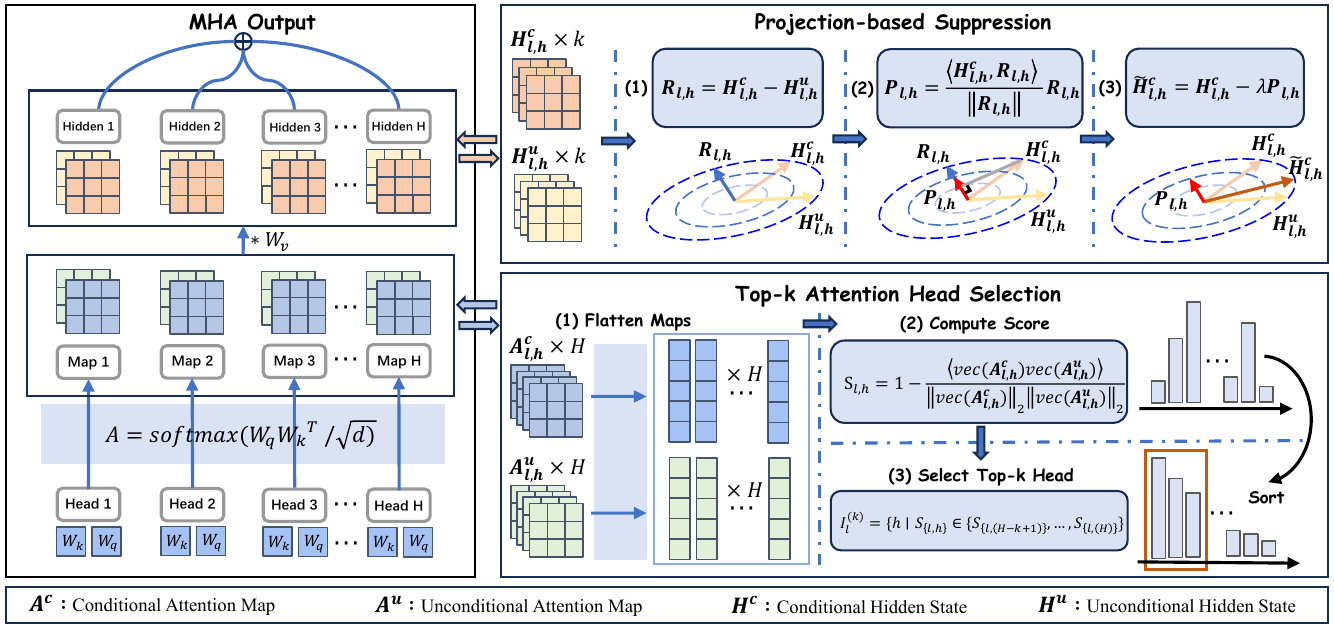}
  \caption{Overview of our framework. The \textbf{left panel} illustrates the computation of multi-head attention. Based on this process, we introduce two modifications on the \textbf{right panel}: \textbf{(1)} Identifying attention heads with strong semantic control through attention map analysis. \textbf{(2)} Applying projection-based suppression to mitigate memorization-related components in the selected heads.}
  \label{Overview}
  \vspace{-10pt}
\end{figure*}

\section{Preliminaries}

\paragraph{Cross-Attention in Diffusion Models.}
In text-to-image diffusion models, text-image correspondence is established through the cross-attention mechanism \cite{vaswani2017attention, hao2017end}, which enables text-guided generation \cite{rombach2022high}. Given a text prompt $\mathbf{y} = ({y_1, y_2, \cdots, y_n})$, a pretrained CLIP text encoder \cite{radford2021learning} first embeds it into a sequence representation, yielding the conditioning vector $c = f_{\text{CLIP}}(\mathbf{y})$. Linear projections are then used to construct the key representation $W_k$ and value representation $W_v$ from $c$, while the query representation $W_q$ is derived from the intermediate features of the denoising network. The attention map $A$ is computed as follows:
\begin{equation}
A = \mathrm{Softmax}\left(\frac{W_q W_k^T}{\sqrt{d_k}}\right),
\end{equation}
where $d_k$ denotes the dimensionality of the key feature space. The output of the cross-attention is given by $AW_v$. The denoising process in diffusion models typically involves both unconditional and conditional branches through classifier-free guidance. We denote by $A^u$ the attention map from the unconditional branch and by $A^c$ the attention map from the conditional branch.

\paragraph{Multi-Head Attention (MHA).}
MHA \cite{vaswani2017attention} enables the model to capture diverse representational patterns by projecting the input into multiple subspaces. Prior studies have shown that different attention heads specialize in capturing distinct features and contribute unequally to the behavior of the model \cite{clark2019does, voita2019analyzing}. At the $l$-th layer, MHA with $n$ heads is formulated as follows:
\begin{equation}
\mathrm{MHA}^{(l)} = \left(h_1^{(l)} \oplus h_2^{(l)} \oplus \cdots \oplus h_n^{(l)}\right) W_o^{(l)},
\end{equation}
\begin{equation}
h_i^{(l)} = \mathrm{Softmax} \left( \frac{W_q^{i,(l)} \left(W_k^{i,(l)}\right)^T}{\sqrt{d_{k}}} \right) W_v^{i,(l)},
\end{equation}
where $h_i^{(l)}$ denotes the output of the $i$-th attention head at the $l$-th layer, and $\oplus$ denotes concatenation. $W_q^{i,(l)}$, $W_k^{i,(l)}$, and $W_v^{i,(l)}$ denote the projected query, key, and value representations of the $i$-th head at layer $l$, respectively. $W_o^{(l)}$ is the output projection matrix, which maps the concatenated head features back to the original feature space at layer $l$. In this work, we mainly focus on the multi-head cross-attention in diffusion models throughout our analysis.

\paragraph{Classifier-Free Guidance (CFG).}
CFG \cite{ho2022classifier} is widely used in text-to-image diffusion models to strengthen prompt alignment during inference. Specifically, the denoising model simultaneously performs an unconditional prediction $\epsilon_{\theta}(x_t)$ and a conditional prediction $\epsilon_{\theta}(x_t, c)$ conditioned on the input text prompt. The final guided prediction is obtained by extrapolating from the unconditional prediction toward the conditional prediction:
\begin{equation}
\hat{\epsilon}_{\theta}(x_t, c)
=
\epsilon_{\theta}(x_t)
+
w \left(
\epsilon_{\theta}(x_t, c)
-
\epsilon_{\theta}(x_t)
\right),
\end{equation}
where $w$ is the guidance scale controlling the strength of conditional guidance. Therefore, during inference, diffusion models naturally contain both unconditional and conditional branches, whose differences reflect the semantic influence introduced by the text condition.

%% file: sec/4_method.tex
\section{Methodology}
In this section, we provide an overview of our method. We first analyze the cross-attention mechanism and show that memorization is mainly driven by a small subset of shallow-layer attention heads with strong semantic control. Based on this observation, we propose projection-based Top-$k$ attention head suppression. Our method identifies these memorization-related heads and selectively removes their condition-specific feature components through directional projection, thereby mitigating memorization while preserving image quality. The framework is illustrated in Fig.~\ref{Overview}.

\subsection{Cross-Attention Analysis of Memorization}
\label{motivation}

Prior studies \cite{wen2024detecting, ren2024unveiling, somepalli2023understanding} have suggested that memorization in diffusion models is closely related to overfitting on conditional information. A natural mitigation strategy is therefore to weaken the conditional signal during generation. However, such interventions often treat conditional influence in a relatively uniform manner, overlooking how it is organized across different functional units of the model. Since conditional generation is jointly governed by attention heads across multiple layers, we ask a more structural question: \emph{how is conditional influence distributed across attention heads and layers during memorized generation?}

To answer this question, we analyze the cross-attention mechanism in the unconditional and conditional branches under classifier-free guidance. Following the observation that memorized samples exhibit a large discrepancy between conditional and unconditional noise predictions \cite{wen2024detecting}, we examine whether such a discrepancy is reflected in the attention distributions. For the $h$-th attention head in layer $l$, let $A_{l,h}^{c}$ and $A_{l,h}^{u}$ denote the cross-attention maps in the conditional and unconditional branches, respectively. Then, we define the \textbf{semantic-control score} as follows:
\begin{equation}
\label{cos}
S_{l,h} = 1 -
\frac{
\left\langle \mathrm{vec}(A_{l,h}^{c}), \mathrm{vec}(A_{l,h}^{u}) \right\rangle
}{
\left\|\mathrm{vec}(A_{l,h}^{c})\right\|_2
\left\|\mathrm{vec}(A_{l,h}^{u})\right\|_2
},
\end{equation}
where $\mathrm{vec}(\cdot)$ denotes vectorization, $\langle \cdot,\cdot \rangle$ denotes the inner product between two vectors, and $\|\cdot\|_2$ denotes the $\ell_2$ norm. A larger $S_{l,h}$ indicates a greater discrepancy between the conditional and unconditional attention maps.

We present the statistical results for layer $3$ of the model in Fig.~\ref{Attentoin_show}. For memorized images, a small number of attention heads exhibit substantially higher semantic-control scores than the others. In contrast, non-memorized images show a much smoother and more uniform distribution of semantic strength across attention heads. These observations suggest that semantic injection is not uniformly distributed across all heads. Instead, a few attention heads are highly sensitive to the condition and may dominate the generation trajectory for memorized prompts.

To further quantify this phenomenon, we analyze the distribution of $S_{l,h}$ across different cross-attention layers using Stable Diffusion v1.4 as a case study. For each layer $l$ with $H$ attention heads, we collect the score set $\{ S_{l,h} \}_{h=1}^{H}$ and sort the scores in ascending order:
\begin{equation}
S_{l,(1)} \leq S_{l,(2)} \leq \cdots \leq S_{l,(H)}.
\end{equation}
The first and third quartiles are defined as:
\begin{equation}
Q_1(\mathcal{S}_l) = S_{l,(\lfloor 0.25H \rfloor)}, 
\;
Q_3(\mathcal{S}_l) = S_{l,(\lfloor 0.75H \rfloor)}.
\end{equation}
We then compute the interquartile range:
\begin{equation}
\mathrm{IQR}_l = Q_3(\mathcal{S}_l) - Q_1(\mathcal{S}_l).
\end{equation}
An attention head is considered to have abnormally strong semantic control if:
\begin{equation}
\label{outlier}
S_{l,h} > Q_3(\mathcal{S}_l) + \eta \cdot \mathrm{IQR}_l,
\end{equation}
where $\eta=0.15$ controls outlier sensitivity.

\vspace{-5pt}
\begin{figure}[h]
  \centering
  \includegraphics[width=\linewidth]{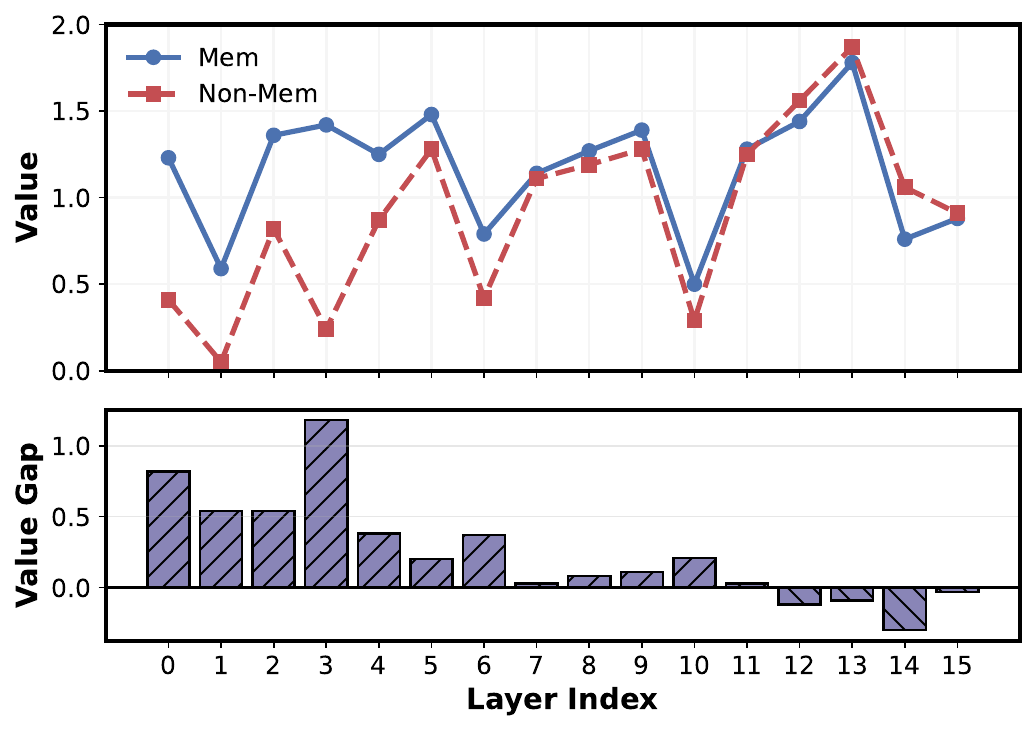}
  \vspace{-15pt}
  \caption{Layer-wise analysis of attention heads with strong semantic control. Memorized images exhibit a larger number of high semantic-control attention heads in the shallow layers, especially layers 0$\sim$3. In deeper layers, the statistics of memorized and non-memorized samples become similar.}
  \label{16_layers}
  \vspace{-5pt}
\end{figure}

We present the layer-wise statistics of strong semantic control heads at timestep $t=T$ in Fig.~\ref{16_layers}. Memorized samples contain significantly more heads with strong semantic control in the shallow layers than non-memorized samples. In contrast, non-memorized samples exhibit more uniform score distributions, with few or no abnormal heads. Notably, this discrepancy is primarily concentrated in the shallow layers, while in deeper layers, the two types of samples show largely similar patterns with no significant differences. \textit{We provide additional timestep-wise analysis in the Supplementary Material~\ref{timestep_stability}, showing that this phenomenon persists across almost all timesteps.}

\begin{tcolorbox}[rounded corners, colback=white, colframe=black]
\textbf{Takeaway.} Memorized samples exhibit a distinctive pattern in shallow layers, where a small subset of attention heads is subject to abnormally strong semantic control.
\end{tcolorbox}

\subsection{Attention Head Ablation}
\label{head}
The above analysis identifies attention heads with strong semantic control. We next investigate whether these heads directly affect memorization through targeted ablation experiments. We focus on the first four cross-attention layers, where strong semantic-control heads are frequently observed in memorized images and differ substantially from those in non-memorized images, as discussed in Sec.~\ref{motivation}. For each layer, we compare three groups: \textbf{(1) the Top-3 heads} with the highest semantic-control scores at $t=T$; \textbf{(2) the Bottom-3 heads} with the lowest scores at $t=T$; and \textbf{(3) three randomly selected heads}. We use the SSCD Score \cite{pizzi2022self} to measure memorization, where lower scores indicate better mitigation. If memorization is driven by high semantic-control heads, suppressing the Top-3 heads should be more effective than suppressing the other groups. Formally, for a selected head set $\mathcal{G}_{l}$, we scale the corresponding head outputs by a factor $\epsilon \in [0,1]$:
\begin{equation}
\tilde{h}_{l,i}
=
\begin{cases}
\epsilon h_{l,i}, & i \in \mathcal{G}_{l}, \\
h_{l,i}, & i \notin \mathcal{G}_{l}.
\end{cases}
\end{equation}
When $\epsilon=0$, the selected heads are fully ablated. The modified multi-head attention output is then:
\begin{equation}
\widetilde{\mathrm{MHA}}_{l}
=
\left(
\tilde{h}_{l,1} \oplus \tilde{h}_{l,2} \oplus \cdots \oplus \tilde{h}_{l,H}
\right) W_{l}^{O}.
\end{equation}

\begin{figure}[t]
  \centering
  \includegraphics[width=\linewidth]{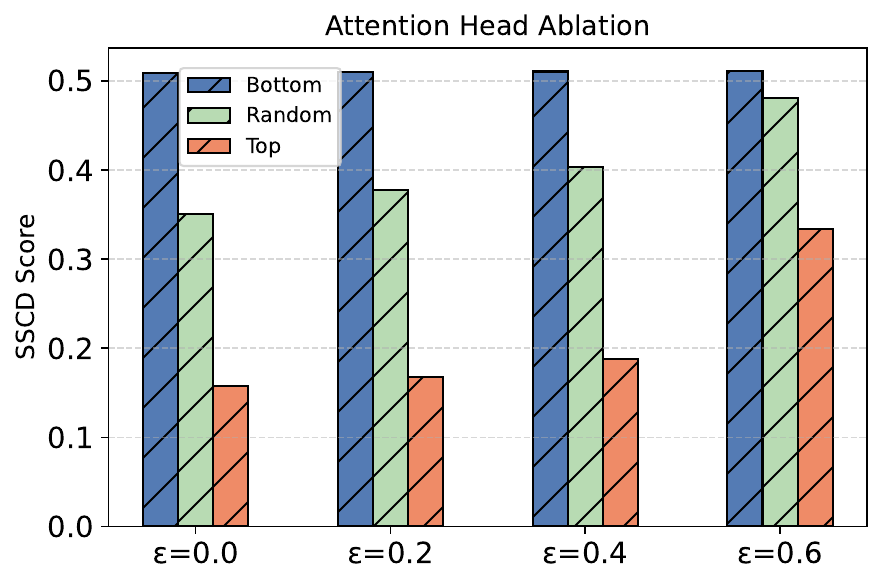}
  \vspace{-5pt}
  \caption{Ablation study on attention head groups. We evaluate different suppression factors $\epsilon \in \{0.0, 0.2, 0.4, 0.6\}$. Suppressing the \textbf{Top-3 heads} produces the largest reduction in SSCD Score, while suppressing the \textbf{Bottom-3 heads} has little effect.}
  \label{Ablation_head}
  \vspace{-10pt}
\end{figure}

The results in Fig.~\ref{Ablation_head} show that suppressing the Top-3 heads substantially reduces SSCD across different suppression strengths $\epsilon$. As $\epsilon$ decreases, the SSCD of the Top-3 group consistently declines, suggesting that weakening these heads effectively suppresses memorization. Random head suppression also yields a moderate reduction in SSCD, as randomly selected heads may occasionally include high semantic-control heads but fail to consistently target memorization-related heads. In contrast, suppressing the Bottom-3 heads produces almost no change in SSCD, even under stronger suppression.

\begin{tcolorbox}[rounded corners, colback=white, colframe=black]
\textbf{Takeaway.} 
For memorized samples, high semantic-control attention heads in shallow layers have a direct causal relationship with memorization. Suppressing these heads significantly weakens training data reproduction, indicating that they are the primary drivers of memorization.
\end{tcolorbox}

\subsection{Top-$k$ Attention Head Suppression}
\label{sec:projection_suppression}
The ablation study confirms that high semantic-control attention heads are closely associated with memorization behavior. However, directly scaling or zeroing out these heads is a coarse intervention, as it may disrupt the original functionality of the attention heads and consequently affect image quality. To achieve more fine-grained suppression, we propose Projection-based Top-$k$ Attention Head Suppression, which selectively removes the condition-specific component from the selected heads. For each pre-selected shallow layer $l$, we first compute the semantic-control score $S_{l,h}$ for each attention head $h$ at the initial denoising timestep $t=T$ according to Eq.~(\ref{cos}). We then select the Top-$k$ attention heads with the highest scores:
\begin{equation}
\label{top-k}
\mathcal{I}_{l}^{(k)}
=
\{\, h \mid S_{l,h} \in \{S_{l,(H-k+1)}, \dots, S_{l,(H)}\} \,\}.
\end{equation}
For each selected head $h \in \mathcal{I}_{l}^{(k)}$, let $\mathbf{H}_{l,h}^{c}$ and $\mathbf{H}_{l,h}^{u}$ denote the hidden activations in the conditional and unconditional branches, respectively. We define the conditional discrepancy direction as:
\(
\mathbf{R}_{l,h}
=
\mathbf{H}_{l,h}^{c}
-
\mathbf{H}_{l,h}^{u}.
\)
This direction captures the component of the head activation that is specifically induced by the text condition. Inspired by model editing techniques \cite{belrose2023leace, arditi2024refusal}, we compute the projection of the conditional activation onto this discrepancy direction:
\begin{equation}
\mathbf{P}_{l,h}
=
\frac{
\left\langle \mathbf{H}_{l,h}^{c}, \mathbf{R}_{l,h} \right\rangle
}{
\left\|\mathbf{R}_{l,h}\right\|^2
}
\mathbf{R}_{l,h},
\end{equation}
where $\mathbf{P}_{l,h}$ denotes the component of $\mathbf{H}_{l,h}^{c}$ aligned with the conditional discrepancy direction. Based on this projection, we suppress the memorization-related component by subtracting it from the original conditional activation:
\begin{equation}
\tilde{\mathbf{H}}_{l,h}^{c}
=
\mathbf{H}_{l,h}^{c}
-
\lambda \mathbf{P}_{l,h},
\end{equation}
where $\lambda$ controls the suppression strength. Compared with direct head ablation, this projection-based strategy is milder and more selective: it suppresses memorization-related conditional features while retaining the remaining information in the selected attention heads.

Overall, our method follows a targeted mitigation principle. First, we identify the attention heads most responsible for abnormal semantic control. Then, rather than globally weakening the prompt condition or disabling entire heads, we selectively remove only the condition-specific direction associated with memorization.

%% file: sec/5_exper.tex
\section{Experiments}

\begin{figure*}[ht]
  \centering
  \begin{subfigure}{0.485\textwidth}
    \centering
    \includegraphics[width=\linewidth]{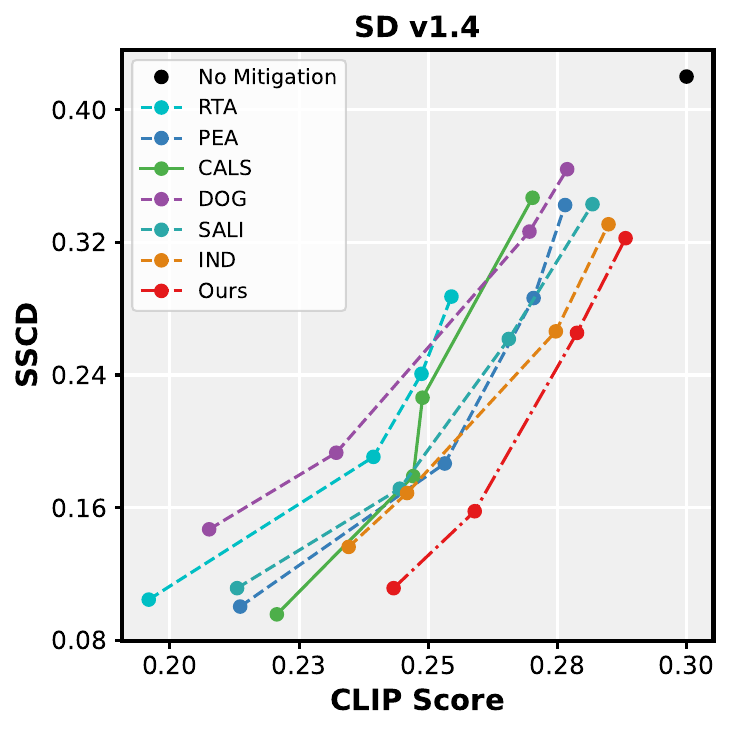}
  \end{subfigure}
  \hspace{1mm}
  \begin{subfigure}{0.485\textwidth}
    \centering
    \includegraphics[width=\linewidth]{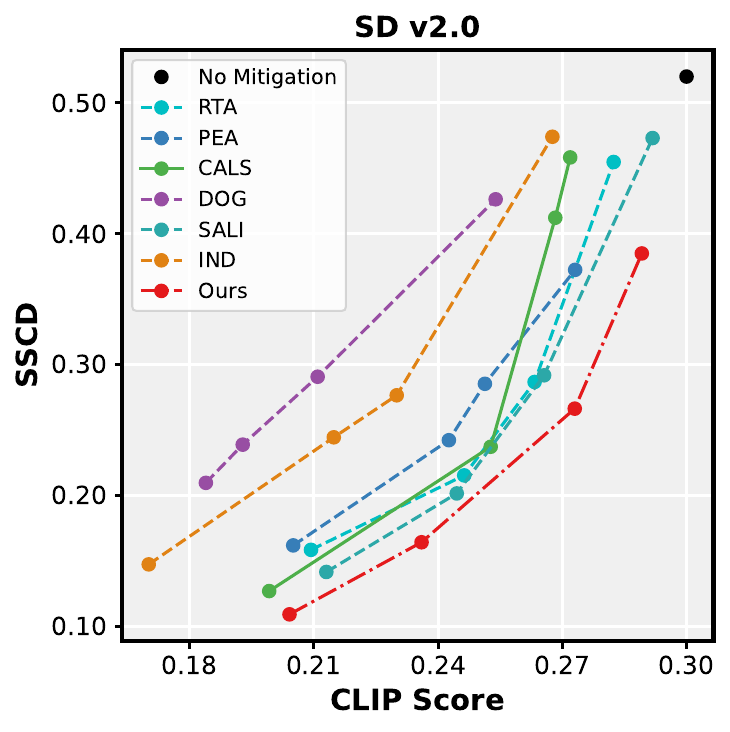}
  \end{subfigure}
  \caption{SSCD-CLIP trade-off curve comparison with baselines. Our method consistently lies in the lower-right region compared with all baselines. This demonstrates that, under the same SSCD level, our approach achieves higher CLIP scores. In other words, given comparable semantic alignment, our method provides the most effective suppression of memorization.}
  \label{main_result}
  \vspace{-5pt}
\end{figure*}

\subsection{Experimental Setup}
\label{setup}

\paragraph{Models and Datasets.} Following prior work on mitigating memorization in diffusion models \cite{ren2024unveiling, jain2025classifier, han2025adjusting, jeon2025understanding}, we evaluate our method on Stable Diffusion v1.4 (SD v1.4) \cite{rombach2022high} and Stable Diffusion v2.0 (SD v2.0) \cite{rombach2022high} in the pretrained setting. For SD v1.4, we use 500 memorized prompts extracted from the LAION dataset \cite{schuhmann2022laion} by \citet{webster2023reproducible}, and use their corresponding training images as memorized images for evaluation. For SD v2.0, we use 219 memorized prompts released by \citet{ren2024unveiling, jeon2025understanding}, together with their corresponding memorized images. In addition, following the evaluation setting of prior work \cite{jain2025classifier}, we further examine memorization under fine-tuning scenarios. Specifically, we fine-tune SD v1.4 on 2,500 image-text pairs from MS-COCO \cite{lin2014microsoft}, and use the corresponding text prompts as memorized prompts and the training images as memorized images.

\paragraph{Evaluation Metrics.} We employ four metrics to evaluate memorization, image-text alignment, image quality, and image diversity. Specifically, we measure memorization using the Self-Supervised Copy Detection (SSCD) score \cite{pizzi2022self}, which quantifies the similarity between a generated image and its nearest training image. We assess image-text alignment using the CLIP score \cite{radford2021learning}, which measures the semantic consistency between generated images and their corresponding text prompts. To further evaluate generation performance, we use Fréchet Inception Distance (FID) \cite{heusel2017gans} to measure image quality and distribution fidelity, and Learned Perceptual Image Patch Similarity (LPIPS) \cite{zhang2018unreasonable} to assess image diversity. Lower SSCD scores indicate more effective memorization mitigation, whereas higher CLIP and LPIPS scores, together with lower FID values, indicate better overall generation performance.

\paragraph{Baselines.} To evaluate performance, we compare our method against six baselines. All baselines operate at inference time and do not require access to the training data. They include random token addition (RTA) \cite{somepalli2023understanding}, prompt embedding adjustment (PEA) \cite{wen2024detecting}, cross-attention logit scaling (CALS) \cite{ren2024unveiling}, dynamic opposite guidance (DOG) \cite{jain2025classifier}, initial noise adjustment (IND) \cite{han2025adjusting}, and sharpness-aware latent initialization (SALI) \cite{jeon2025understanding}.

\paragraph{Implementation Details.} For evaluation, we generate 8 images per memorized prompt under consistent inference settings across all methods. We use 50 inference steps and a CFG scale of 7.5 for both models, with the DDIM scheduler \cite{song2020denoising} for SD v1.4 and the Euler discrete scheduler \cite{karras2022elucidating} for SD v2.0. For our method, we set $k=3$ for SD v1.4 and $k=4$ for SD v2.0, with $\lambda=1$ by default. All experiments are conducted on a single NVIDIA A100 GPU (80G).

\subsection{Overall Performance}
\label{main}
Following prior studies \cite{chen2024exploring, jain2025classifier, jeon2025understanding, han2025adjusting}, we assess different methods by analyzing the trade-off between SSCD and CLIP scores under varying mitigation strengths. Experiments are conducted on both SD v1.4 and SD v2.0. For each method, we evaluate four hyperparameter settings to ensure a comprehensive comparison and plot the corresponding SSCD-CLIP trade-off curve. In these curves, trajectories closer to the lower-right corner indicate a more favorable balance, with stronger memorization mitigation and better preservation of semantic fidelity in generated images. As shown in Fig.~\ref{main_result}, our method, represented by the red curve, consistently lies closer to the lower-right region than all baselines across both model versions. This result shows that our method achieves higher CLIP scores at the same SSCD level. Equivalently, at a comparable level of semantic alignment, our method achieves lower SSCD scores, indicating more effective memorization suppression than all compared approaches. \textit{We provide additional experiments on SD3 in the Supplementary Material~\ref{SD3}. Further experimental results with more evaluation metrics are presented in the FID and LPIPS section of the Supplementary Material~\ref{FID-LPIPS}.}

\begin{figure*}[ht]
    \centering
    \begin{subfigure}{0.12\textwidth}
        \centering
        \includegraphics[height=1.85cm]{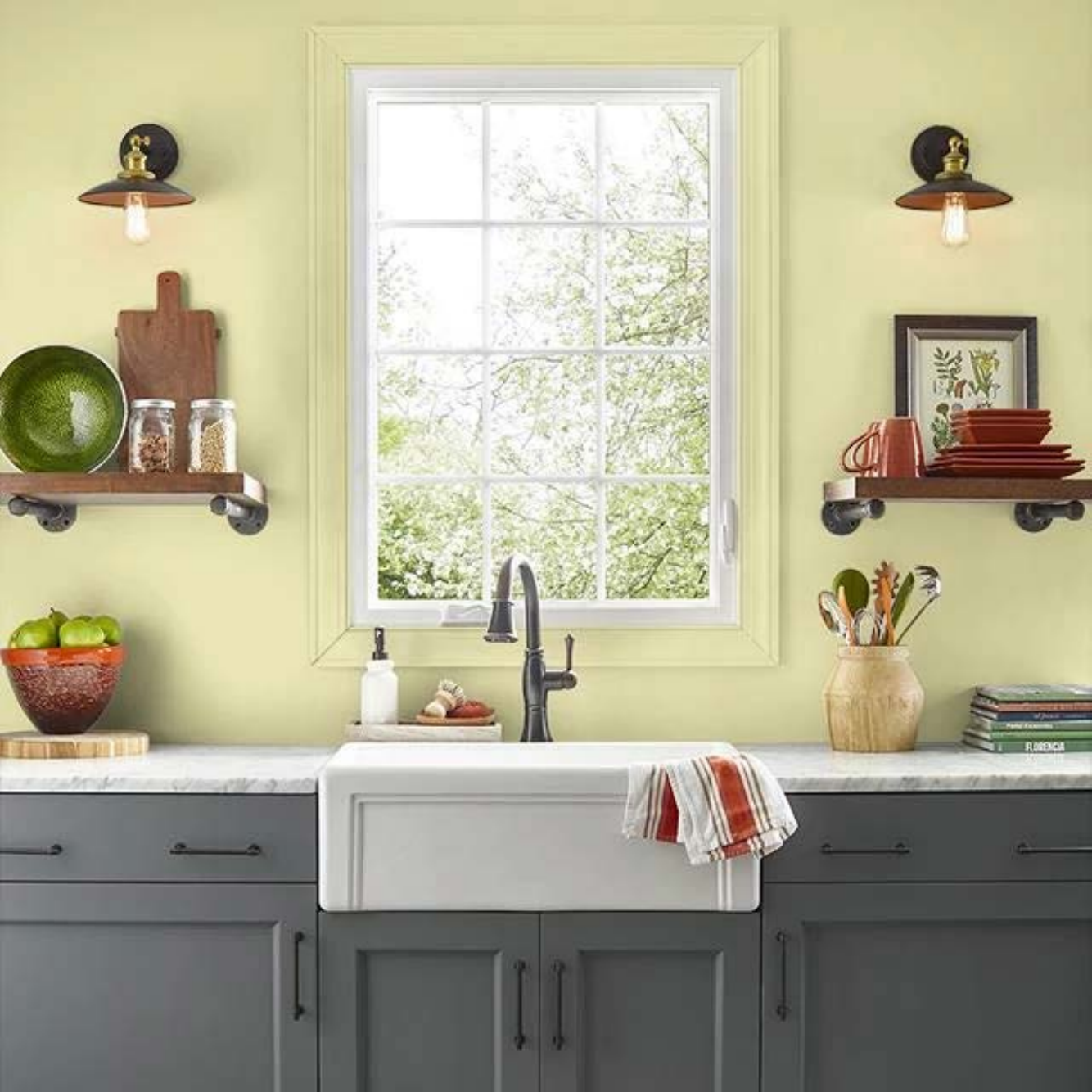}
    \end{subfigure}
    \hfill
    \begin{subfigure}{0.12\textwidth}
        \centering
        \includegraphics[height=1.85cm]{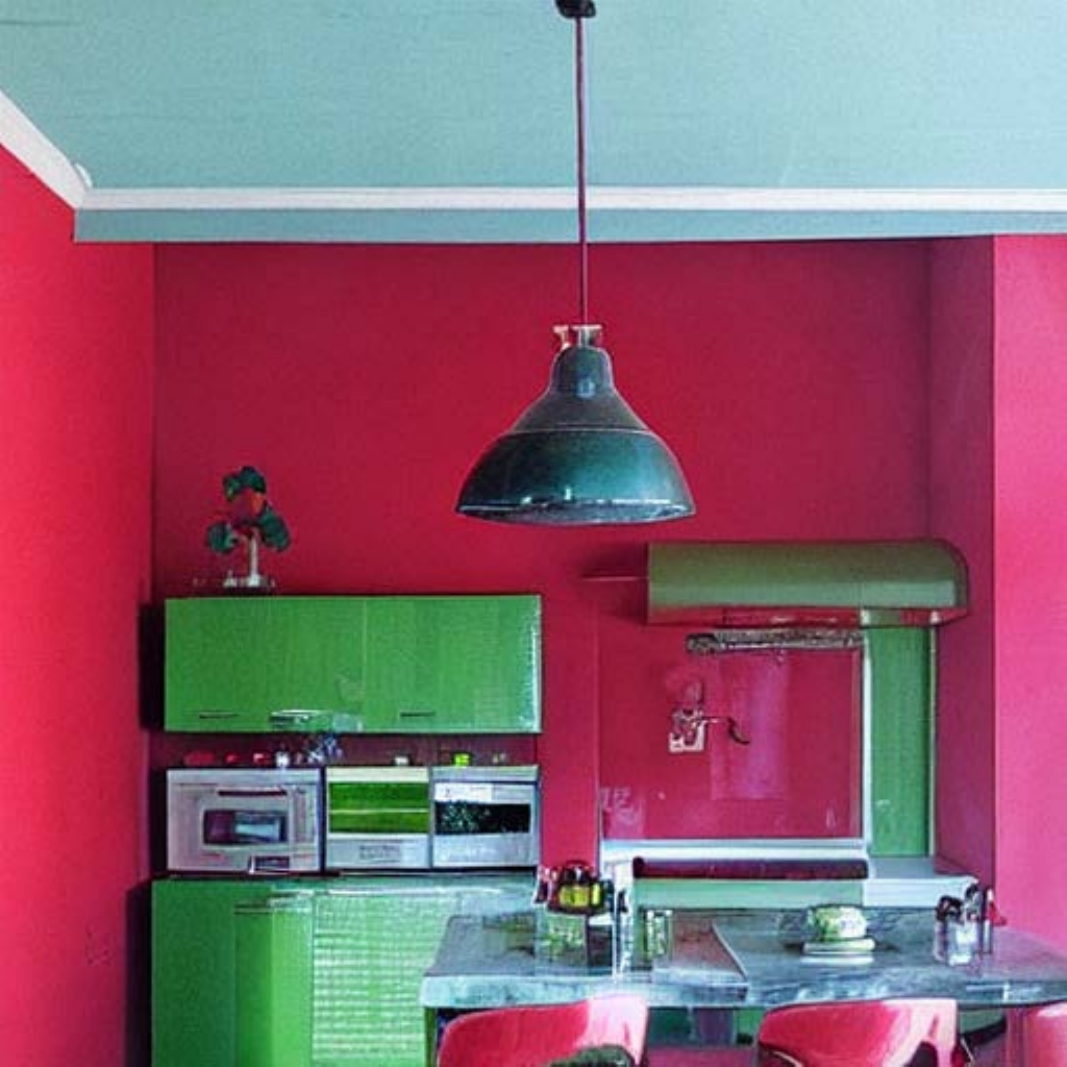}
    \end{subfigure}
    \hfill
    \begin{subfigure}{0.12\textwidth}
        \centering
        \includegraphics[height=1.85cm]{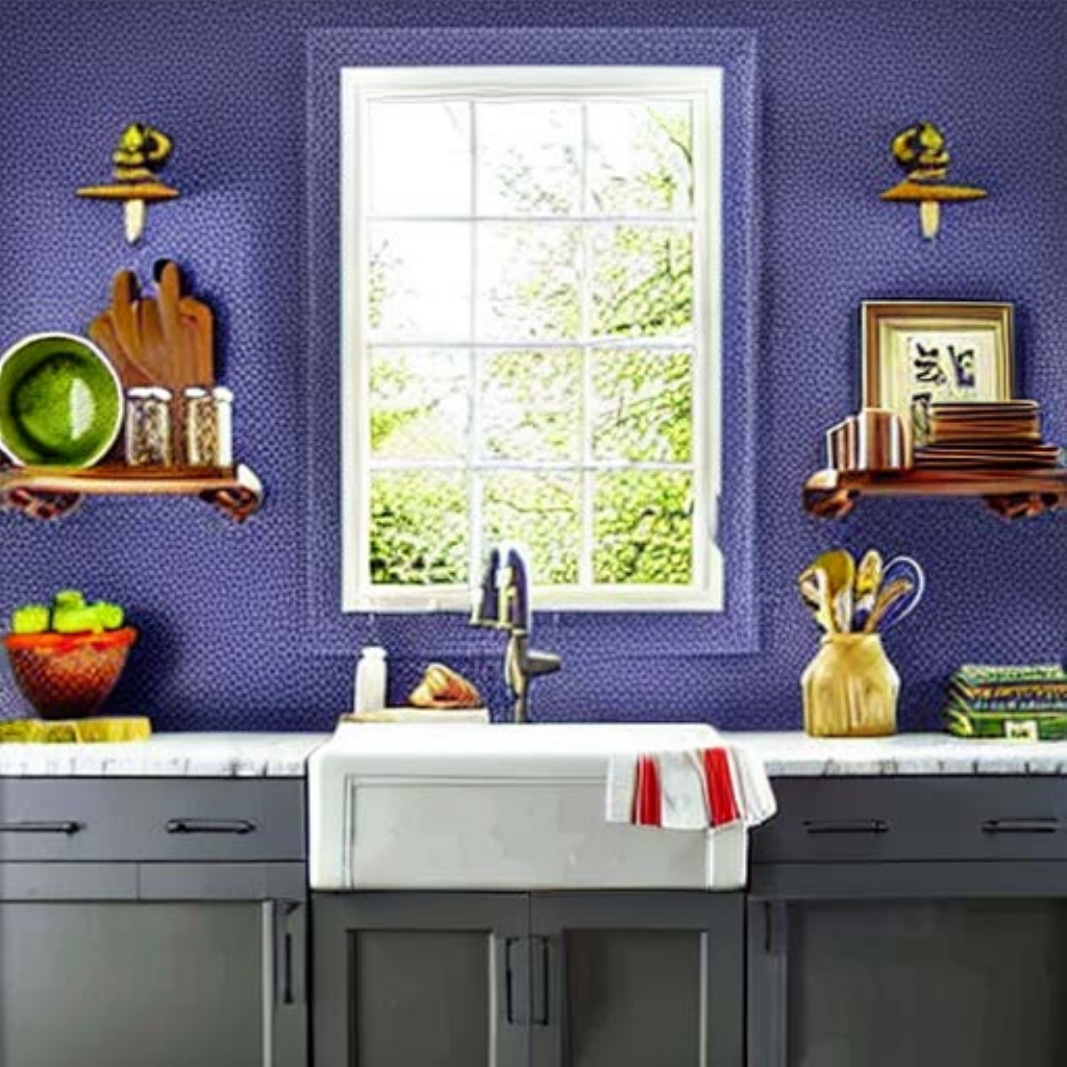}
    \end{subfigure}
    \hfill
    \begin{subfigure}{0.12\textwidth}
        \centering
        \includegraphics[height=1.85cm]{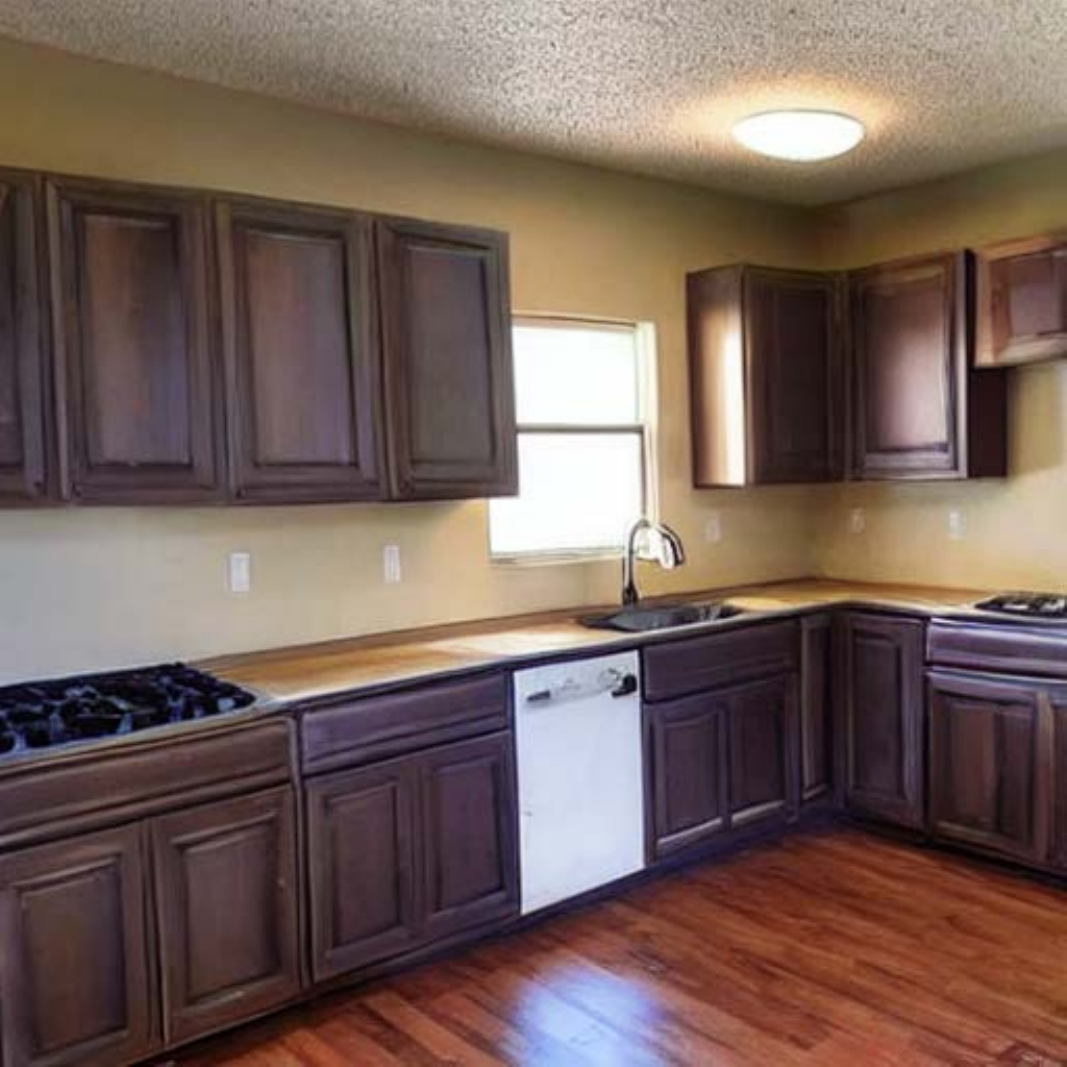}
    \end{subfigure}
    \hfill
    \begin{subfigure}{0.12\textwidth}
        \centering
        \includegraphics[height=1.85cm]{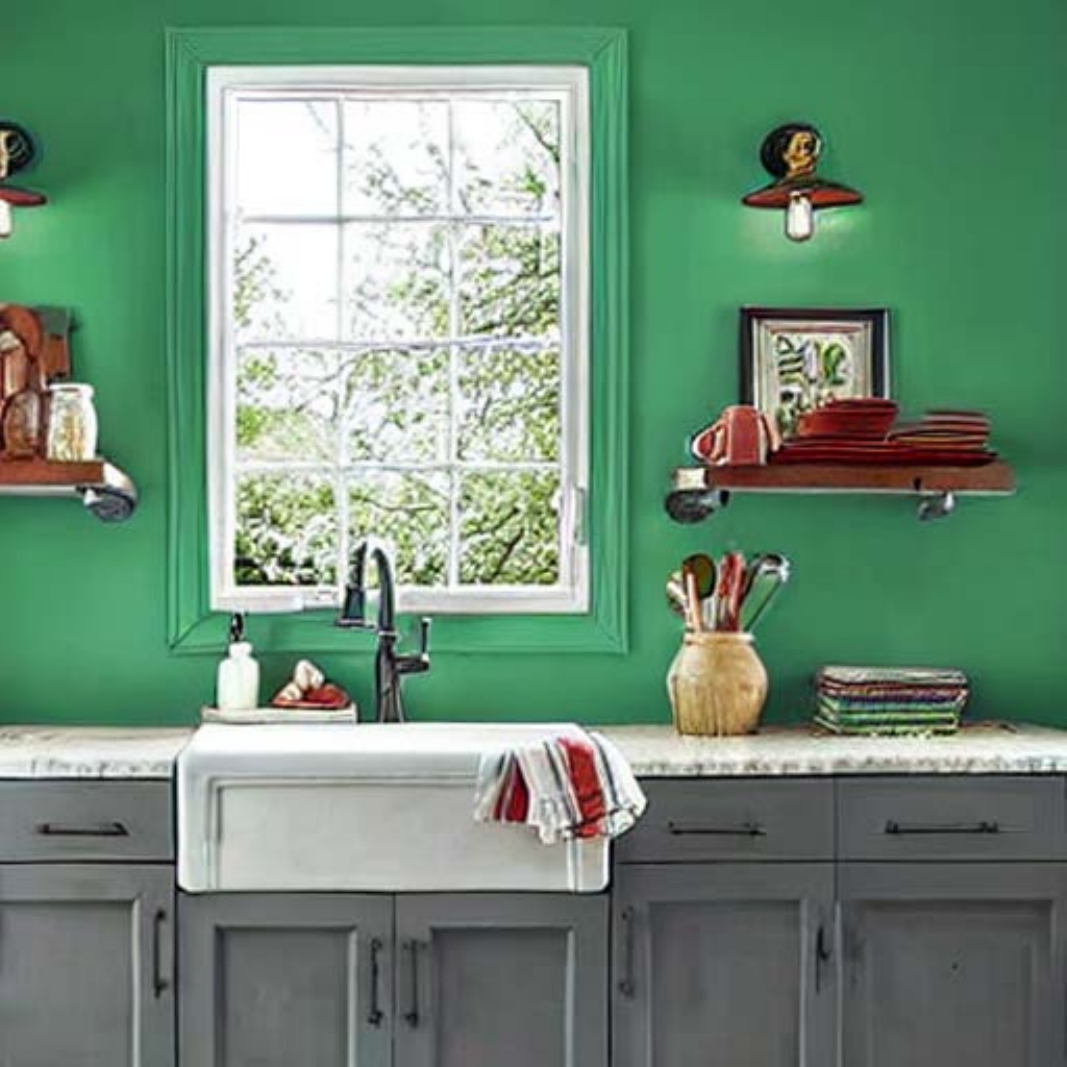}
    \end{subfigure}
    \hfill
    \begin{subfigure}{0.12\textwidth}
        \centering
        \includegraphics[height=1.85cm]{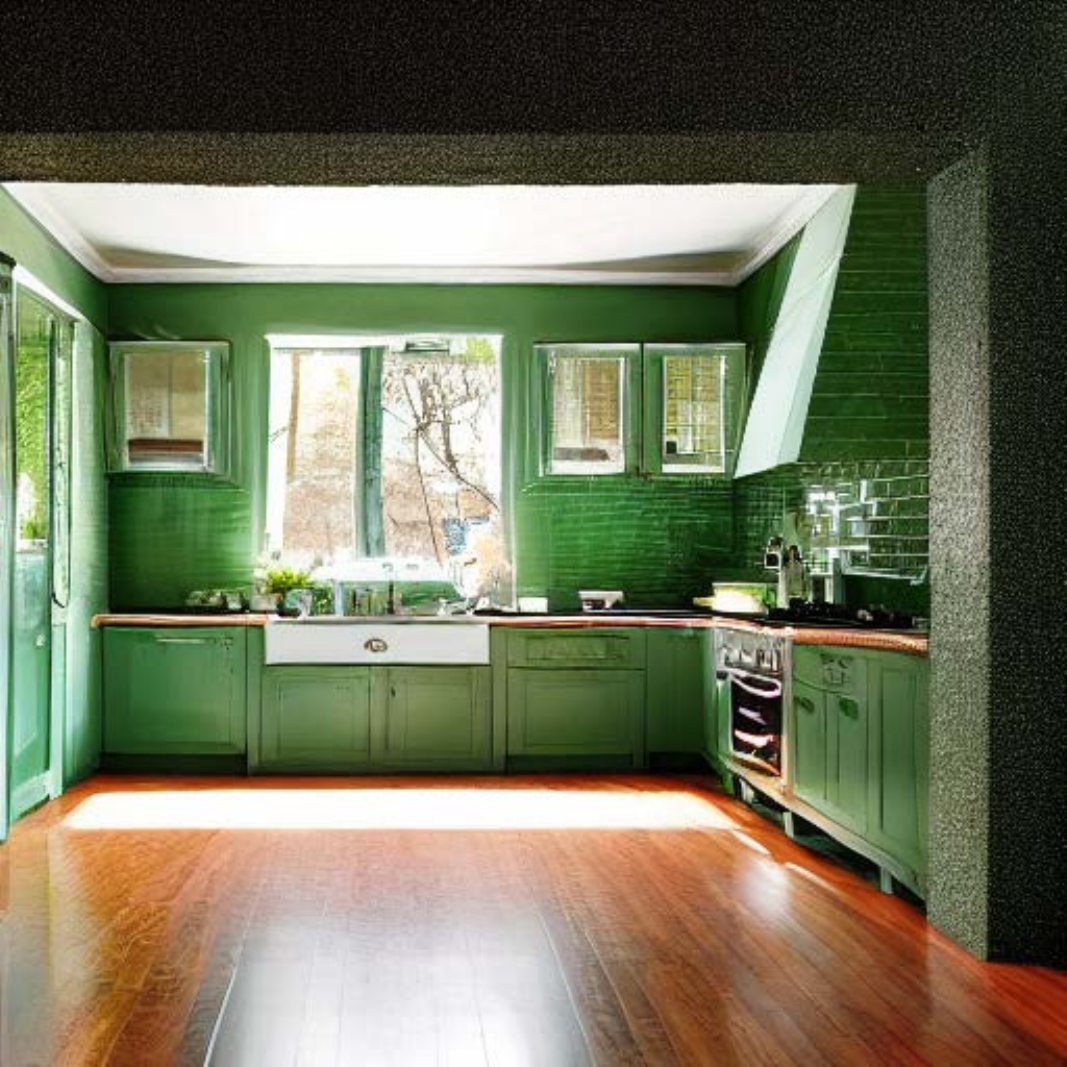}
    \end{subfigure}
    \hfill
    \begin{subfigure}{0.12\textwidth}
        \centering
        \includegraphics[height=1.85cm]{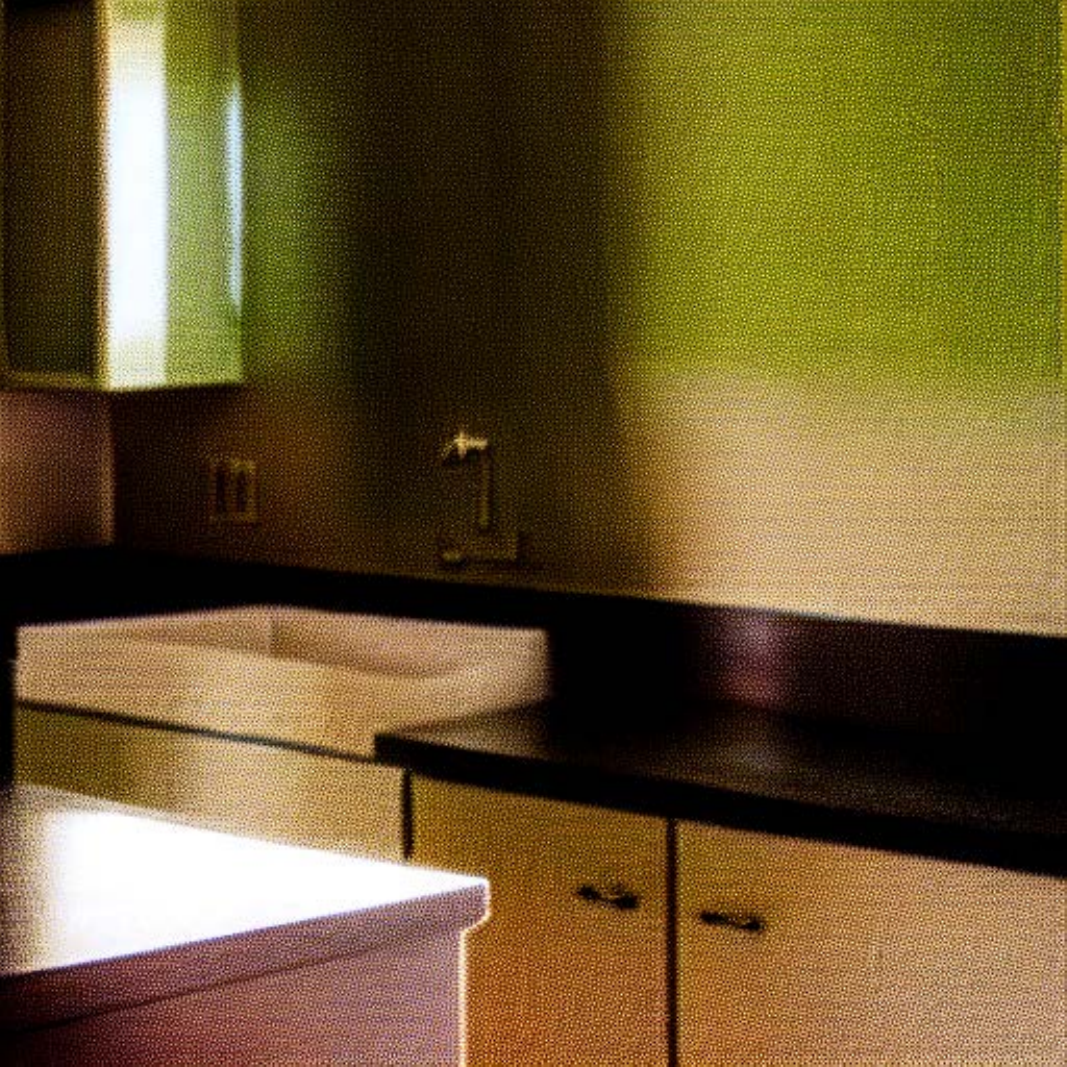}
    \end{subfigure}
    \hfill
    \begin{subfigure}{0.12\textwidth}
        \centering
        \includegraphics[height=1.85cm]{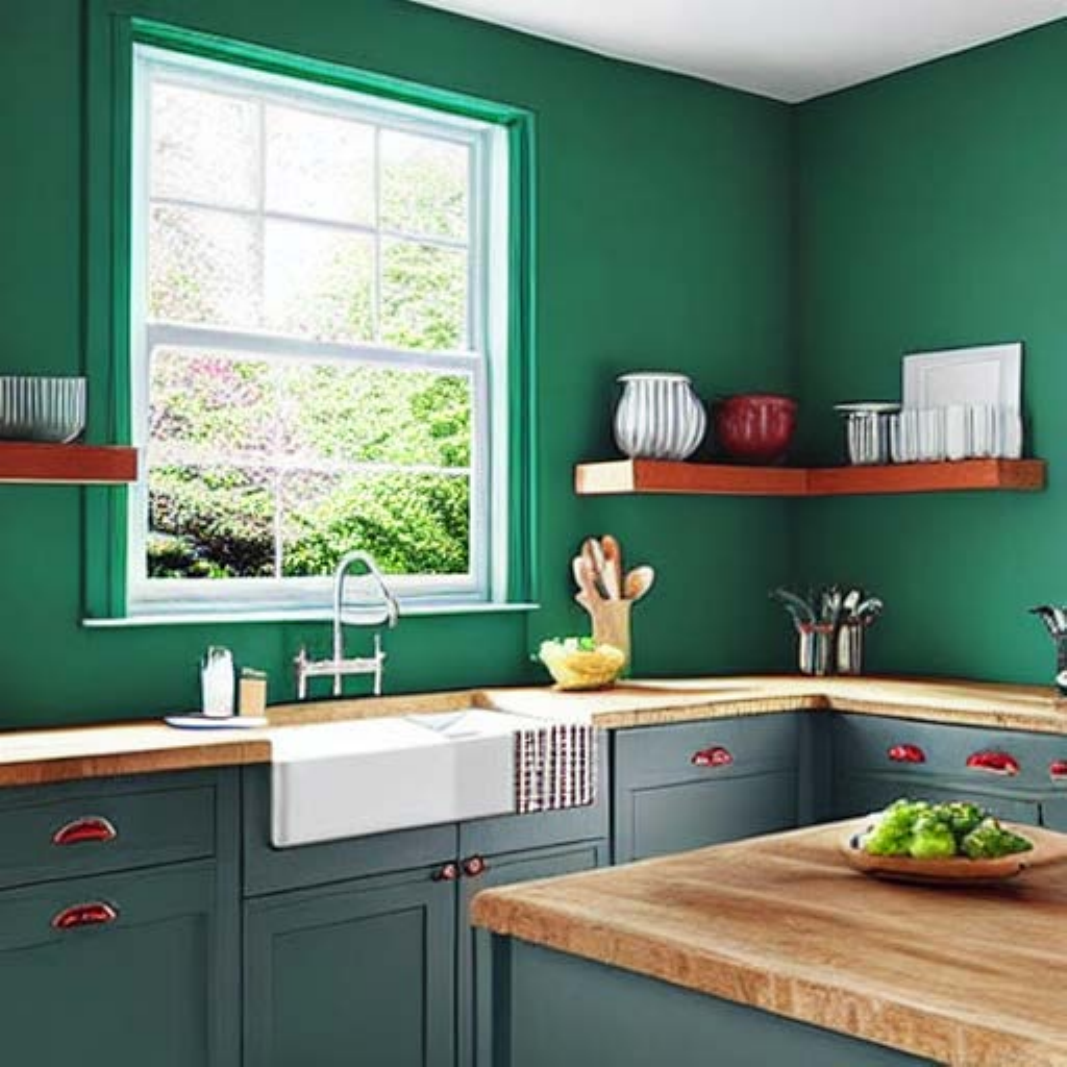}
    \end{subfigure}

    \vspace{0.3em} 
    \begin{subfigure}{0.12\textwidth}
        \centering
        \includegraphics[height=1.85cm]{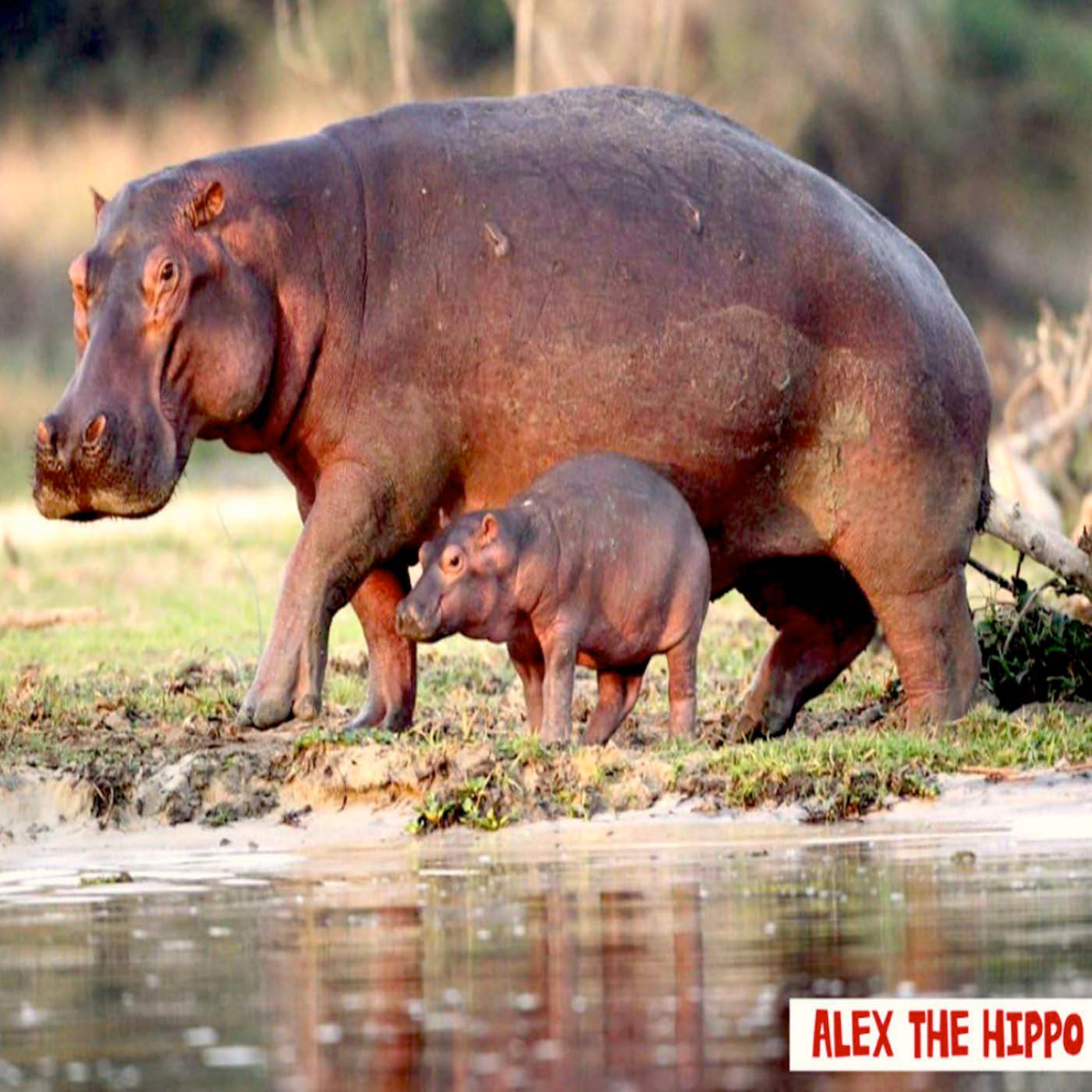}
    \end{subfigure}
    \hfill
    \begin{subfigure}{0.12\textwidth}
        \centering
        \includegraphics[height=1.85cm]{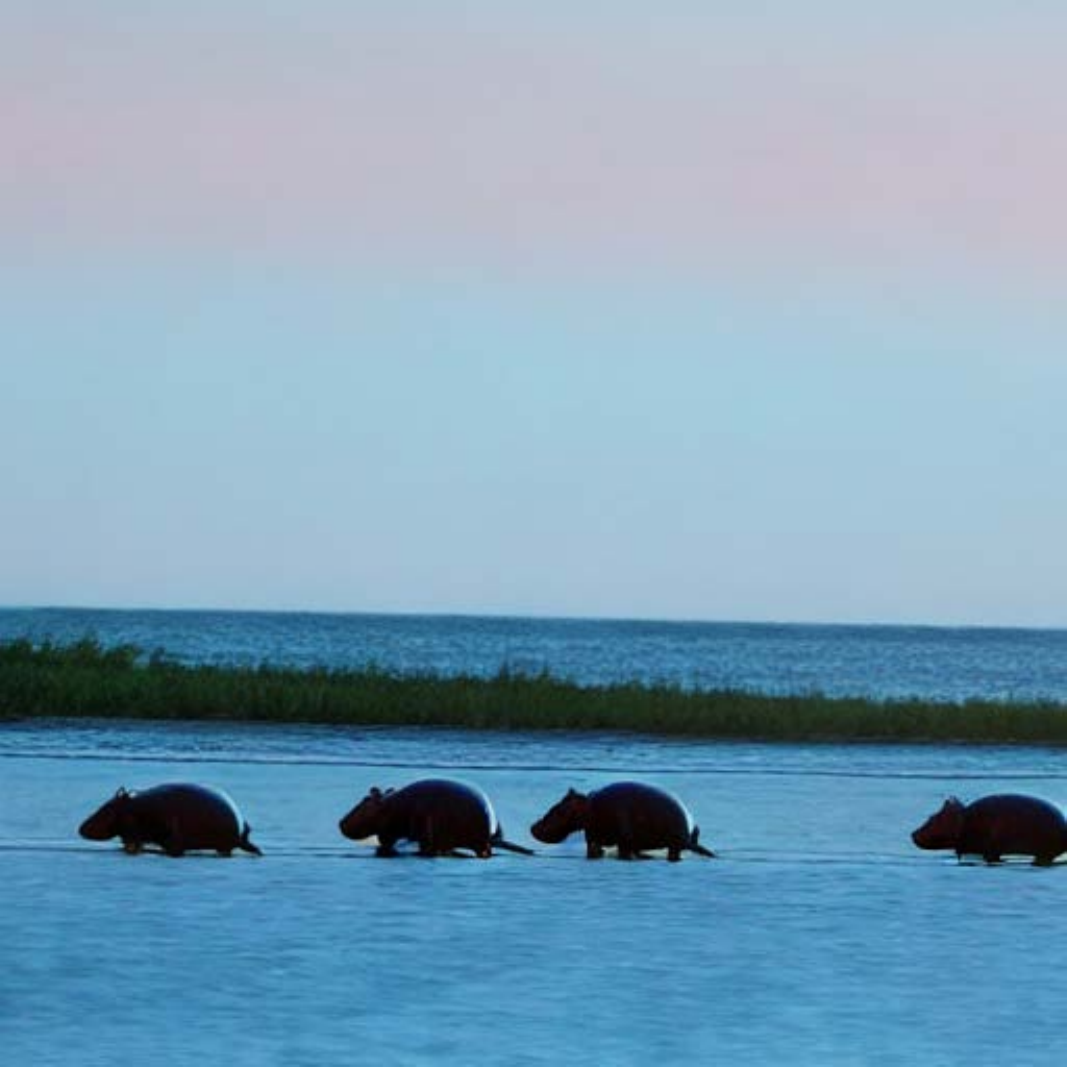}
    \end{subfigure}
    \hfill
    \begin{subfigure}{0.12\textwidth}
        \centering
        \includegraphics[height=1.85cm]{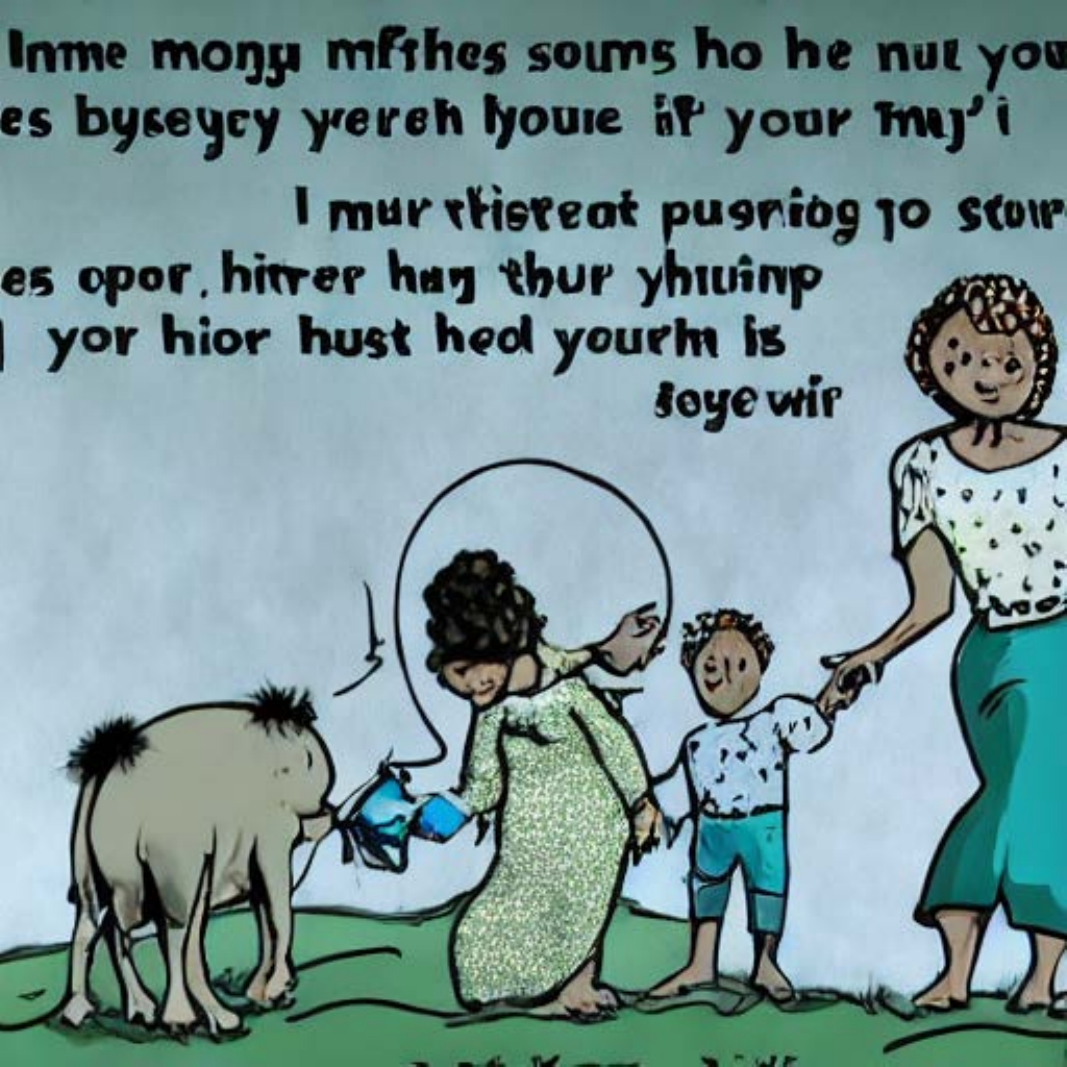}
    \end{subfigure}
    \hfill
    \begin{subfigure}{0.12\textwidth}
        \centering
        \includegraphics[height=1.85cm]{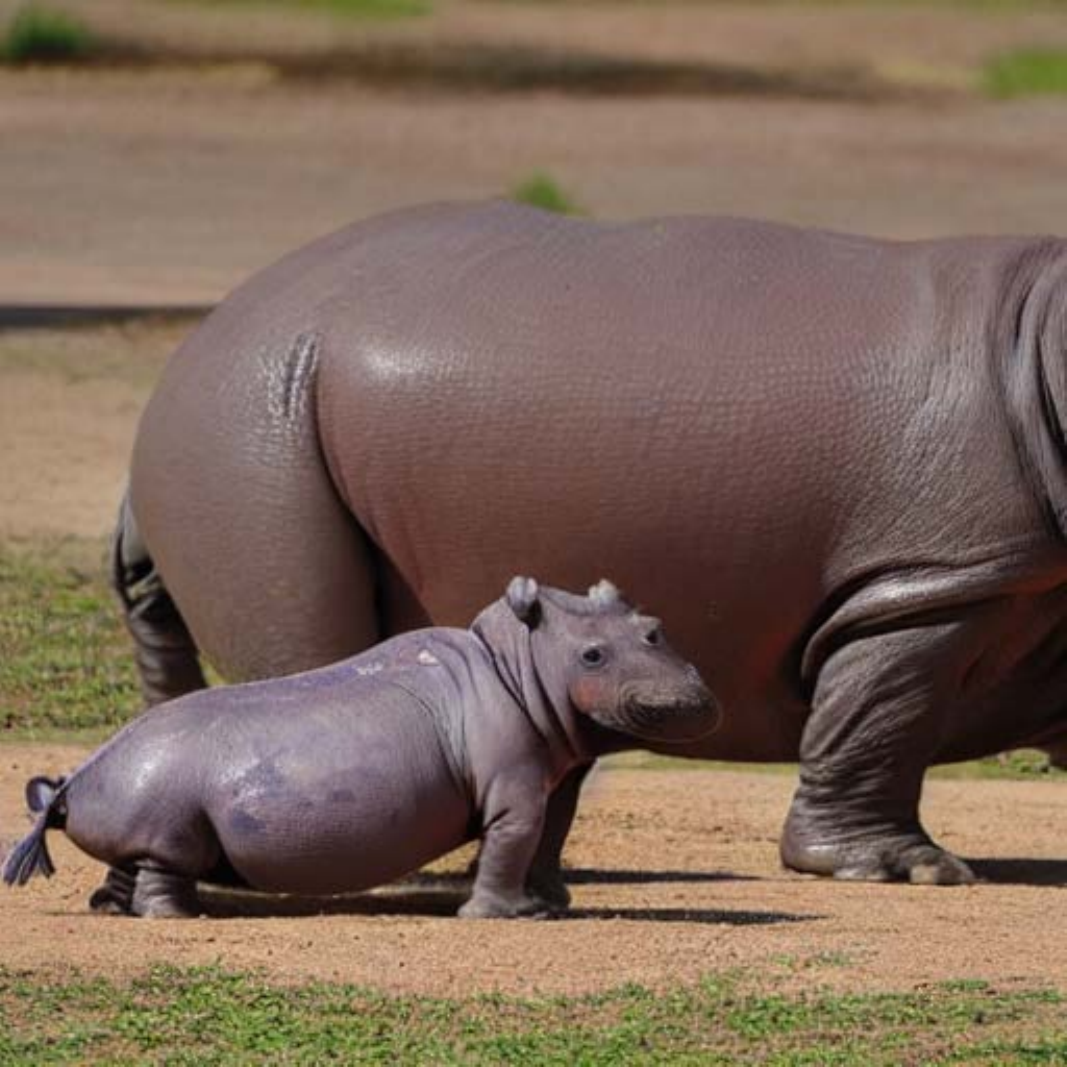}
    \end{subfigure}
    \hfill
    \begin{subfigure}{0.12\textwidth}
        \centering
        \includegraphics[height=1.85cm]{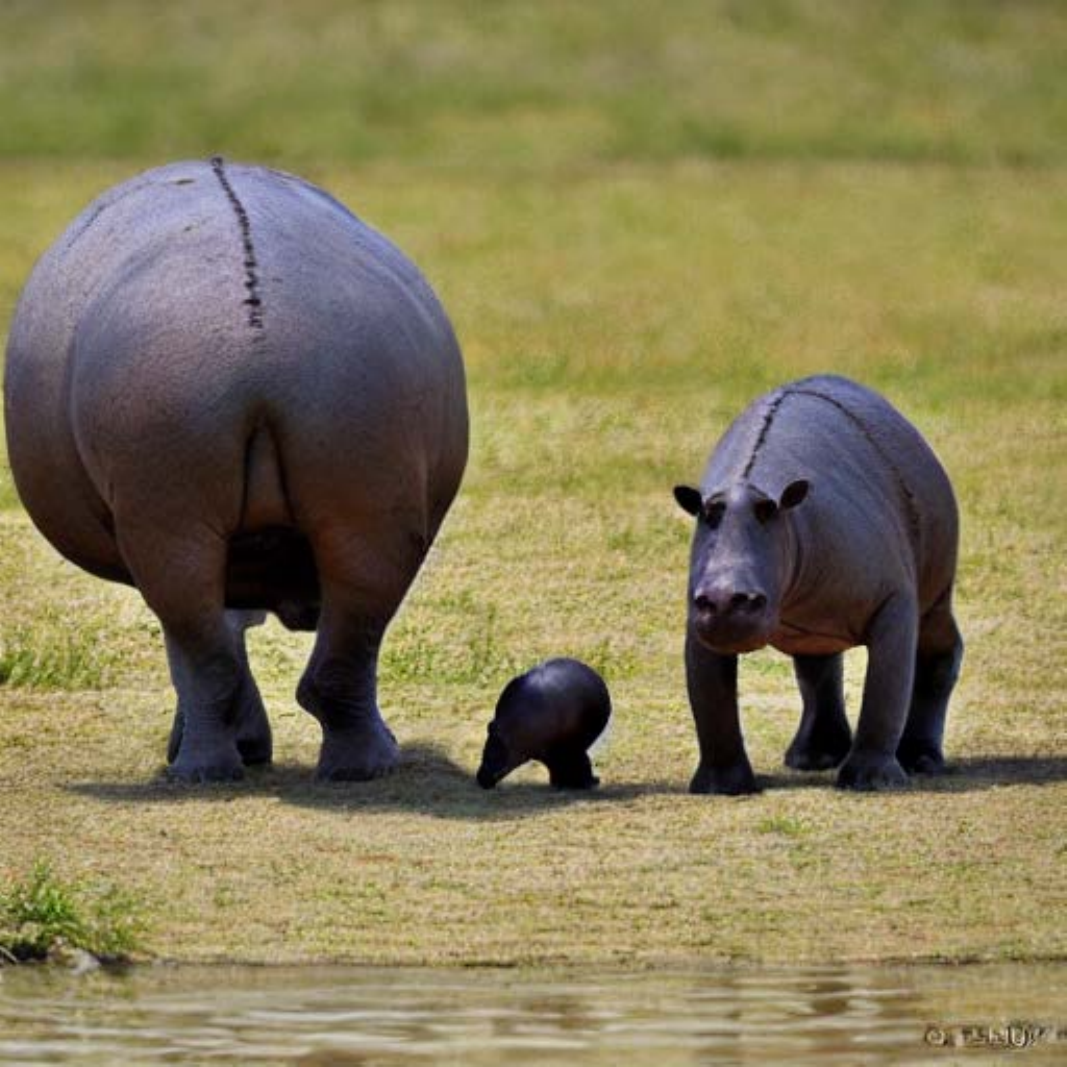}
    \end{subfigure}
    \hfill
    \begin{subfigure}{0.12\textwidth}
        \centering
        \includegraphics[height=1.85cm]{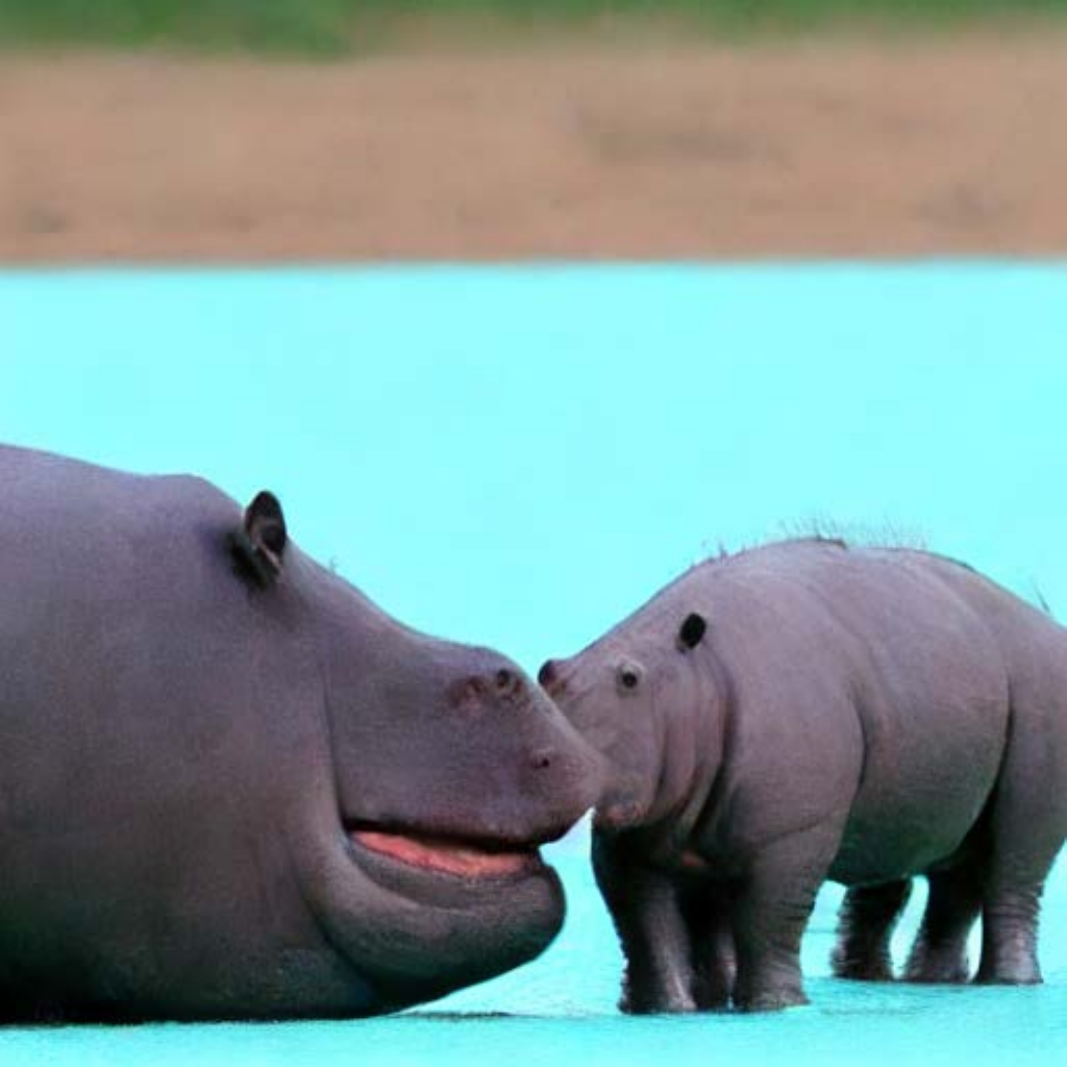}
    \end{subfigure}
    \hfill
    \begin{subfigure}{0.12\textwidth}
        \centering
        \includegraphics[height=1.85cm]{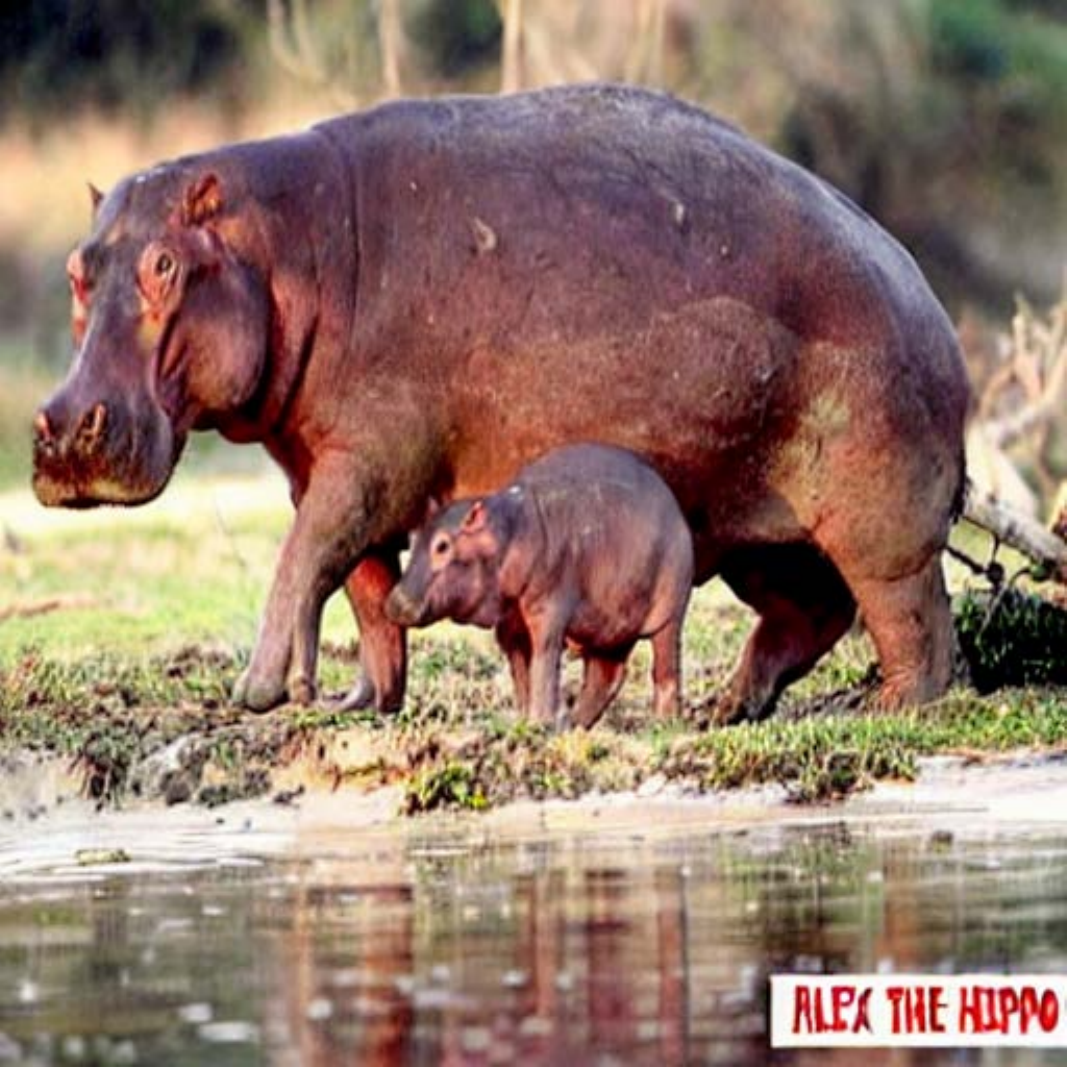}
    \end{subfigure}
    \hfill
    \begin{subfigure}{0.12\textwidth}
        \centering
        \includegraphics[height=1.85cm]{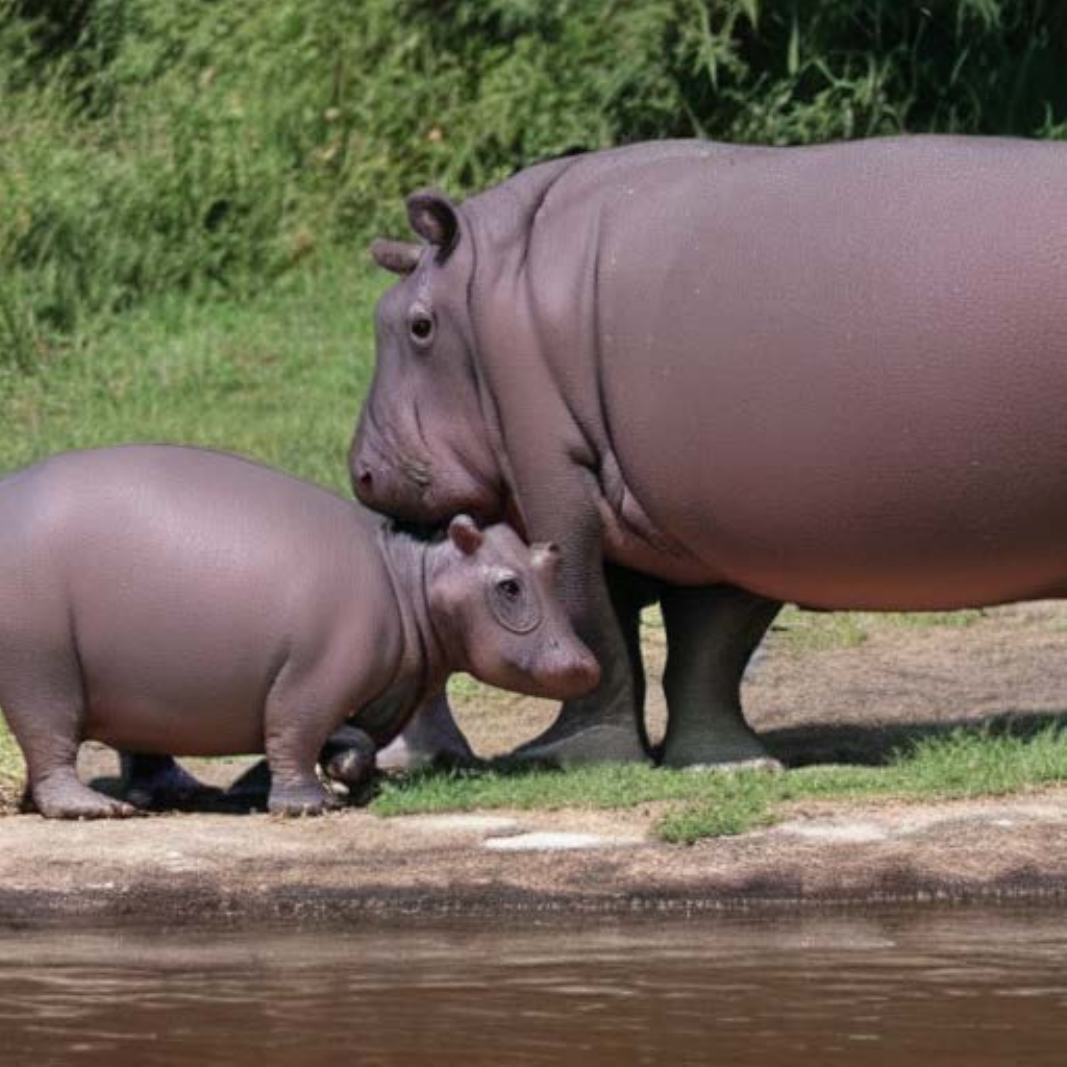}
    \end{subfigure}

    \vspace{0.3em}
    \begin{subfigure}{0.12\textwidth}
        \centering
        \includegraphics[height=1.85cm]{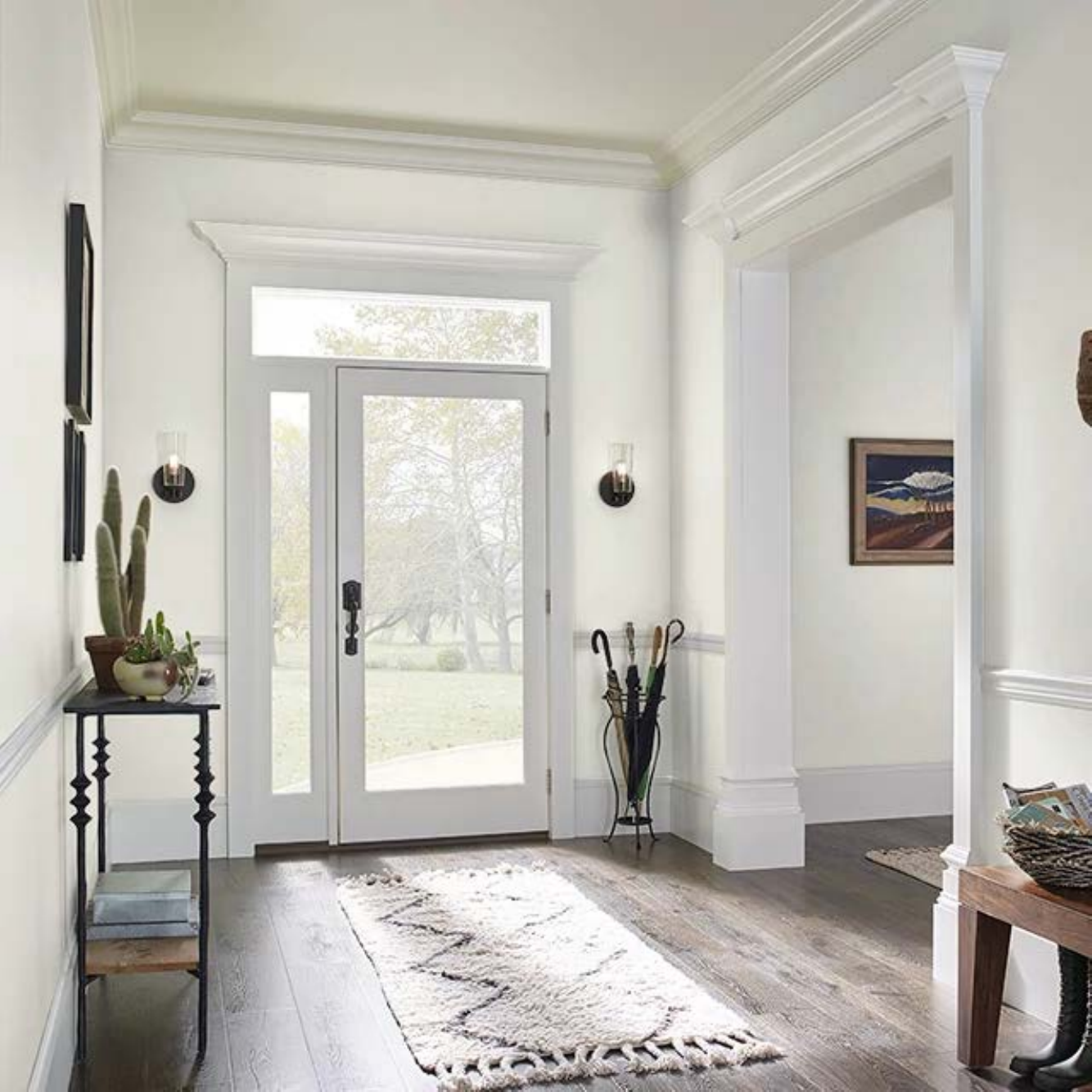}
    \end{subfigure}
    \hfill
    \begin{subfigure}{0.12\textwidth}
        \centering
        \includegraphics[height=1.85cm]{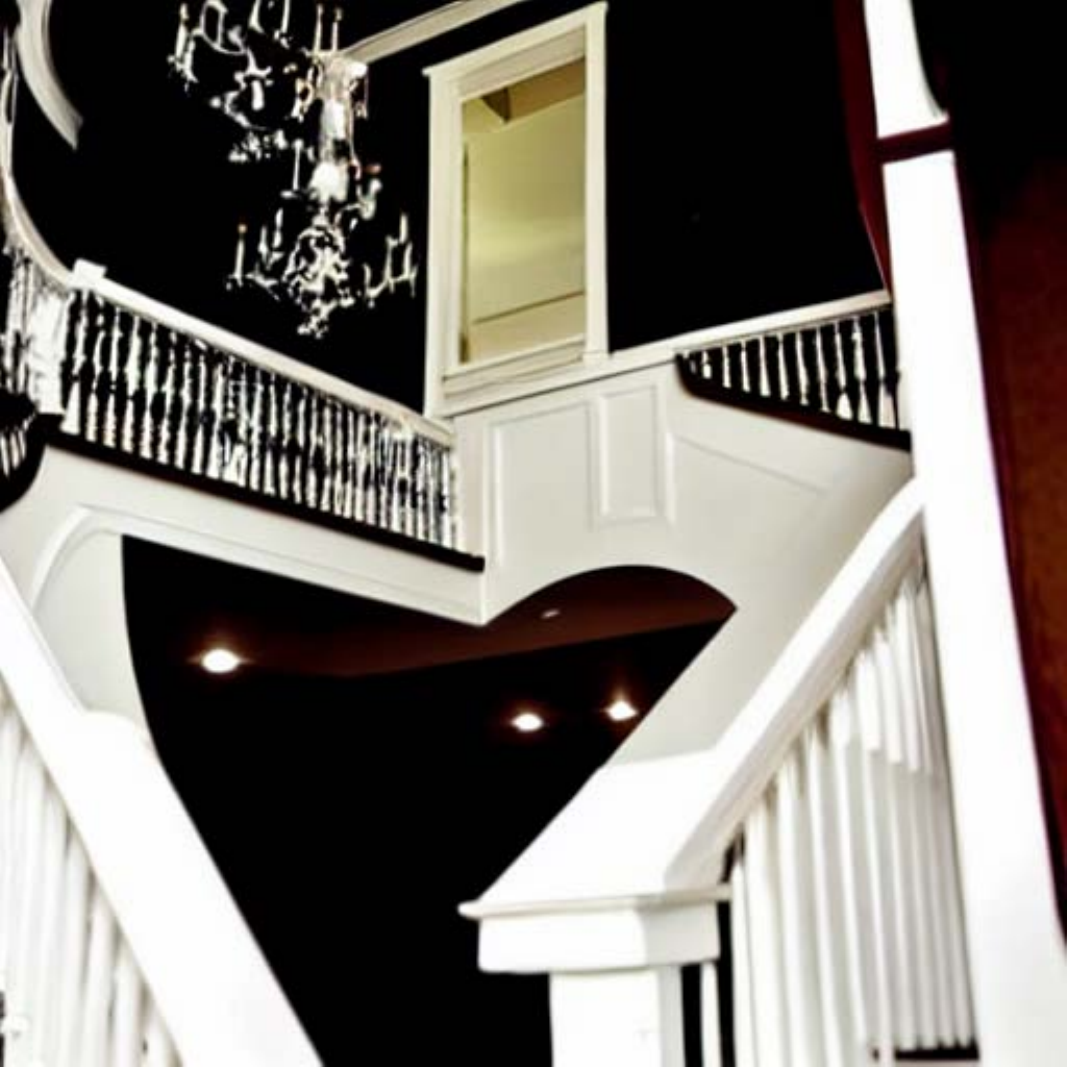}
    \end{subfigure}
    \hfill
    \begin{subfigure}{0.12\textwidth}
        \centering
        \includegraphics[height=1.85cm]{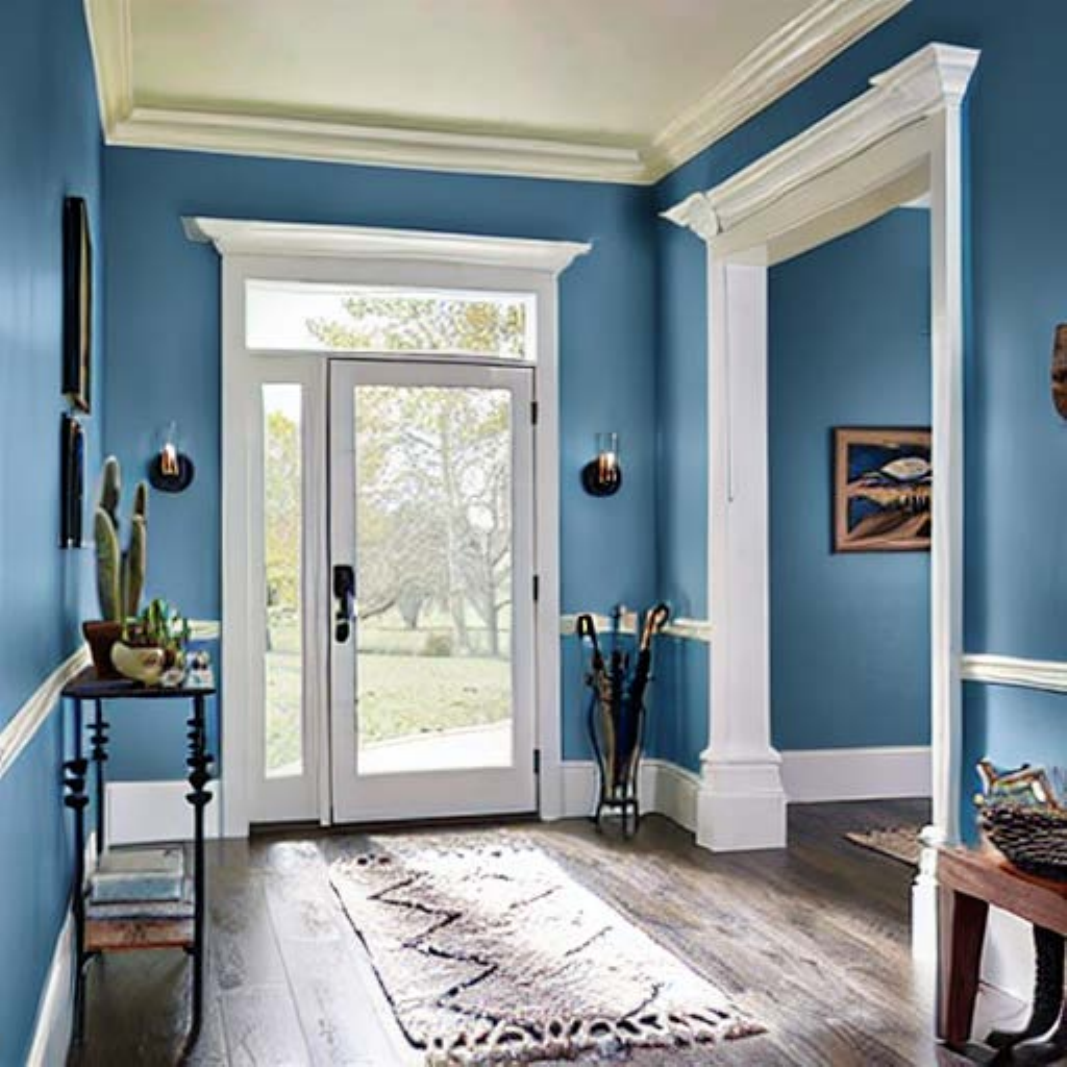}
    \end{subfigure}
    \hfill
    \begin{subfigure}{0.12\textwidth}
        \centering
        \includegraphics[height=1.85cm]{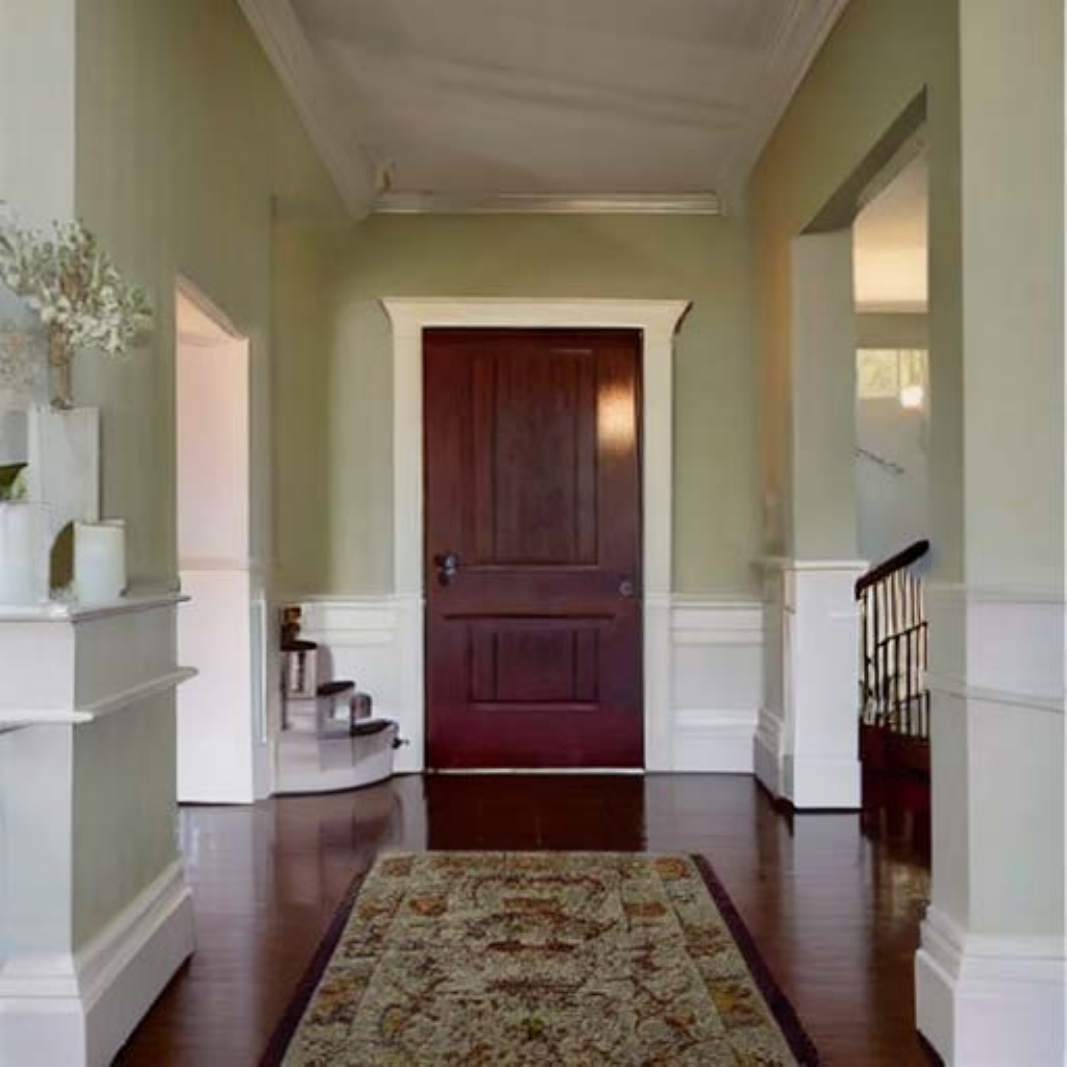}
    \end{subfigure}
    \hfill
    \begin{subfigure}{0.12\textwidth}
        \centering
        \includegraphics[height=1.85cm]{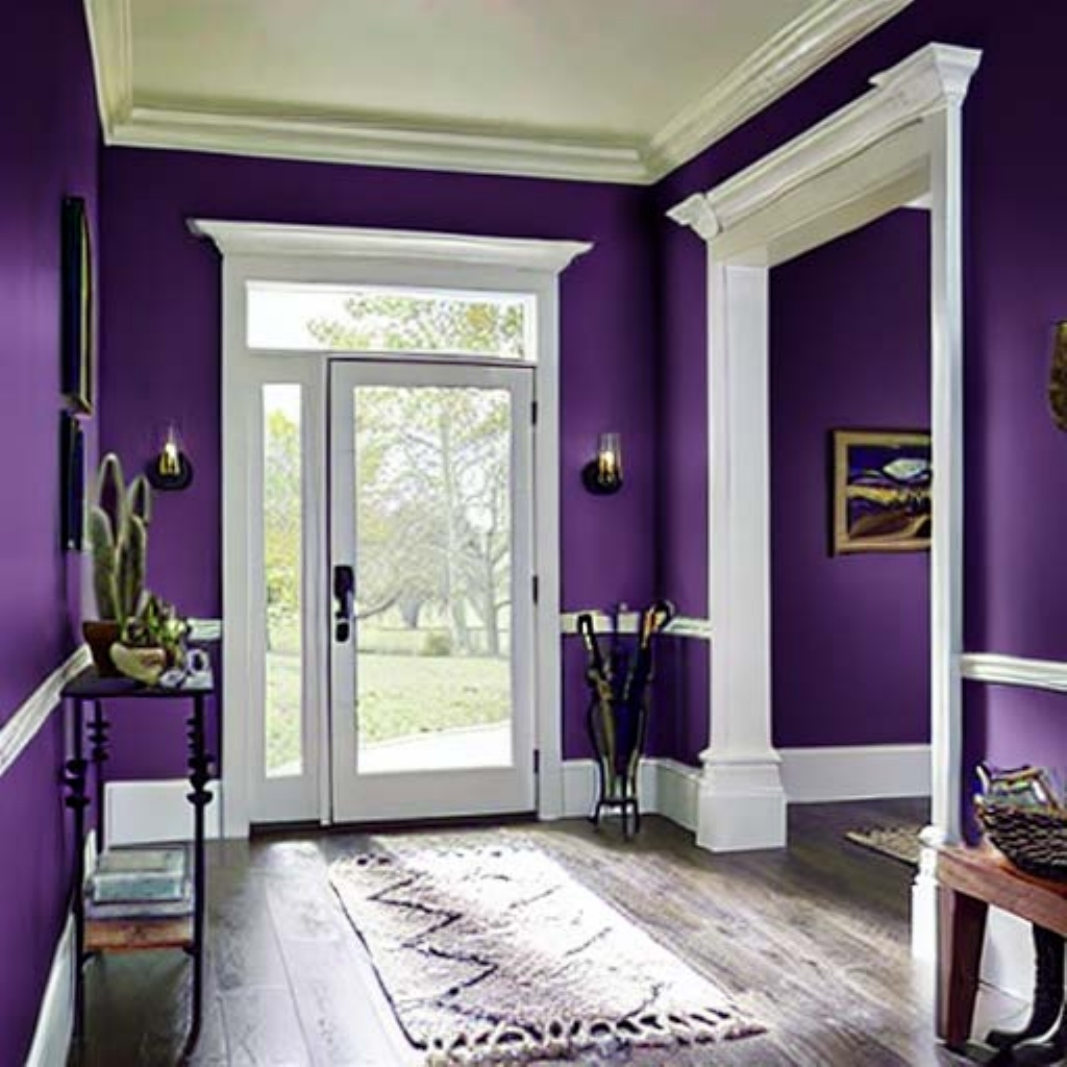}
    \end{subfigure}
    \hfill
    \begin{subfigure}{0.12\textwidth}
        \centering
        \includegraphics[height=1.85cm]{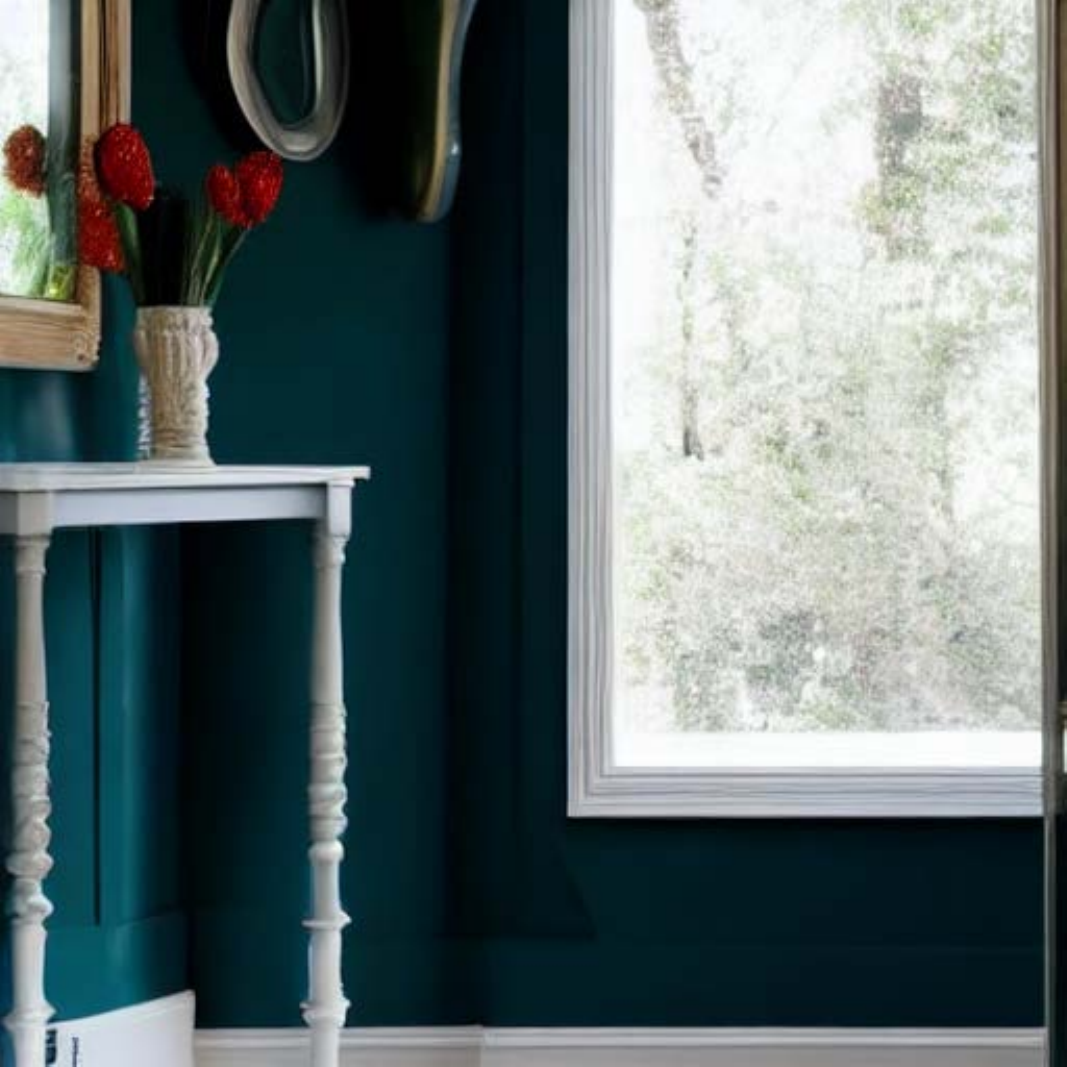}
    \end{subfigure}
    \hfill
    \begin{subfigure}{0.12\textwidth}
        \centering
        \includegraphics[height=1.85cm]{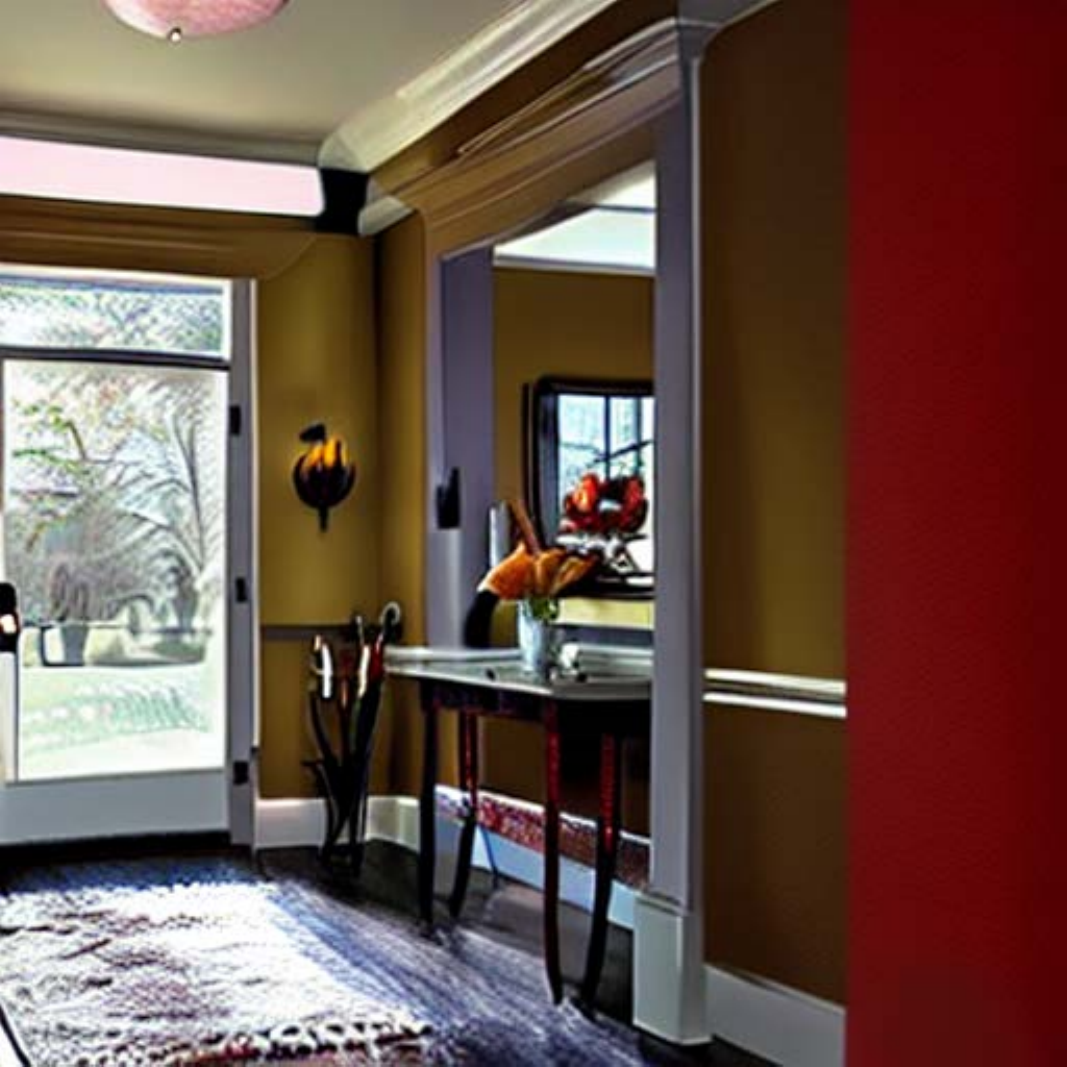}
    \end{subfigure}
    \hfill
    \begin{subfigure}{0.12\textwidth}
        \centering
        \includegraphics[height=1.85cm]{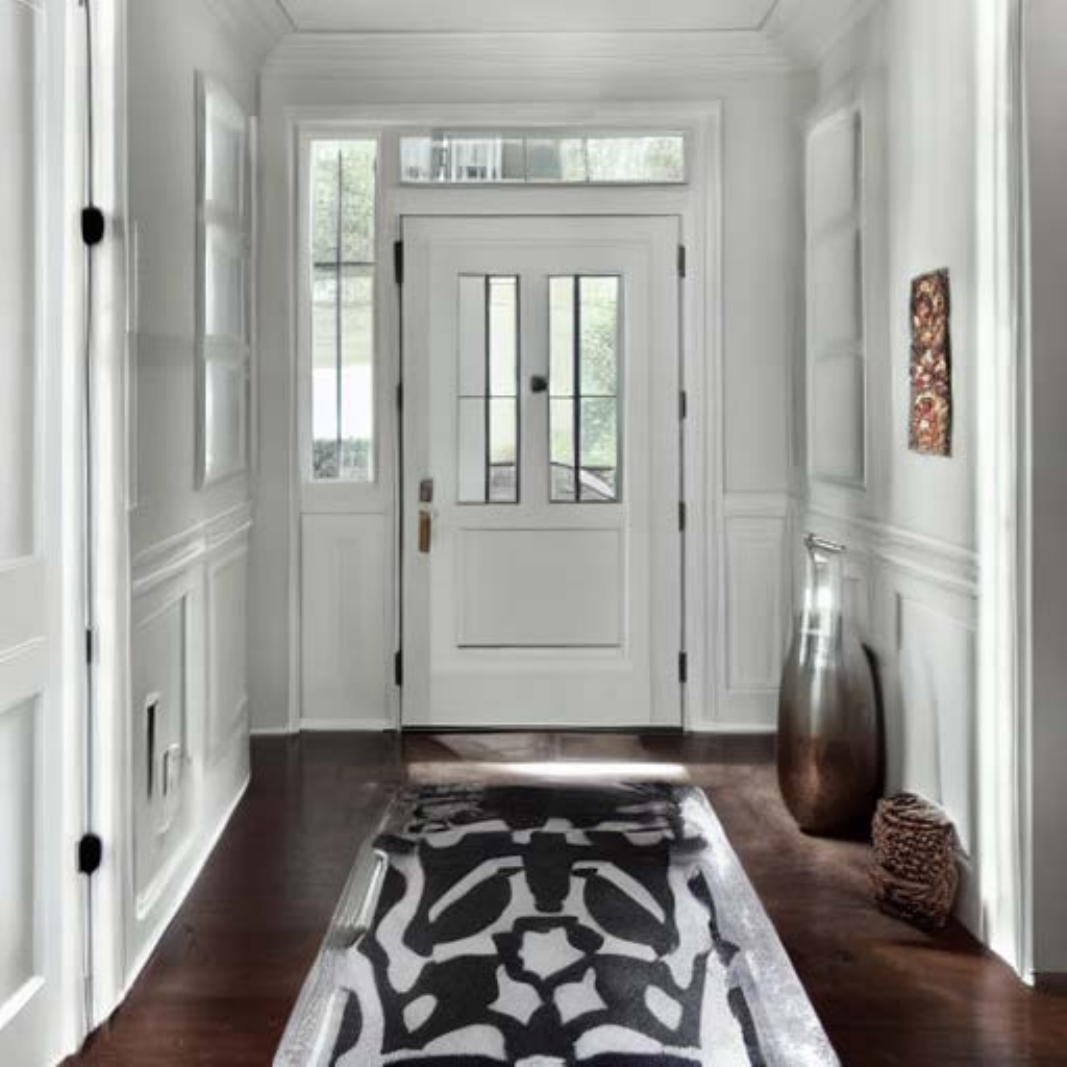}
    \end{subfigure}

    \vspace{0.3em}
    \begin{subfigure}{0.12\textwidth}
        \centering
        \includegraphics[height=1.85cm]{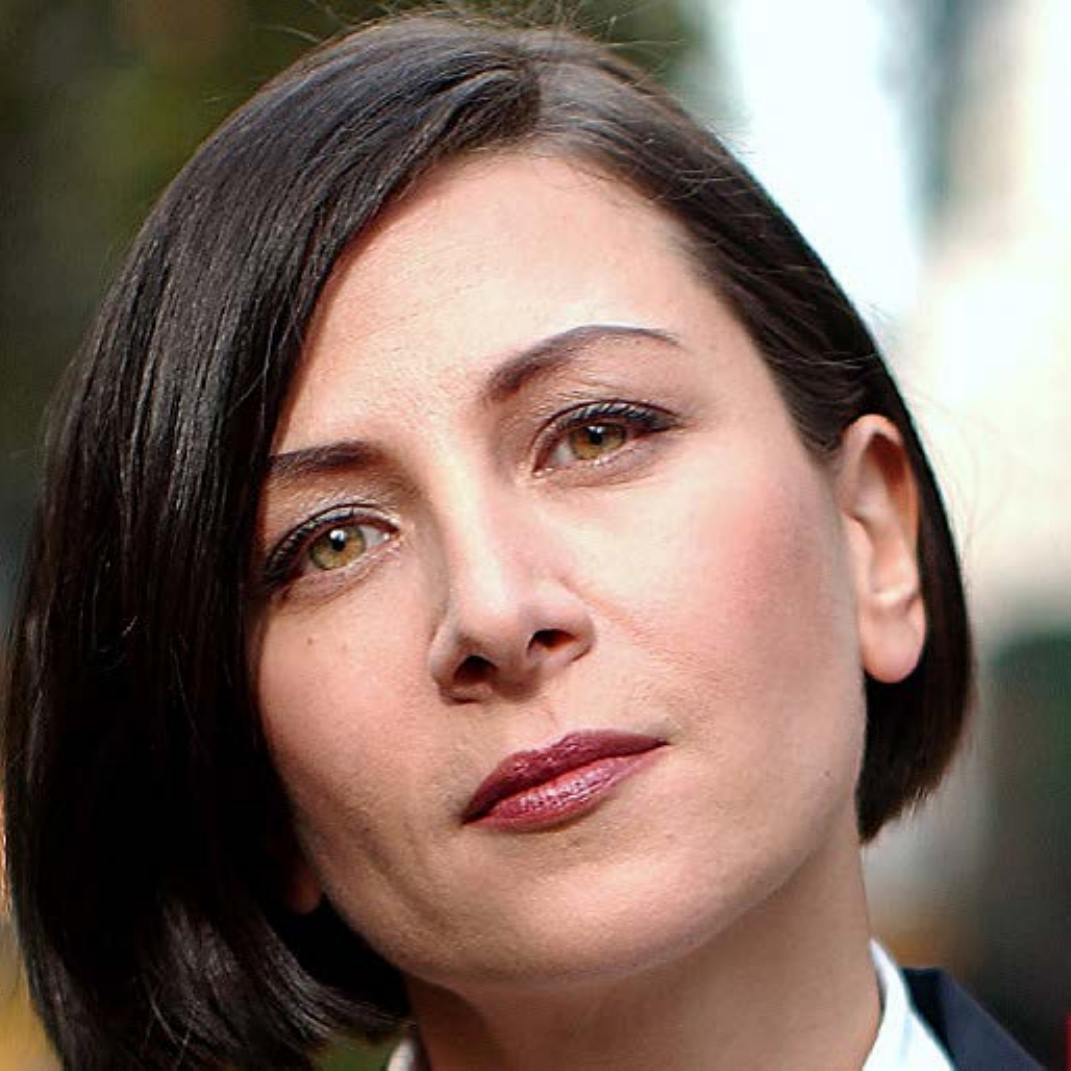}
    \end{subfigure}
    \hfill
    \begin{subfigure}{0.12\textwidth}
        \centering
        \includegraphics[height=1.85cm]{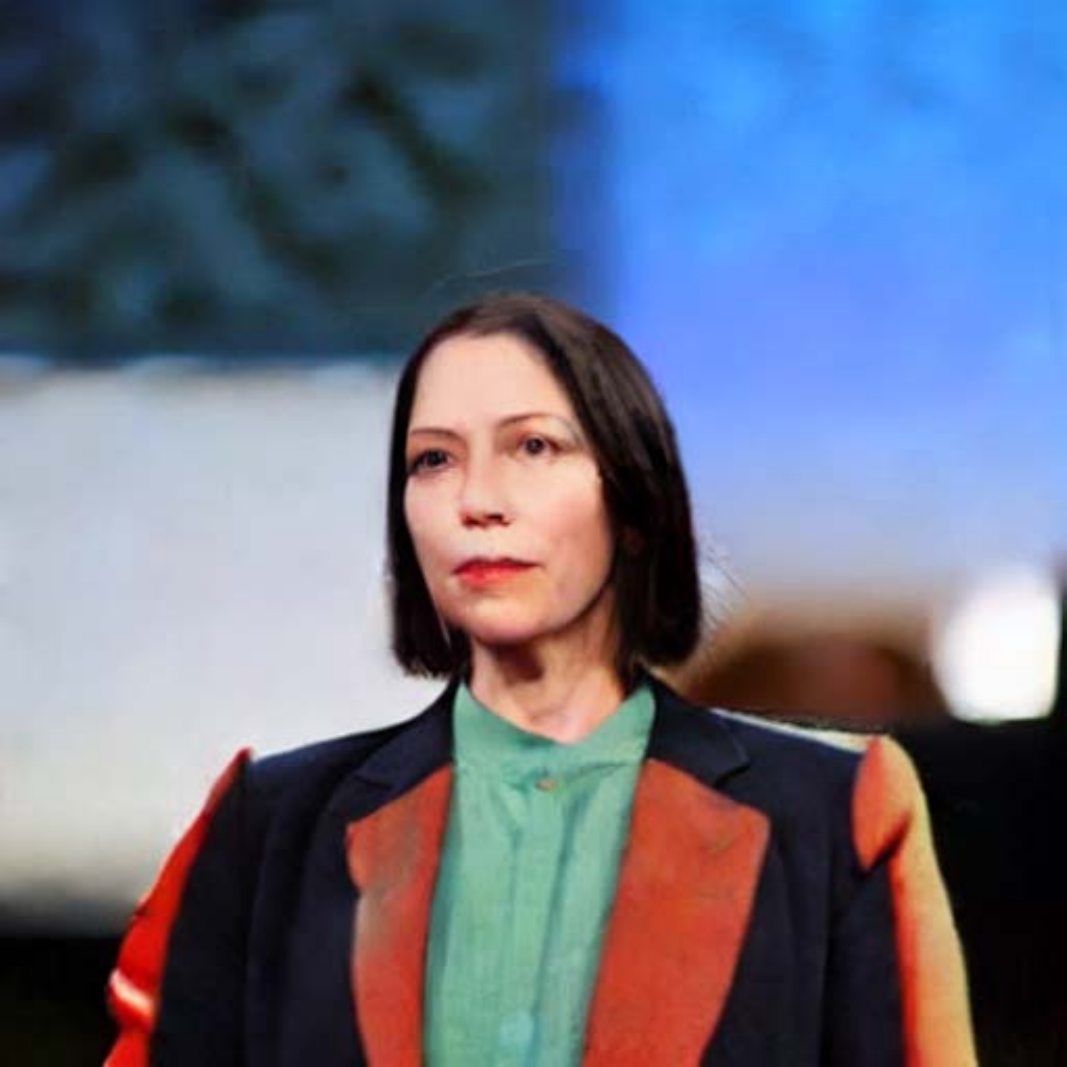}
    \end{subfigure}
    \hfill
    \begin{subfigure}{0.12\textwidth}
        \centering
        \includegraphics[height=1.85cm]{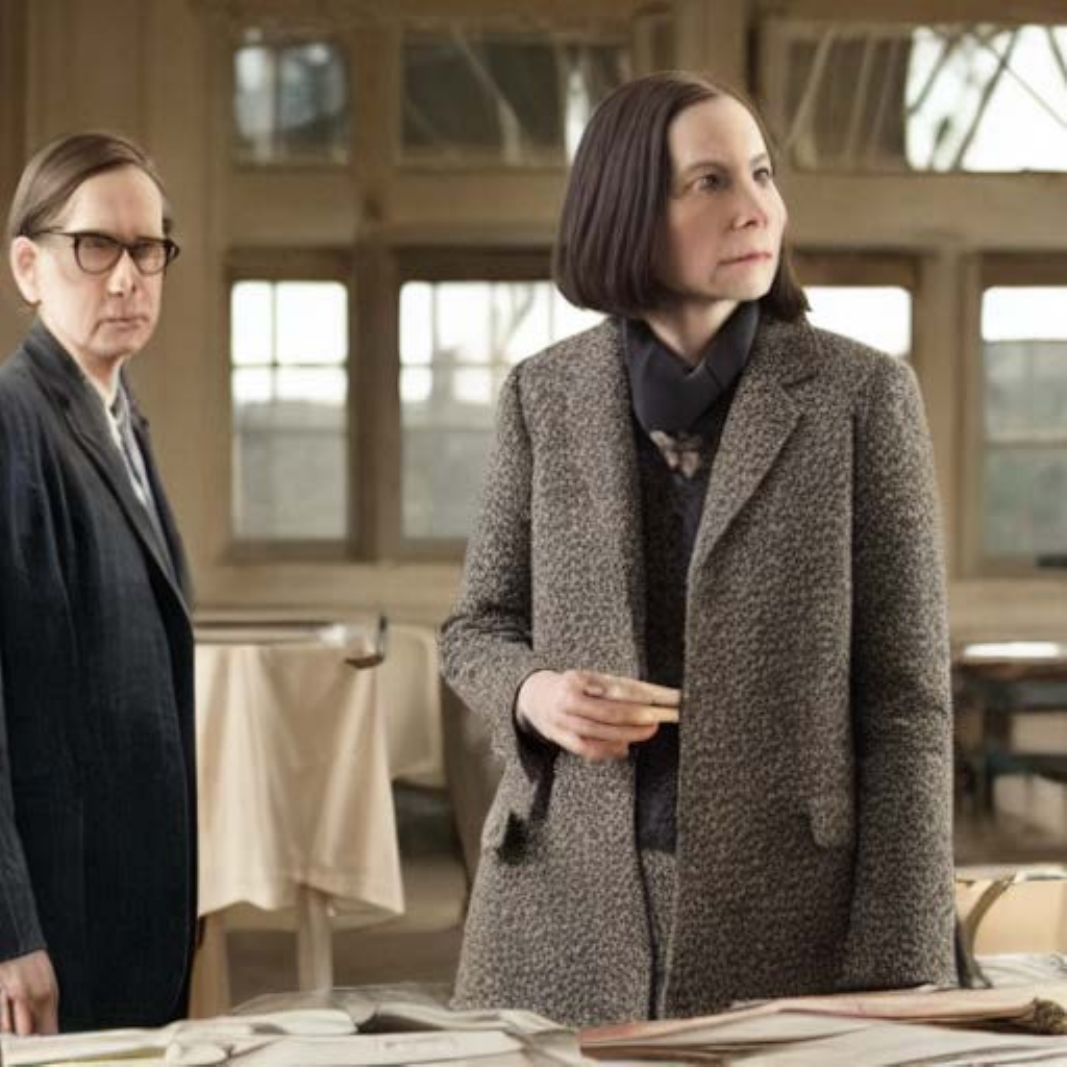}
    \end{subfigure}
    \hfill
    \begin{subfigure}{0.12\textwidth}
        \centering
        \includegraphics[height=1.85cm]{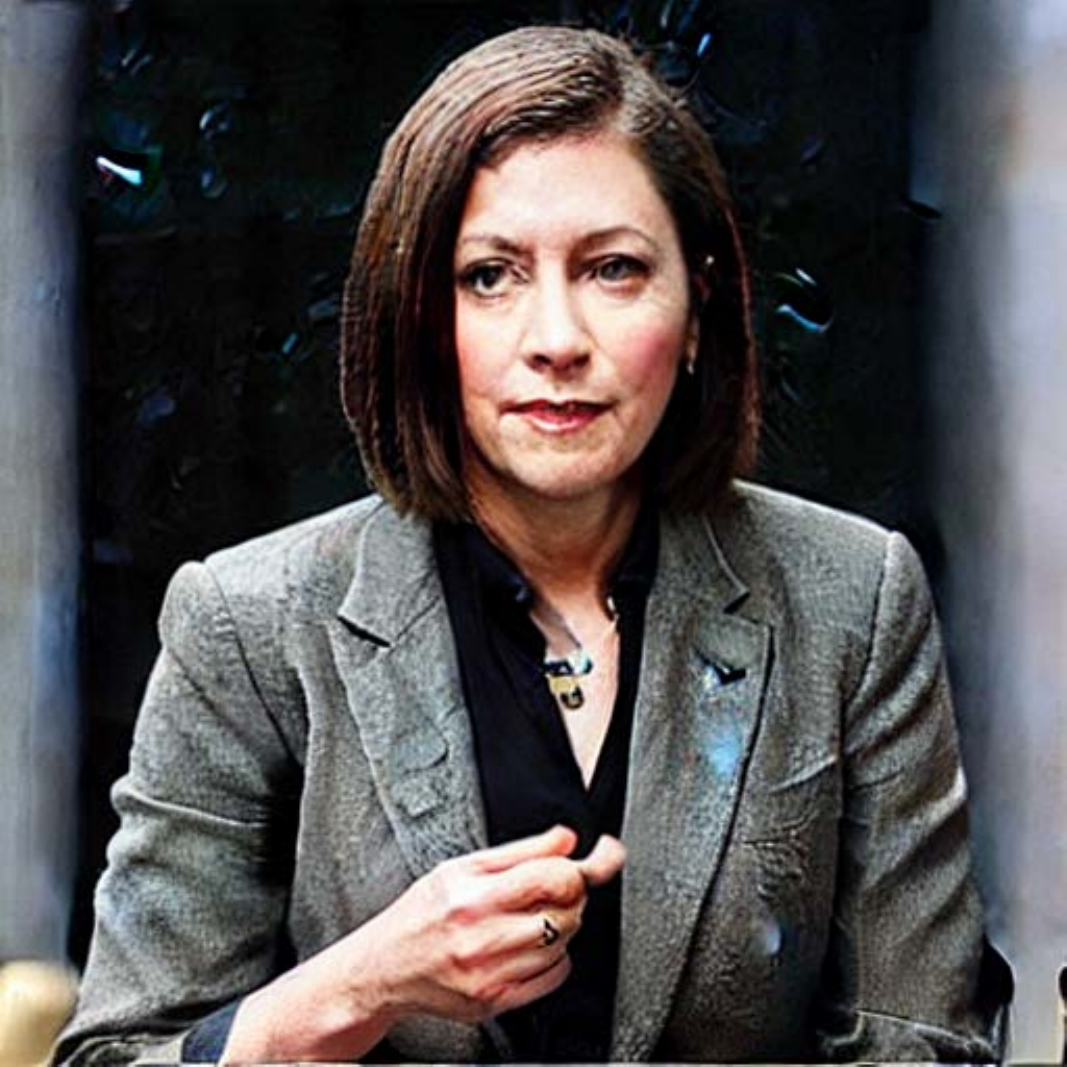}
    \end{subfigure}
    \hfill
    \begin{subfigure}{0.12\textwidth}
        \centering
        \includegraphics[height=1.85cm]{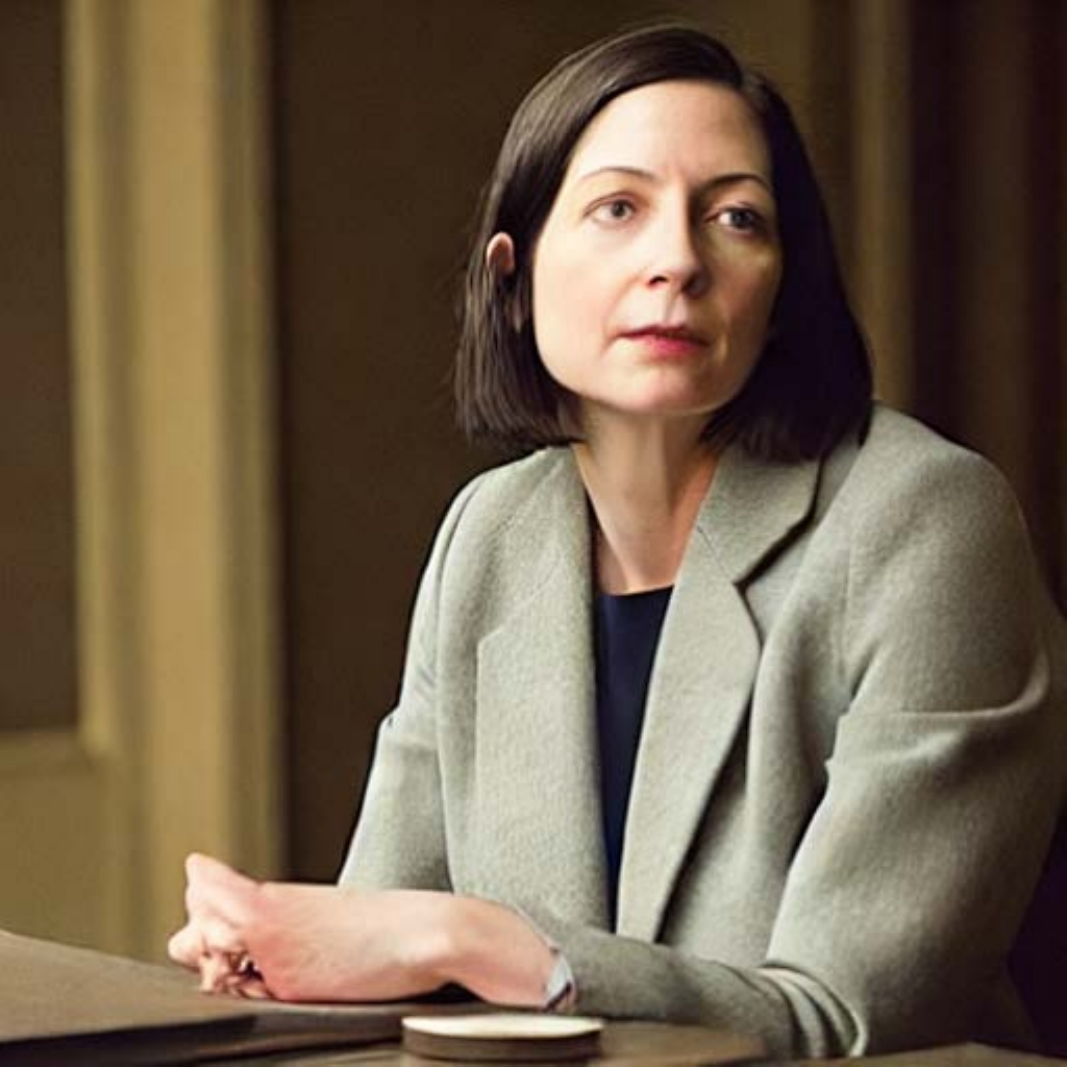}
    \end{subfigure}
    \hfill
    \begin{subfigure}{0.12\textwidth}
        \centering
        \includegraphics[height=1.85cm]{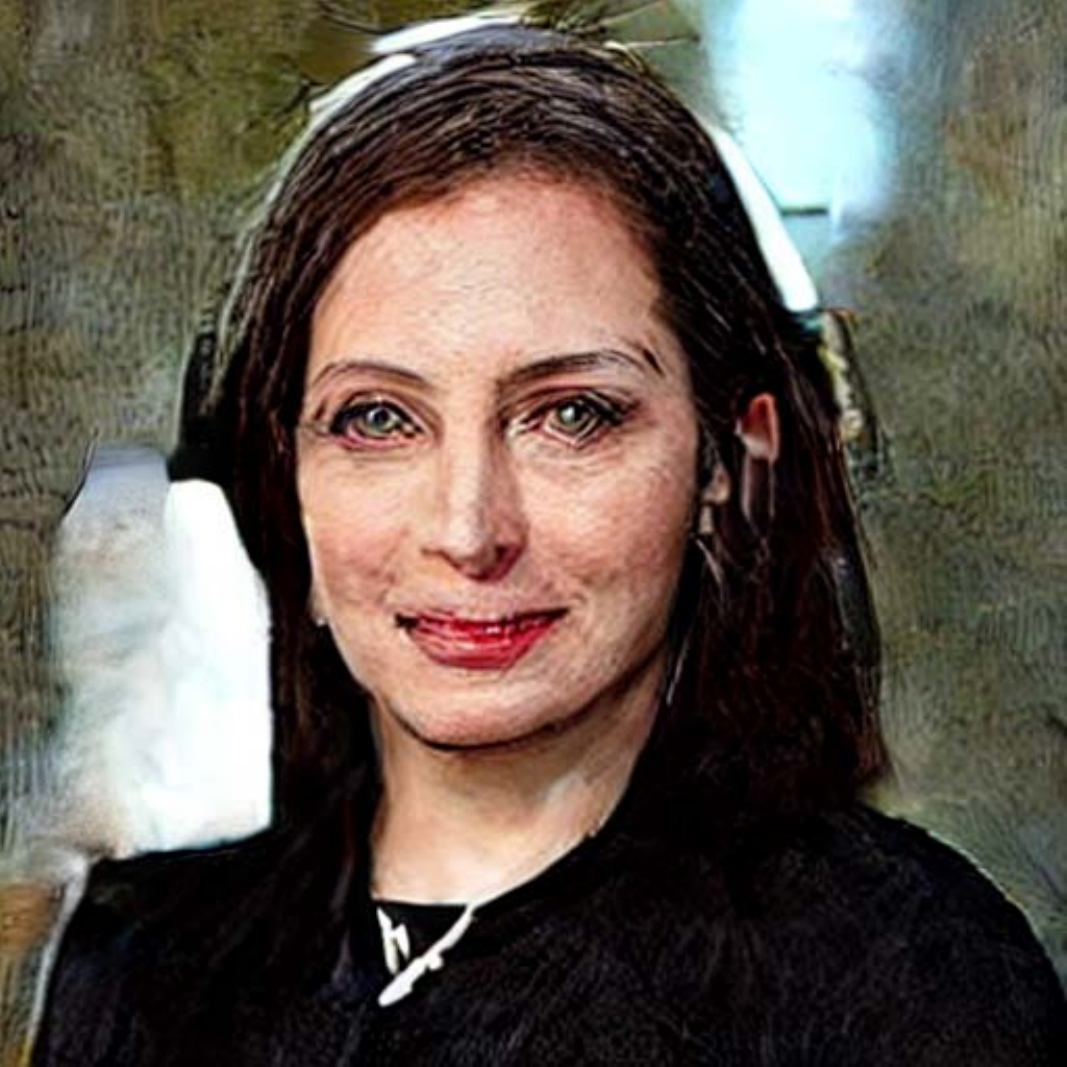}
    \end{subfigure}
    \hfill
    \begin{subfigure}{0.12\textwidth}
        \centering
        \includegraphics[height=1.85cm]{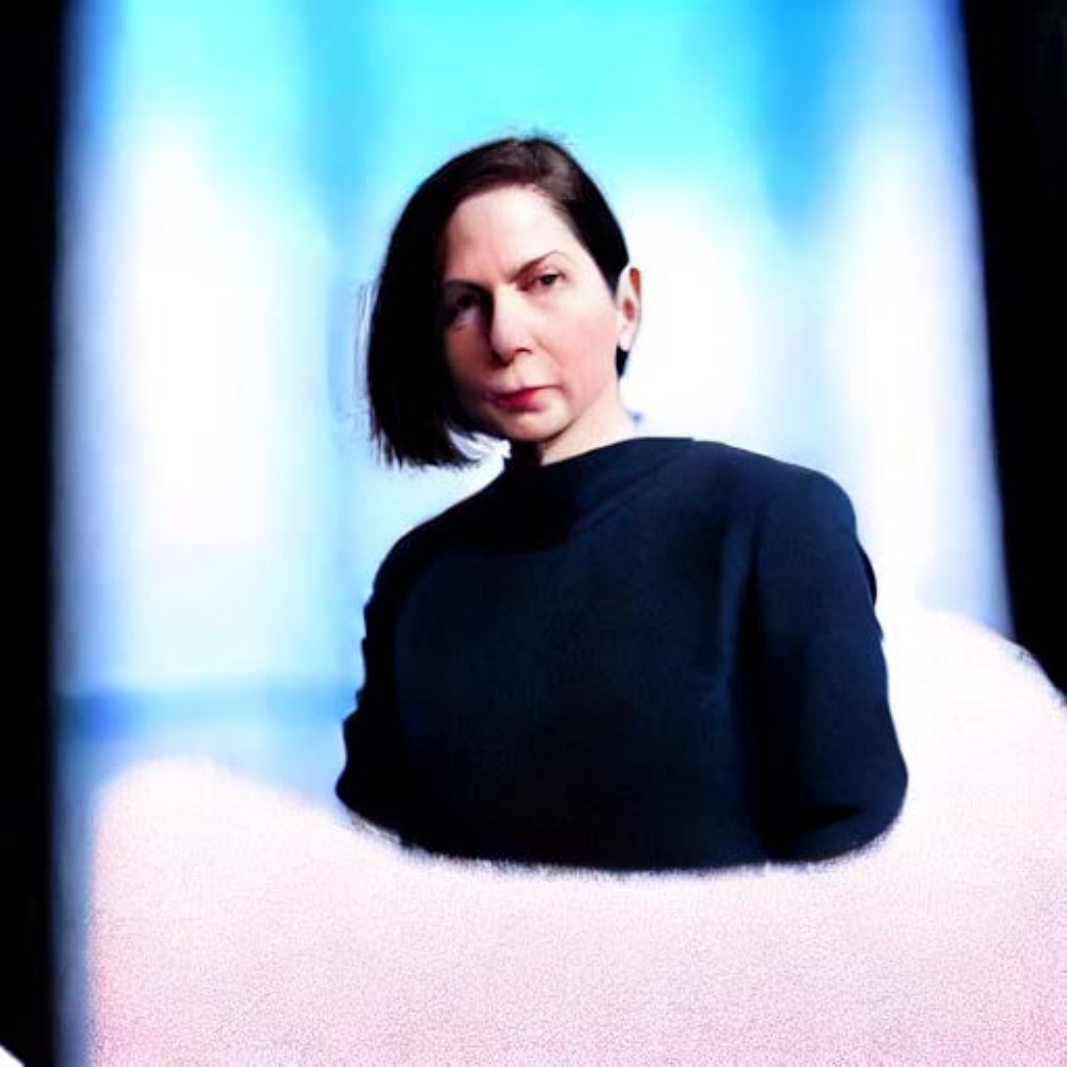}
    \end{subfigure}
    \hfill
    \begin{subfigure}{0.12\textwidth}
        \centering
        \includegraphics[height=1.85cm]{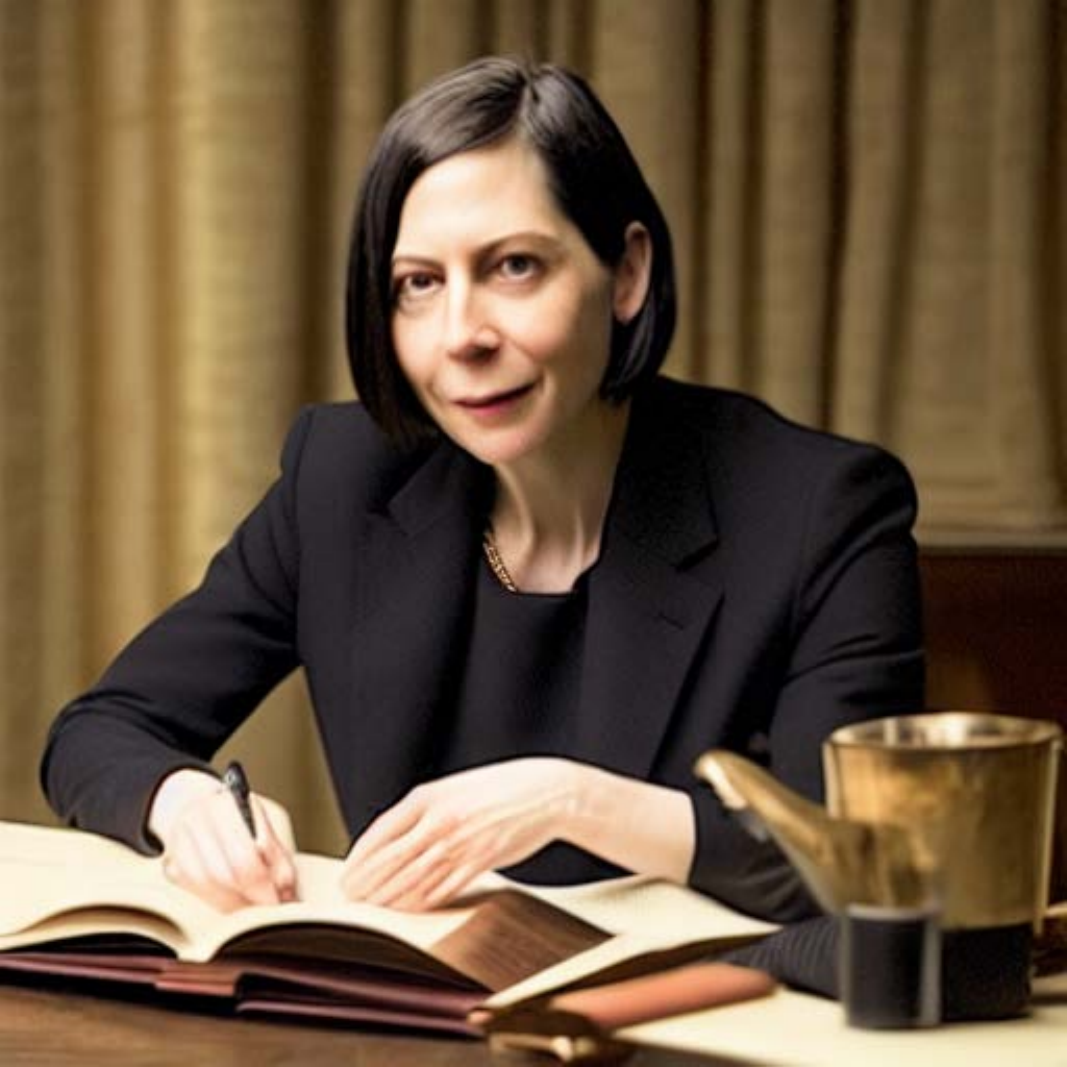}
    \end{subfigure}

    \vspace{0.3em}
    \begin{subfigure}{0.12\textwidth}
        \centering
        Training
    \end{subfigure}
 \hfill
    \begin{subfigure}{0.12\textwidth}
        \centering
        DOG
    \end{subfigure}
 \hfill
    \begin{subfigure}{0.12\textwidth}
        \centering
        PEA
    \end{subfigure}
 \hfill
    \begin{subfigure}{0.12\textwidth}
        \centering
        RTA 
    \end{subfigure}
 \hfill
    \begin{subfigure}{0.12\textwidth}
        \centering
        CALS 
    \end{subfigure}
 \hfill
    \begin{subfigure}{0.12\textwidth}
        \centering
        IND 
    \end{subfigure}
 \hfill
    \begin{subfigure}{0.12\textwidth}
        \centering
        SALI 
    \end{subfigure}
 \hfill
    \begin{subfigure}{0.12\textwidth}
        \centering
        \textbf{Ours}
    \end{subfigure}
    \vspace{-5pt}
    \caption{Qualitative comparison with baselines. The first column shows the memorized training images, while the remaining seven columns present the results of baselines and our method. Our method achieves a superior balance between memorization mitigation, semantic consistency, and image quality compared with existing baselines.}
    \label{Fig32}
    \vspace{-10pt}
\end{figure*}

\subsection{Layer-Wise Ablation Study} 

\paragraph{Progressive Layers Suppression.} In cross-attention analysis of memorization, we observe that memorized prompts produce stronger semantic responses in shallow layers. To investigate the layer-wise contribution to memorization, we progressively apply our suppression strategy to the 16 cross-attention layers, beginning with the shallowest layers. As shown in Tab.~\ref{Ablation_layers}, suppressing layers 0--3 substantially reduces the SSCD score from $0.5122$ to $0.1578$ for SD v1.4 and from $0.6306$ to $0.2662$ for SD v2.0, while maintaining CLIP scores of $0.2590$ and $0.2730$, respectively. Extending suppression to deeper layers further reduces SSCD, but only marginally. In particular, suppressing all layers achieves the lowest SSCD scores but reduces the CLIP scores to $0.1738$ and $0.2207$. These results indicate that shallow cross-attention layers account for most of the memorization effect, whereas suppressing deeper layers provides diminishing mitigation benefits while increasingly compromising semantic fidelity. 

\begin{table}[h]
\caption{Progressive Layers ablation study for mitigation. Suppressing layers 0--3 provides the most substantial mitigation, while extending suppression to deeper layers yields diminishing benefits at the cost of semantic fidelity.}
\label{Ablation_layers}
\centering
\setlength{\tabcolsep}{10pt}
\begin{tabular}{@{}c|cc|cc@{}}
\toprule
\multirow{2}{*}{Layers}
& \multicolumn{2}{c}{SD v1.4}
& \multicolumn{2}{c}{SD v2.0} \\
\cmidrule(lr){2-3}
\cmidrule(lr){4-5}
& SSCD $\downarrow$ & CLIP $\uparrow$
& SSCD $\downarrow$ & CLIP $\uparrow$ \\
\midrule
None  & 0.5122 & 0.3009 & 0.6306 & 0.2947 \\
\midrule
0--1  & 0.3742 & 0.2887 & 0.5975 & 0.2973 \\
\rowcolor[gray]{0.9}
0--3  & 0.1578 & 0.2590 & 0.2662 & 0.2730 \\
0--5  & 0.1301 & 0.2446 & 0.2067 & 0.2564 \\
0--7  & 0.1098 & 0.2264 & 0.1602 & 0.2465 \\
0--9  & 0.0990 & 0.2117 & 0.1580 & 0.2403 \\
0--11 & 0.0888 & 0.2046 & 0.1500 & 0.2215 \\
0--13 & 0.0873 & 0.2035 & 0.1492 & 0.2208 \\
0--15 & 0.0690 & 0.1738 & 0.1463 & 0.2207 \\
\bottomrule
\end{tabular}
\vspace{-5pt}
\end{table}

\paragraph{Segmented Layers Ablation Study.}
\label{Segmented}
We divide the 16 cross-attention layers into four groups: early layers (0 $\sim$ 3), early-middle layers (4 $\sim$ 7), late-middle layers (8 $\sim$ 11), and late layers (12 $\sim$ 15), and apply our suppression strategy to each group separately. As shown in Tab.~\ref{Segmented_layers}, suppressing only the early layers reduces the SSCD score of SD v1.4 from $0.5122$ to $0.1578$, achieving the strongest mitigation while retaining a CLIP score of $0.2590$. In contrast, suppressing later layers yields much weaker mitigation. Even suppressing all remaining layers (4 $\sim$ 15) is less effective than suppressing only layers 0 $\sim$ 3 and causes a severe drop in CLIP score. These results further confirm that memorization is mainly concentrated in shallow cross-attention layers rather than uniformly distributed across the network.

\vspace{-5pt}
\begin{table}[h]
\caption{Segmented layers ablation study for mitigation. Early layers (0 $\sim$ 3) contribute most significantly to mitigation, while suppression in deeper layers yields limited benefits.}
\label{Segmented_layers}
\vspace{-5pt}
\centering
\setlength{\tabcolsep}{8pt}
\begin{tabular}{l|cc|cc}
\toprule
 \multirow{2}{*}{Layers} & \multicolumn{2}{c|}{SD v1.4}
 & \multicolumn{2}{c}{SD v2.0} \\
\cmidrule(lr){2-3}
\cmidrule(lr){4-5}
& SSCD $\downarrow$ & CLIP $\uparrow$
& SSCD $\downarrow$ & CLIP $\uparrow$ \\
\midrule
None & 0.5122 & 0.3009 & 0.6306 & 0.2947 \\
\midrule
\rowcolor[gray]{0.9}
0 $\sim$ 3 & 0.1578 & 0.2590 & 0.2662 & 0.2730 \\
4 $\sim$ 7 & 0.4260 & 0.2690 & 0.3978 & 0.2667 \\
8 $\sim$ 11 & 0.4412 & 0.2934 & 0.6165 & 0.2588 \\
12 $\sim$ 15 & 0.3962 & 0.2969 & 0.6264 & 0.2922 \\
4 $\sim$ 15 & 0.2175 & 0.1925 & 0.3697 & 0.2352 \\
0 $\sim$ 15 & 0.0690 & 0.1738 & 0.1463 & 0.2207 \\
\bottomrule
\end{tabular}%
\vspace{-5pt}
\end{table}

\subsection{Hyperparameter Sensitivity Analysis}
We investigate the sensitivity of our method to the two key hyperparameters, $k$ and the scale factor $\lambda$. As shown in Tab.~\ref{tab:hyperparameter_sweep}, increasing $k$ generally leads to lower SSCD scores with a slight decrease in CLIP scores, whereas decreasing $\lambda$ mildly improves CLIP scores while increasing SSCD scores. Importantly, both metrics vary smoothly across all evaluated configurations, without abrupt performance degradation. In particular, the CLIP scores remain within a narrow range of $0.2561$--$0.2639$, while consistently low SSCD scores are maintained under different values of $k$ and $\lambda$. These results demonstrate that our method is robust to the choice of hyperparameters and consistently achieves effective memorization mitigation across different settings.

\begin{table}[h]
\caption{Ablation study of $k$ and $\lambda$ on SD v1.4. The stable performance across different settings demonstrates the robustness of our method to hyperparameter choices.}
\label{tab:hyperparameter_sweep}
\centering
{
\setlength{\tabcolsep}{2pt}
\begin{tabular}{@{}c cc cc cc@{}}
\toprule
\multirow{2}{*}{$\lambda$}
& \multicolumn{2}{c}{$k=2$}
& \multicolumn{2}{c}{$k=3$}
& \multicolumn{2}{c}{$k=4$} \\
\cmidrule(lr){2-3}
\cmidrule(lr){4-5}
\cmidrule(lr){6-7}
& SSCD $\downarrow$ & CLIP $\uparrow$
& SSCD $\downarrow$ & CLIP $\uparrow$
& SSCD $\downarrow$ & CLIP $\uparrow$ \\
\midrule
1.00 & 0.1692 & 0.2610 & 0.1578 & 0.2590 & 0.1534 & 0.2576 \\
0.95 & 0.1697 & 0.2603 & 0.1590 & 0.2592 & 0.1554 & 0.2581 \\
0.90 & 0.1718 & 0.2607 & 0.1580 & 0.2593 & 0.1558 & 0.2589 \\
0.85 & 0.1752 & 0.2611 & 0.1602 & 0.2606 & 0.1566 & 0.2592 \\
0.80 & 0.1768 & 0.2609 & 0.1626 & 0.2613 & 0.1594 & 0.2604 \\
0.75 & 0.1774 & 0.2625 & 0.1628 & 0.2620 & 0.1617 & 0.2623 \\
0.70 & 0.1823 & 0.2633 & 0.1698 & 0.2637 & 0.1678 & 0.2639 \\
\bottomrule
\end{tabular}
}
\vspace{-10pt}
\end{table}

\subsection{Visualization of Mitigation Results}
\label{Visualization}
We present qualitative comparisons among different mitigation methods in Fig.~\ref{Fig32}. As shown, methods such as PEA, DOG, and RTA often introduce substantial semantic deviations, generating irrelevant objects or additional elements inconsistent with the original prompt. In contrast, latent-optimization-based methods, such as IND and SALI, can mitigate memorization to some extent, but typically at the cost of noticeably degraded image quality. Their generated images often exhibit missing fine-grained texture details, overall blurriness, or visible artifacts, thereby compromising perceptual quality. In addition, baselines such as PEA, CALS, and SALI sometimes show only limited effectiveness in mitigating memorization: they achieve only marginal suppression. Compared with these methods, our approach more effectively mitigates memorization while better preserving semantic consistency with the prompt and structural details in the generated images.

\subsection{Runtime Comparison}
\label{overhead}
Beyond mitigation effectiveness, computational efficiency is also important for inference-time deployment. We therefore compare the runtime of different memorization mitigation methods on 500 samples using SD v1.4, as reported in Tab.~\ref{mitigation_time}. Our method requires only $1115$ seconds, achieving the second-fastest runtime among all methods, just $25$ seconds slower than CALS. Despite this minor overhead, our method achieves better memorization mitigation and semantic fidelity than CALS, demonstrating a favorable balance between effectiveness and computational efficiency.

\begin{table}[h]
\centering
\caption{Runtime comparison of memorization mitigation methods on SD v1.4 for 500 samples. Our method achieves the second-fastest runtime, with only a minor overhead compared with CALS.}
\label{mitigation_time}
{
\setlength{\tabcolsep}{3pt}
\begin{tabular}{@{}c c c c c c c c@{}}
\toprule
Method
& RTA
& IND
& DOG
& PEA
& CALS
& SALI
& \cellcolor[gray]{0.9}\textbf{Ours} \\
\midrule
Time (s) $\downarrow$
& 2765
& 2385
& 5420
& 1490
& \textbf{1090}
& 2075
& \cellcolor[gray]{0.9}1115 \\
\bottomrule
\end{tabular}
}
\end{table}

%% file: sec/6_con.tex
\section{Conclusion}
In this paper, we investigate memorization in diffusion models from the perspective of cross-attention. We find that memorization is closely associated with a small number of semantically dominant attention heads in shallow layers. Based on this observation, we propose Projection-Based Top-$k$ Attention Head Suppression, which selectively suppresses condition-specific components in high semantic-control heads. Extensive experiments across diverse settings show that our method effectively mitigates memorization while preserving semantic consistency and image quality, achieving a better trade-off than existing approaches.


\section{LLM Usage Statement}

During the preparation of this manuscript, we used a large language model (LLM) solely for language polishing and improving overall clarity and readability. The LLM was not used to generate scientific ideas, conduct analyses, design experiments, or draw conclusions. The authors take full responsibility for the originality, accuracy, and integrity of all content presented in this work.

%% file: sec/X_suppl.tex
\clearpage
\setcounter{page}{1}
\maketitlesupplementary
\appendix

\section{More Related Works}
\paragraph{Detecting Memorization in Diffusion Models.}
The detection of memorization has attracted increasing attention. \citet{somepalli2023diffusion} used the SSCD metric, which is based on feature extractors, to quantify the degree of memorization by measuring the similarity between generated and training samples. However, this approach requires searching over the training dataset and performing pairwise comparisons with individual images, making it impractical for real-time evaluation during inference. To address this limitation, several real-time memorization detection methods have been proposed. \citet{wen2024detecting} identified memorized samples based on the observation that prompts associated with memorization tend to induce a larger discrepancy between unconditional and conditional noise predictions. \citet{chen2024exploring} further improved this idea by leveraging attention on the end token as a mask. Similarly, \citet{ren2024unveiling} detected memorization by computing the entropy of cross-attention distributions. In addition, \citet{jeon2025understanding} analyzed memorization from the perspective of the sharpness of the log-probability density and used this signal to identify memorized samples in diffusion models.

\section{More Experiments}

\subsection{Mitigation on Fine-tuned Models}
\label{app:finetuned_coco}

\begin{figure}[h]
  \centering
  \includegraphics[width=\linewidth]{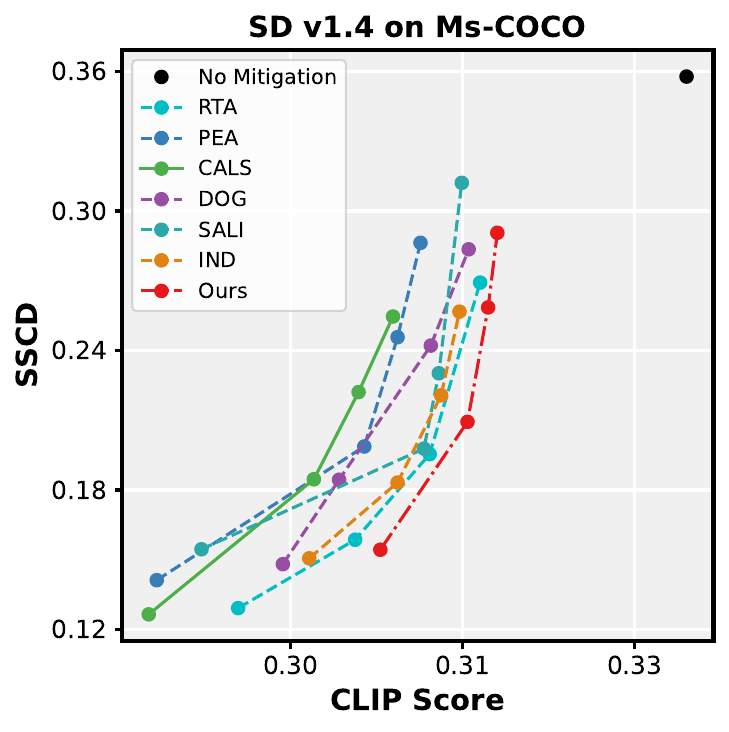}
  \caption{SSCD-CLIP trade-off on MS-COCO fine-tuned SD v1.4. Our method achieves a better balance between memorization mitigation and semantic alignment than baselines.}
  \label{sd1_coco}
\end{figure}

We further examine whether our method remains effective under a fine-tuning scenario, where memorization can be more directly induced by a limited training set. Specifically, we fine-tune SD v1.4 on 2,500 image-text pairs from MS-COCO and evaluate memorization mitigation using the corresponding text prompts and training images. The model is fine-tuned for 500 epochs with a learning rate of $1\times10^{-4}$; the detailed training script follows the Hugging Face Diffusers implementation \cite{finetune}. In Fig.~\ref{sd1_coco}, our method consistently achieves a more favorable SSCD-CLIP trade-off. Under comparable CLIP scores, our approach obtains a lower SSCD score than the baselines, while at similar SSCD levels, it better preserves image-text alignment. This result demonstrates that the memorization-related attention heads identified by our semantic-control criterion are not specific to pretrained memorized prompts, but also emerge in fine-tuned models. Therefore, the results on the fine-tuned model further validate the robustness of our method.

\subsection{Detecting Memorization}
\label{sec:detecting_memorization}

Our analysis in Secs.~\ref{motivation} and ~\ref{head} shows that memorized samples tend to activate a small number of shallow-layer attention heads with strong semantic control. To further verify this observation, we examine whether these heads can also serve as an indicator for memorization detection. Given the semantic-control score $S_{l,h}$ defined in Eq.~(\ref{cos}), we first select the Top-$k$ attention heads with the largest scores according to Eq.~(\ref{top-k}). Since the layer-wise analysis in Fig.~\ref{16_layers} shows that memorization-related discrepancies are most pronounced in shallow layers, especially layers $0$ and $3$, we use these two layers for detection. The memorization detection score is defined as the sum of the semantic-control scores over the selected Top-$k$ heads:
\begin{equation}
\mathcal{M}
=
\sum_{l \in \{0,3\}}
\sum_{h \in \mathcal{I}_{l}^{(k)}}
S_{l,h}.
\end{equation}
A larger value of $\mathcal{M}$ indicates that the sample is more likely to be memorized. For evaluation, following the memorized prompt setting in Sec.~\ref{setup}, we additionally use 500 non-memorized prompts sourced from MS-COCO~\cite{lin2014microsoft}, Lexica~\cite{lexica2024dataset}, and GPT-4~\cite{achiam2023gpt}, where each prompt is used to generate one image. We report the Area Under the ROC Curve (AUC) and the true positive rate at a fixed false positive rate of $1\%$ (T@F=$1\%$), where higher values indicate better detection performance. 

Tab.~\ref{Detecting} reports the detection results on SD v1.4 and SD v2.0. We compare our method with existing memorization detection methods from~\cite{wen2024detecting, ren2024unveiling, jeon2025understanding}. Although our method is not specifically optimized as a dedicated detector, it achieves competitive performance compared with these methods. The detection results provide additional evidence for our main observation: memorization is closely associated with a small subset of shallow-layer attention heads with strong semantic control.

\begin{table}[t]
\centering
\caption{Memorization detection results on SD v1.4 and SD v2.0. The competitive performance further supports our finding that memorization is associated with shallow-layer attention heads exhibiting large conditional--unconditional attention discrepancies.}
\label{Detecting}
{
\setlength{\tabcolsep}{3pt}
\begin{tabular}{@{}lcccc@{}}
\toprule
\multirow{2}{*}{Method}
& \multicolumn{2}{c}{SD v1.4}
& \multicolumn{2}{c}{SD v2.0} \\
\cmidrule(lr){2-3}
\cmidrule(lr){4-5}
& AUC $\uparrow$
& T@F=1\% $\uparrow$
& AUC $\uparrow$
& T@F=1\% $\uparrow$ \\
\midrule
\citeauthor{wen2024detecting}
& 94.32 & 34.32 & 91.43 & 20.63 \\
\citeauthor{ren2024unveiling}
& 95.67 & 42.66 & 84.82 & 10.50 \\
\citeauthor{jeon2025understanding}
& \textbf{99.01} & \textbf{82.80}
& \textbf{96.88} & 50.20 \\
\rowcolor[gray]{0.9}
Ours
& 98.41 & 78.12 & 96.23 & \textbf{54.21} \\
\bottomrule
\end{tabular}
}
\end{table}

\begin{figure*}[h!]
\centering
\setlength{\tabcolsep}{2pt}

\begin{tabular}{cccccc}
\rotatebox{90}{Layer 0} &
\includegraphics[width=0.18\linewidth]{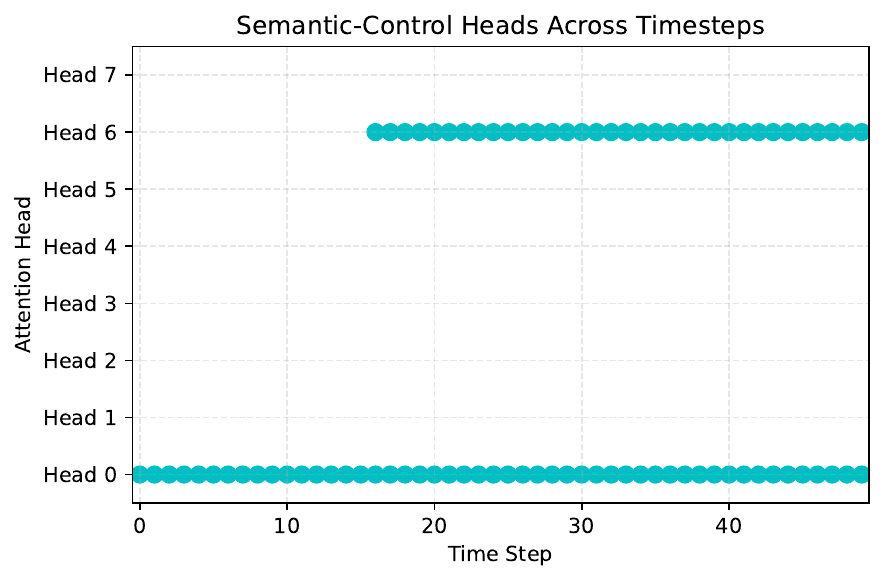} &
\includegraphics[width=0.18\linewidth]{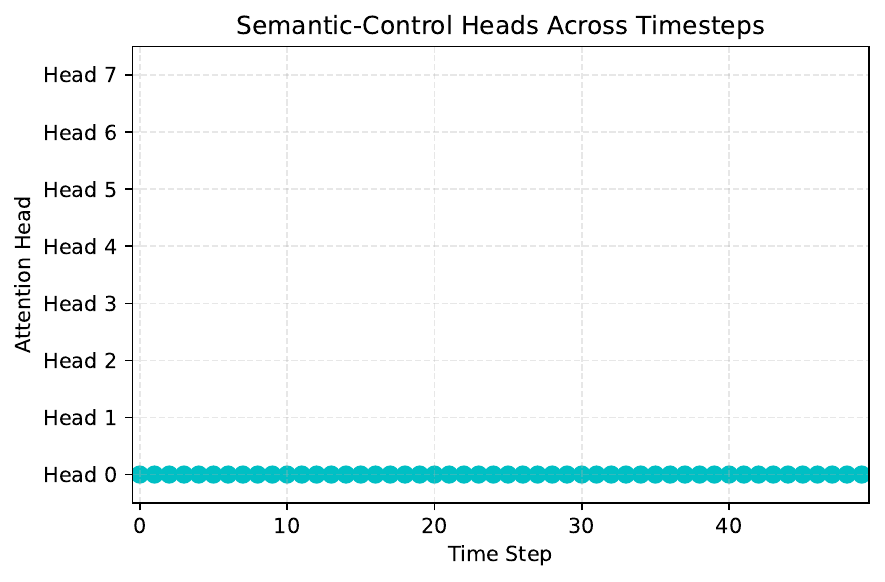} &
\includegraphics[width=0.18\linewidth]{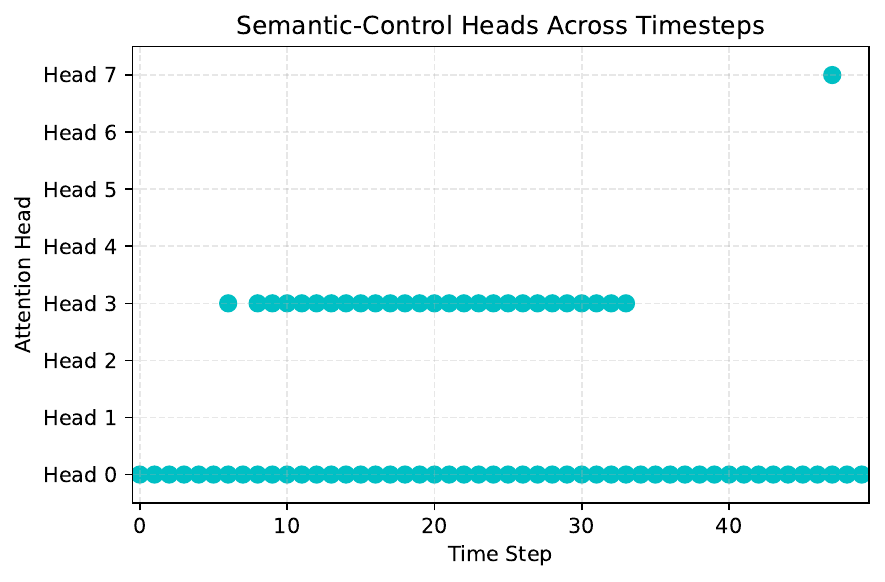} &
\includegraphics[width=0.18\linewidth]{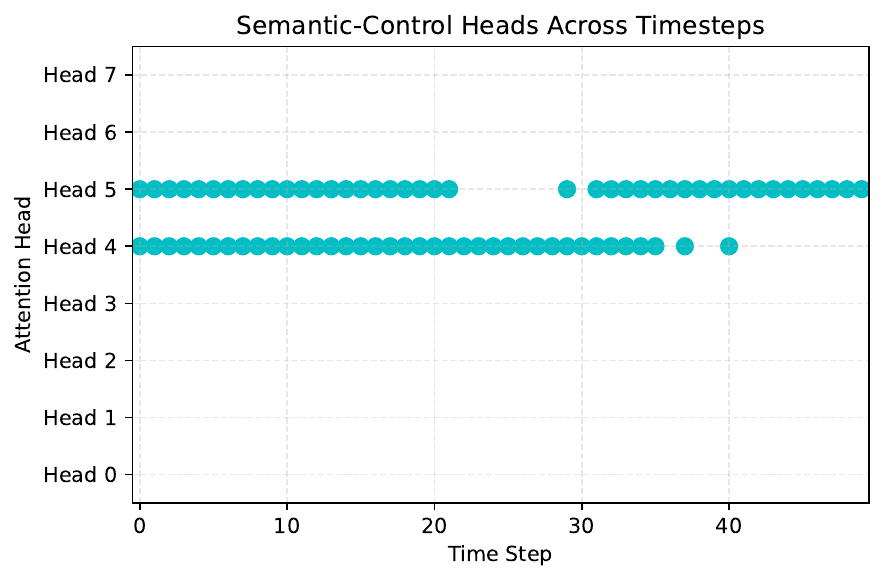} &
\includegraphics[width=0.18\linewidth]{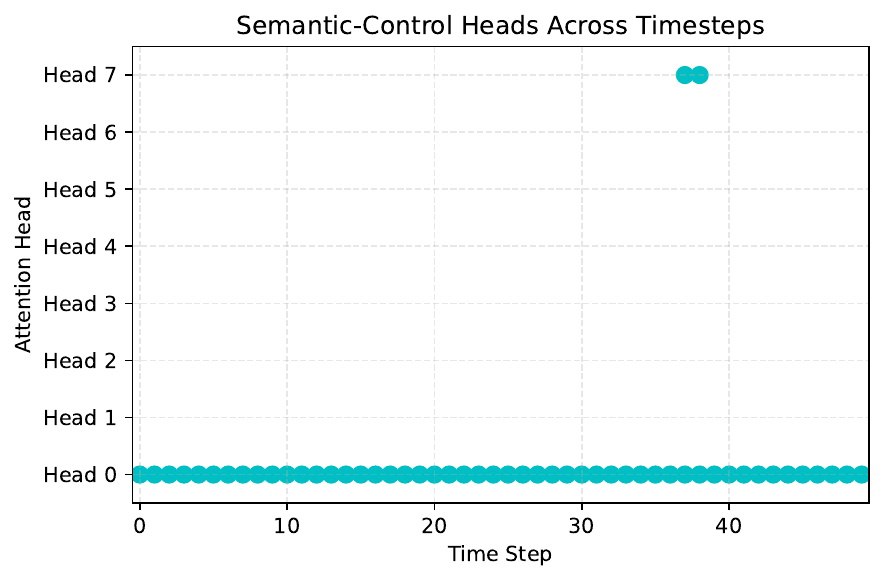}
\\

\rotatebox{90}{Layer 1} &
\includegraphics[width=0.18\linewidth]{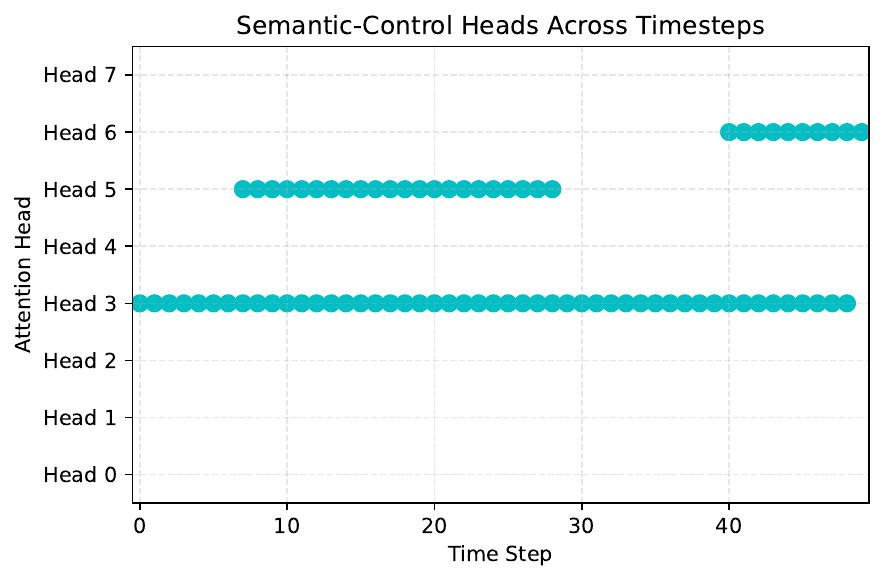} &
\includegraphics[width=0.18\linewidth]{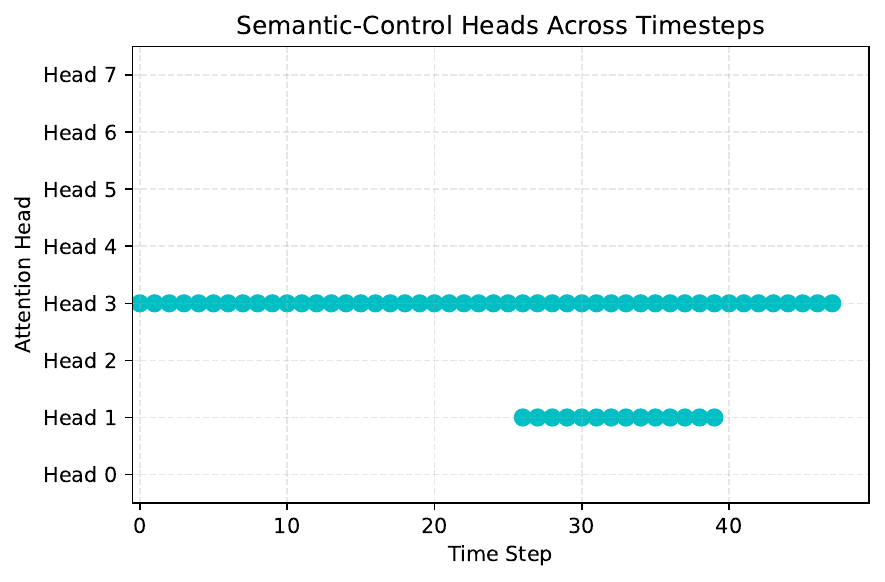} &
\includegraphics[width=0.18\linewidth]{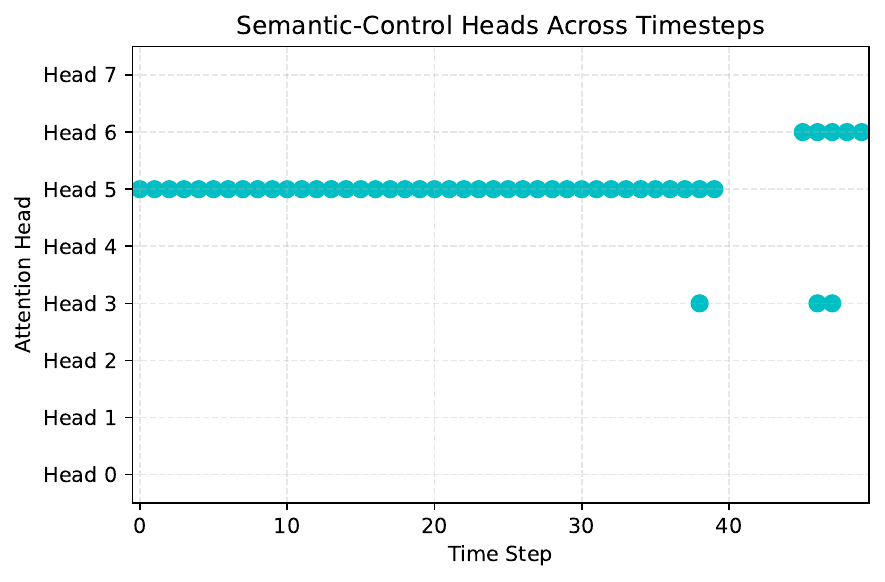} &
\includegraphics[width=0.18\linewidth]{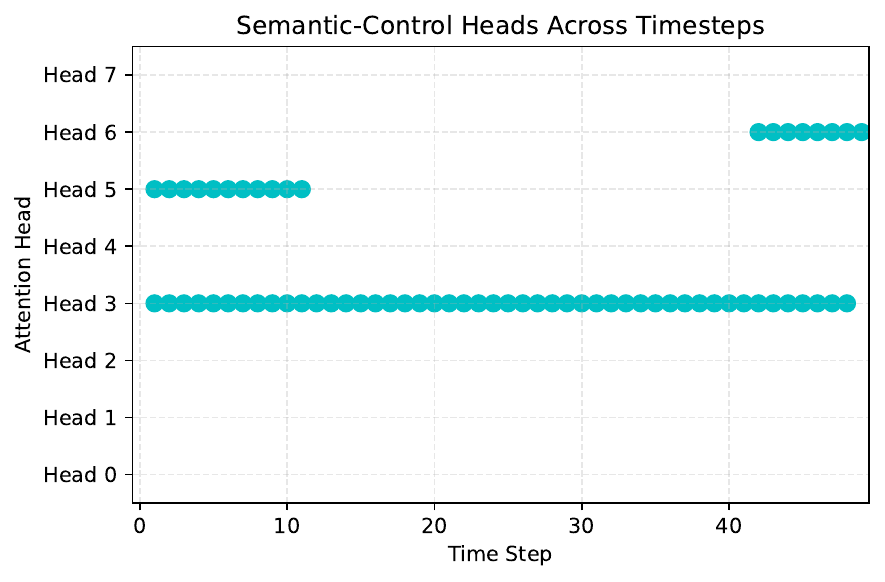} &
\includegraphics[width=0.18\linewidth]{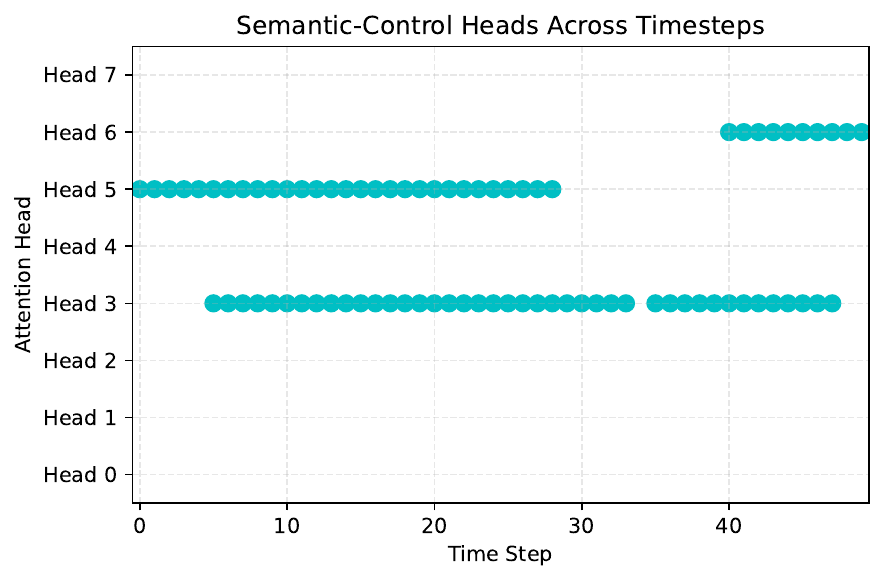}
\\

\rotatebox{90}{Layer 2} &
\includegraphics[width=0.18\linewidth]{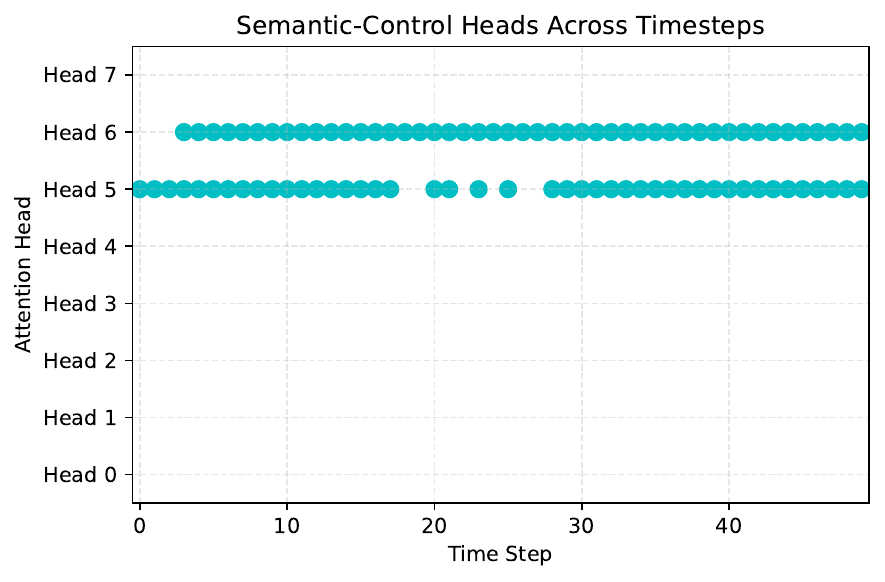} &
\includegraphics[width=0.18\linewidth]{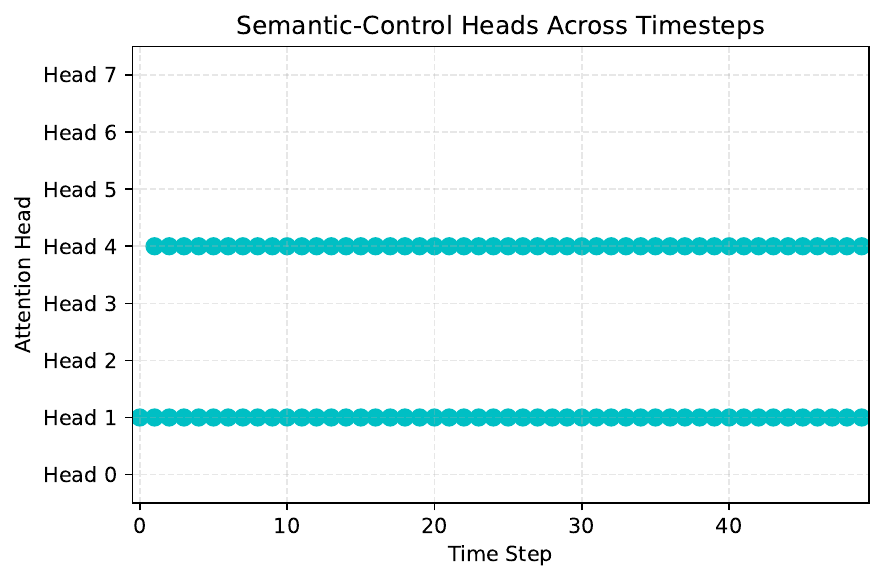} &
\includegraphics[width=0.18\linewidth]{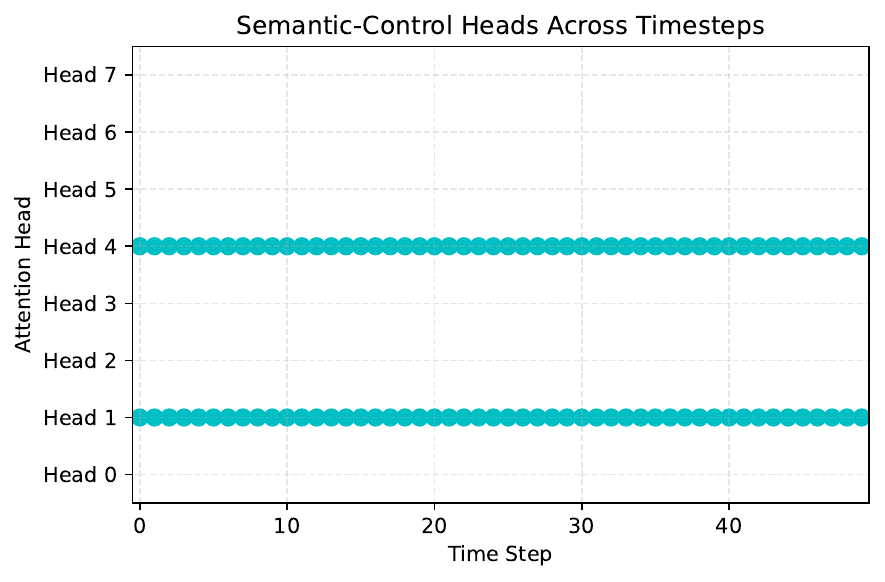} &
\includegraphics[width=0.18\linewidth]{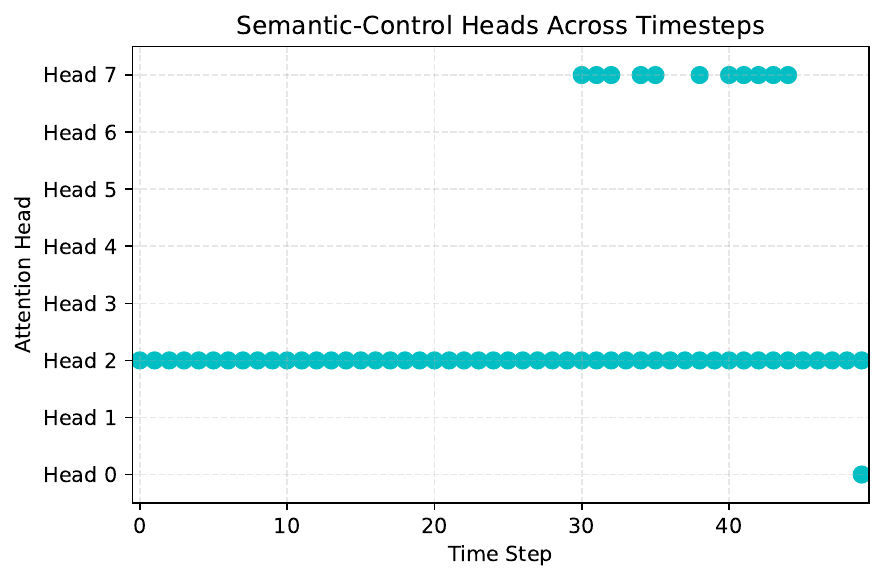} &
\includegraphics[width=0.18\linewidth]{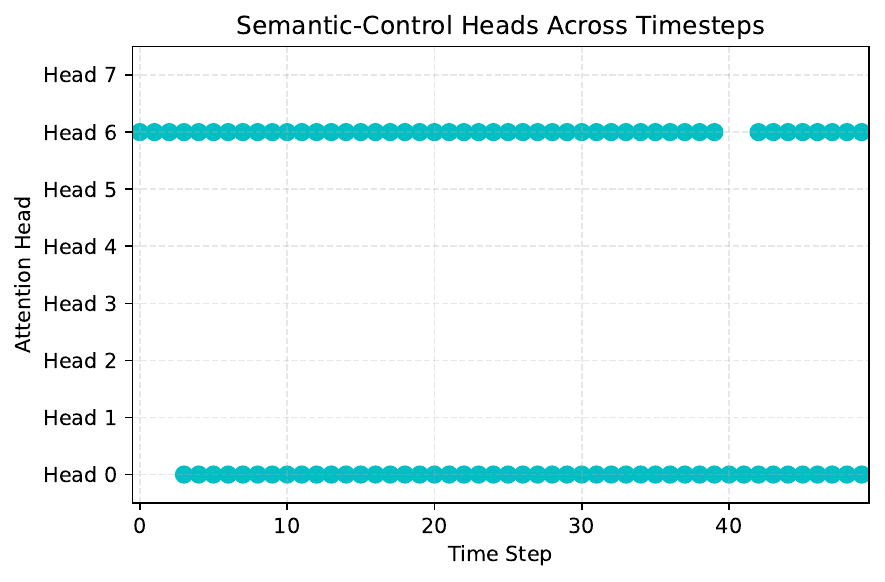}
\\

\rotatebox{90}{Layer 3} &
\includegraphics[width=0.18\linewidth]{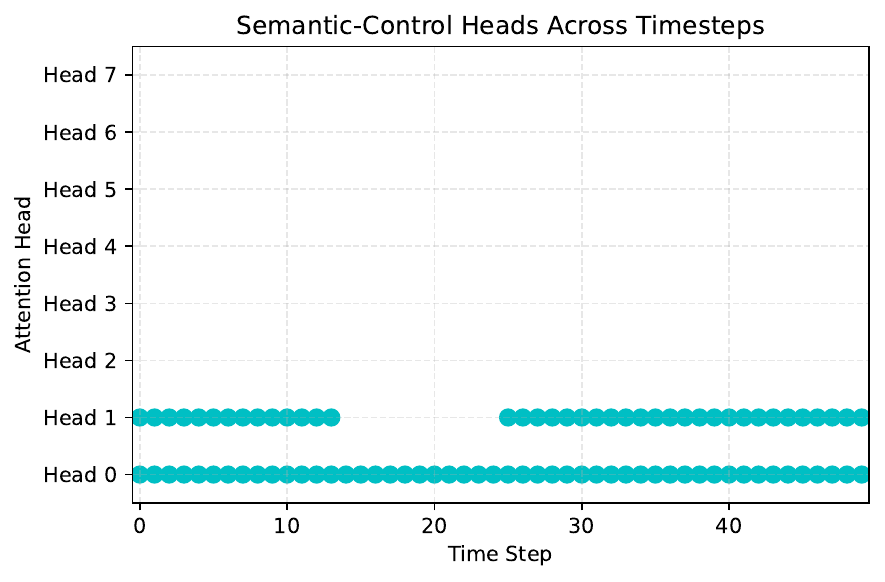} &
\includegraphics[width=0.18\linewidth]{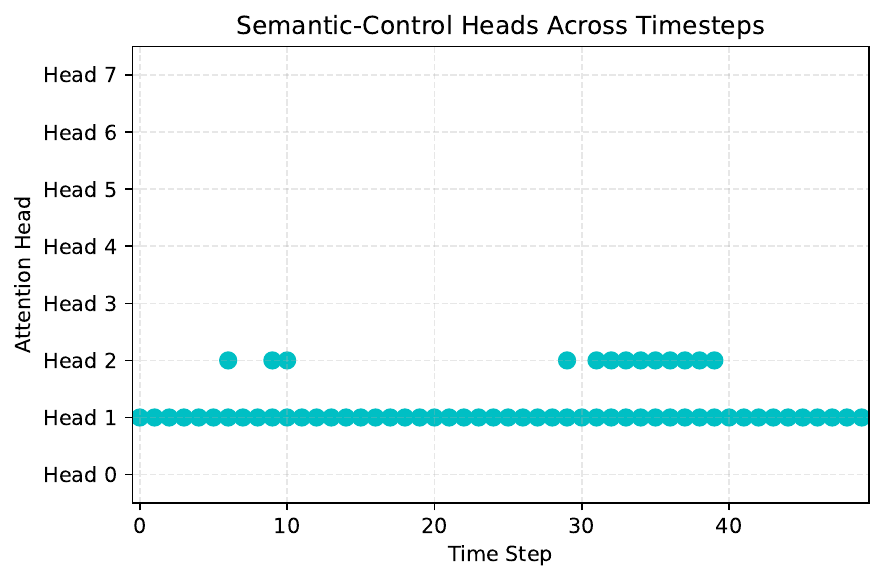} &
\includegraphics[width=0.18\linewidth]{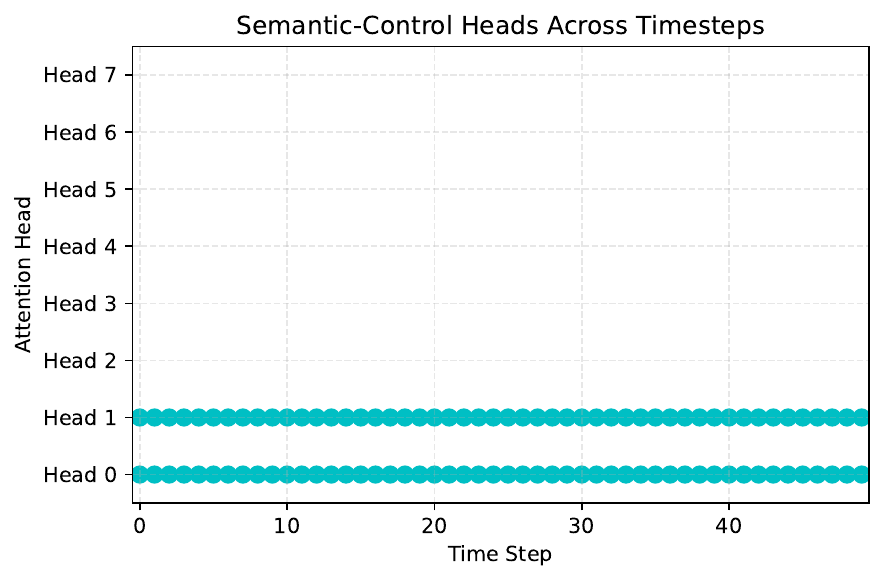} &
\includegraphics[width=0.18\linewidth]{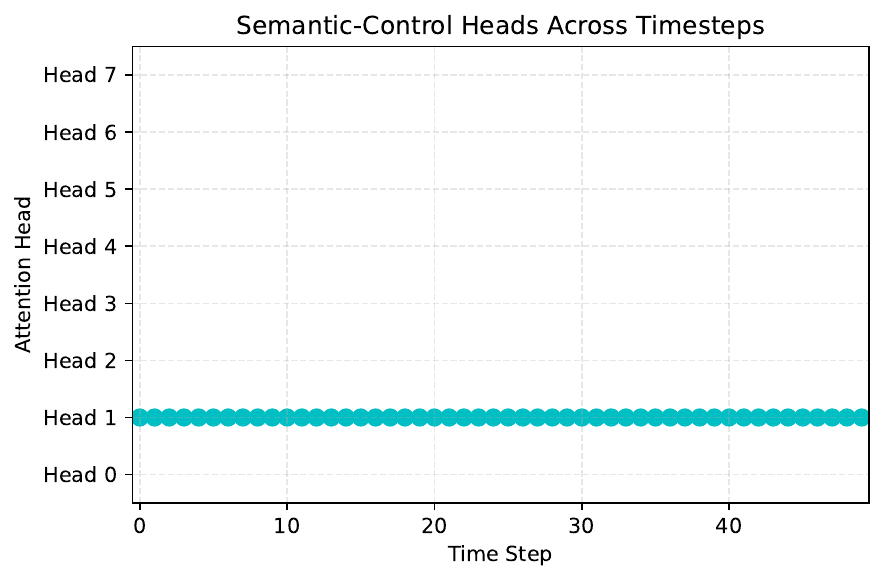} &
\includegraphics[width=0.18\linewidth]{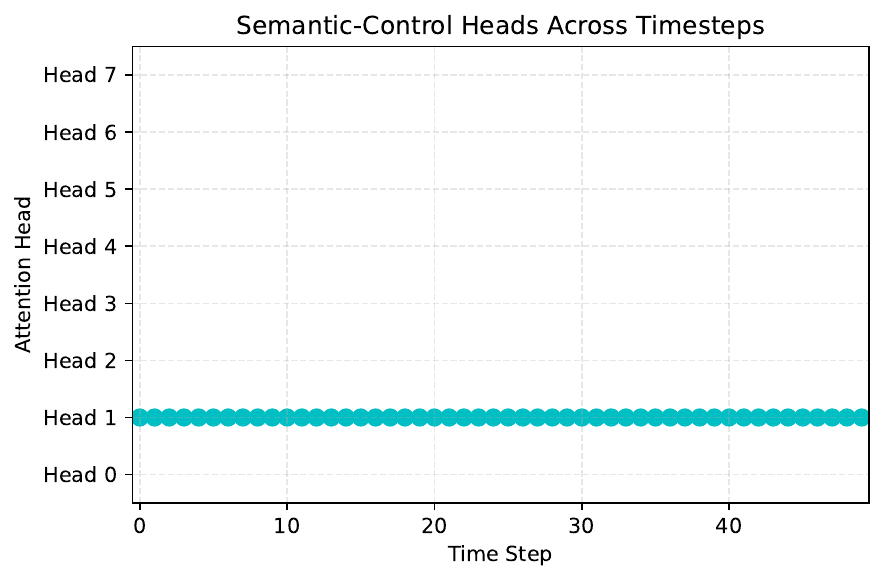}
\\

\end{tabular}

\caption{
Temporal distribution of outlier semantic-control attention heads across denoising timesteps.
Each row corresponds to a layer, and each column corresponds to one generated sample.
In each subplot, the $x$-axis denotes the denoising timestep, and the $y$-axis denotes the attention head index. Each cyan point indicates that the corresponding attention head is identified as an outlier semantic-control head at that timestep. The horizontal and temporally continuous patterns show that memorization-related semantic control is not random or transient, but is mainly carried by a stable subset of attention heads that persist across most denoising timesteps.
}
\label{fig:temporal_heads}
\label{timestep_head}
\end{figure*}

\subsection{Temporal Stability of Semantic Heads}
\label{timestep_stability}

We further examine whether these high semantic-control heads are stable across the denoising process. Specifically, at each denoising timestep, we identify the attention heads with abnormally strong semantic control according to the outlier criterion defined in Eq.~(\ref{outlier}). As shown in Fig.~\ref{timestep_head}, such outlier semantic-control heads persist across most denoising timesteps. This indicates that memorization-related semantic control is not a transient phenomenon that appears only at certain timesteps. Instead, it is mainly governed by a stable subset of attention heads.

This observation also supports the design of our mitigation strategy. Since the memorization-related heads remain largely unchanged across timesteps, it is unnecessary to re-identify the target heads at every denoising step. In practice, we compute the semantic-control scores only at the first denoising step, select the heads to be suppressed, and reuse them throughout the denoising process. This design reduces additional computational overhead while still targeting the attention heads that consistently contribute to memorization.

\begin{figure}[h]
  \centering
    \includegraphics[width=\linewidth]{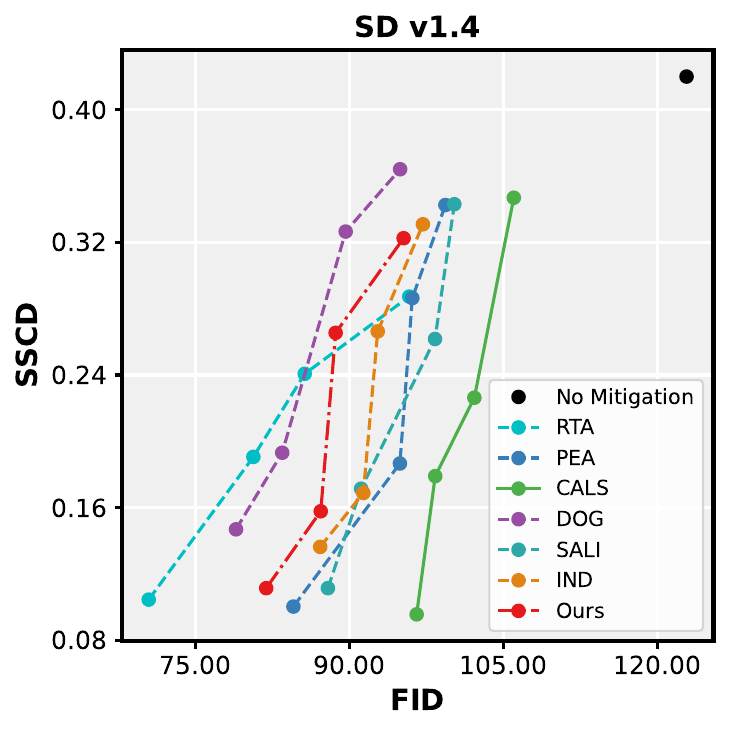} 
    \caption{SSCD--FID trade-off on SD v1.4. Lower FID indicates better generation quality.}
    \label{FID}
\end{figure}

\begin{figure}[h]
  \centering
    \includegraphics[width=\linewidth]{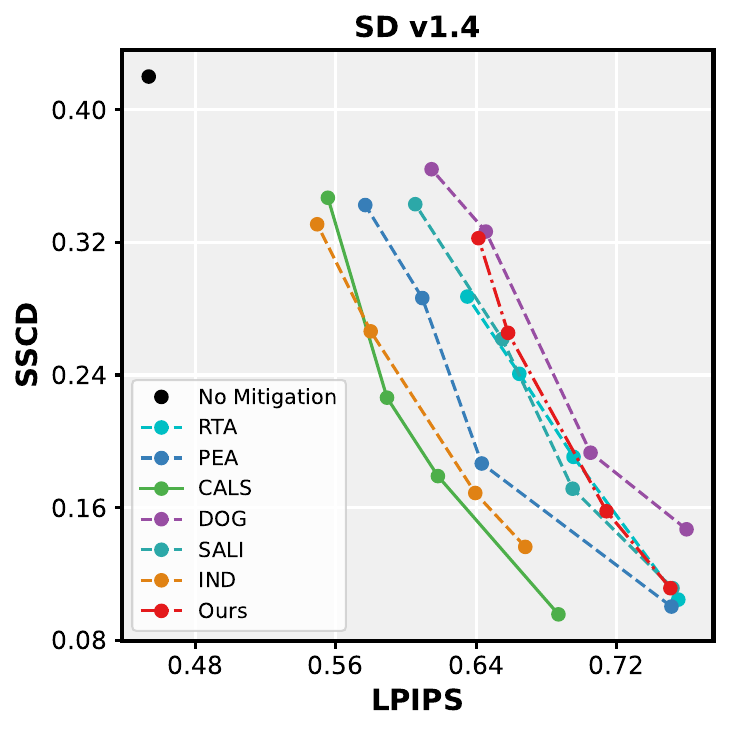}
    \caption{SSCD--LPIPS trade-off on SD v1.4. Higher LPIPS indicates better sample diversity.}
    \label{LPIPS}
\end{figure}

\subsection{FID and LPIPS}
\label{FID-LPIPS}
To further evaluate generation performance, we use Fréchet Inception Distance (FID)~\cite{heusel2017gans} to measure distribution fidelity and Learned Perceptual Image Patch Similarity (LPIPS)~\cite{zhang2018unreasonable} to assess image diversity. As shown in Fig.~\ref{FID} and Fig.~\ref{LPIPS}, our method maintains strong generation performance while effectively reducing memorization. Specifically, our method achieves the second-best result on LPIPS, indicating that it preserves high sample diversity, and the third-best result on FID, showing that it maintains competitive visual fidelity. Although RTA and DOG obtain better FID and LPIPS scores, their generated images exhibit poor text-image alignment, as shown in Sec.~\ref{main}. In contrast, our method achieves consistently favorable performance across different metrics, demonstrating that it can reduce memorization while preserving prompt alignment, image diversity, and visual quality.

\subsection{Effectiveness of Projection-Based Suppression}
We compare our projection-based suppression with naive suppression, which, as in the Attention Head Ablation section, directly suppresses the identified memorization-related attention heads by setting their attention outputs to zero. As shown in Tab.~\ref{tab:suppression_comparison}, our method achieves comparable SSCD scores while retaining higher CLIP scores, demonstrating better semantic preservation.

\begin{table}[t]
\centering
\caption{Comparison between naive suppression and our proposed suppression strategy on SD v1.4 and SD v2.0. Our method achieves comparable memorization mitigation while better preserving semantic fidelity on both models.}
\label{tab:suppression_comparison}
\setlength{\tabcolsep}{3pt}
\begin{tabular}{@{}lcccc@{}}
\toprule
\multirow{2}{*}{Method}
& \multicolumn{2}{c}{SD v1.4}
& \multicolumn{2}{c}{SD v2.0} \\
\cmidrule(lr){2-3}
\cmidrule(lr){4-5}
& SSCD ($\downarrow$) & CLIP ($\uparrow$)
& SSCD ($\downarrow$) & CLIP ($\uparrow$) \\
\midrule
Naive
& \textbf{0.1581} & 0.2482
& \textbf{0.2667} & 0.2613 \\
Ours
& 0.1578 & \textbf{0.2590}
& 0.2662 & \textbf{0.2730} \\
\rowcolor[gray]{0.9}
Gain
& -\;0.0003 & +\;0.0108
& -\;0.0005 & +\;0.0117 \\
\bottomrule
\end{tabular}
\end{table}

\subsection{Impact on Non-Memorized Generation}
Given the availability of several effective methods for detecting memorized samples~\cite{wen2024detecting, ren2024unveiling, jeon2025understanding}, existing studies typically assume that memorized and non-memorized samples can be identified beforehand, allowing mitigation to be selectively applied during inference. Nevertheless, it remains important to examine whether our method adversely affects generation quality on non-memorized inputs. To this end, we select 500 prompts from MS-COCO as non-memorized samples and compare images generated by the original models with those generated using our method. We evaluate both SD v1.4 and SD v2.0 using the same set of prompts, with the results reported in Tab.~\ref{non-memorized}. 

Our method produces nearly identical generation quality to the original models. On SD v1.4, the CLIP score slightly increases from $0.3386$ to $0.3390$, while the FID changes only marginally from $85.47$ to $85.49$. On SD v2.0, the CLIP score changes from $0.3509$ to $0.3503$, while the FID improves from $86.73$ to $86.20$. These negligible differences indicate that our intervention does not broadly weaken normal prompt conditioning. Instead, the results further support that our method primarily suppresses the abnormally strong semantic control associated with memorized
generation while preserving ordinary semantic information.

\begin{table}[h]
\centering
\caption{Generation quality on non-memorized prompts. Our method preserves semantic alignment and distributional quality on non-memorized inputs.}
\label{non-memorized}
{
\setlength{\tabcolsep}{5pt}
\begin{tabular}{@{}lcccc@{}}
\toprule
\multirow{2}{*}{Method}
& \multicolumn{2}{c}{SD v1.4}
& \multicolumn{2}{c}{SD v2.0} \\
\cmidrule(lr){2-3}
\cmidrule(lr){4-5}
& CLIP ($\uparrow$) & FID ($\downarrow$)
& CLIP ($\uparrow$) & FID ($\downarrow$) \\
\midrule
No Mitigation
& 0.3386 & 85.47 & 0.3509 & 86.73 \\
\rowcolor[gray]{0.9}
Ours
& 0.3390 & 85.49 & 0.3503 & 86.20 \\
\bottomrule
\end{tabular}
}
\end{table}

\subsection{Evaluation on DiT Architectures}
\label{SD3}

We additionally evaluate our method on Stable
Diffusion 3 (SD 3)~\cite{esser2024scaling} to examine whether its empirical effectiveness extends to a more recent diffusion architecture. Unlike SD v1.4 and SD v2.0, which employ U-Net-based denoising networks, SD 3 adopts a transformer-based backbone with a substantially different attention architecture. Specifically, within each joint-attention module of SD 3, we isolate the text-to-image component, where image queries attend to textual keys
and values, and apply our conditional-unconditional head selection and directional suppression to this component.

We follow the prompt selection protocol in~\cite{zhao2026you}. Specifically, we begin with the same 500 memorized prompts originally constructed for earlier Stable Diffusion models and retain prompts whose generations from the unmitigated SD 3 model achieve an SSCD score greater than $0.30$. Since SD 3 exhibits substantially weaker memorization on this benchmark, only 32 prompts satisfy this criterion. We therefore conduct the evaluation on this subset and use the same prompts and inference settings for all applicable methods. For the baselines, we adapt RTA, PEA, IND, and DOG from their original U-Net-based implementations to the SD 3 architecture. The remaining baselines are not included, as we have not yet found a reliable way to adapt them to SD 3.

As shown in Tab.~\ref{sd3}, our method achieves the lowest SSCD score of 0.2221 while maintaining the highest CLIP score of 0.2532 among all mitigation methods. These results demonstrate that our method remains effective on the DiT-based architecture while preserving prompt alignment. However, given the limited number of memorized samples available for SD 3, these results primarily demonstrate the applicability of our method rather than providing a comprehensive understanding of memorization in DiT-based architectures. A systematic investigation of memorization in DiT-based architectures is left for future work.

\begin{table}[h]
\centering
\caption{Memorization mitigation results on Stable Diffusion 3. Our method achieves the best balance between memorization mitigation and semantic alignment.}
\label{sd3}
\setlength{\tabcolsep}{5pt}
\begin{tabular}{@{}lcc@{}}
\toprule
\multirow{2}{*}{Method}
& \multicolumn{2}{c}{SD 3} \\
\cmidrule(lr){2-3}
& SSCD ($\downarrow$) & CLIP ($\uparrow$) \\
\midrule
No Mitigation & 0.3372 & 0.2625 \\
\midrule
RTA & 0.2252 & 0.2367 \\
PEA & 0.2292 & 0.2438 \\
IND & 0.2442 & 0.2451 \\
DOG & 0.2289 & 0.2219 \\
\rowcolor[gray]{0.9}
Ours & 0.2221 & 0.2532 \\

\bottomrule
\end{tabular}
\end{table}

\subsection{Joint Analysis with NeMo}

NeMo~\cite{hintersdorf2024finding} identifies prompt-specific memorization neurons in cross-attention value projections. To examine the consistency between neuron-level and head-level localization, we measure the proportion of NeMo-identified neurons covered by our selected attention heads.

Let $\mathcal{M}_{p,l}$ denote the memorization neurons identified by NeMo for prompt $p$ at layer $l$, and $\mathcal{I}_{p,l}^{(k)}$ denote the Top-$k$ heads selected by our method. Since value projection dimensions are partitioned among attention heads, neuron $j$ belongs to head $h(j)=\lfloor j/d_h \rfloor$, where $d_h$ is the head dimension. We define the average neuron coverage rate as:
\begin{equation}
\mathrm{NC}^{(k)}
=
\frac{1}{|\mathcal{P}|}
\sum_{p\in\mathcal{P}}
\frac{
\sum_l
\left|
\left\{
j\in\mathcal{M}_{p,l}
\mid
h(j)\in\mathcal{I}_{p,l}^{(k)}
\right\}
\right|
}{
\sum_l |\mathcal{M}_{p,l}|
},
\end{equation}
where $\mathcal{P}$ denotes the set of memorized prompts.

Our selected attention heads cover 61.40\% of the memorization neurons identified by NeMo, increasing to 77.14\% when restricting the comparison to Layers 0--3. These results suggest consistency between neuron-level and head-level localization, further supporting our observation that memorization-related signals are concentrated in a small subset of shallow-layer attention heads.

Beyond neuron-level localization, our analysis provides further insight into how memorization-related conditional influence is organized within the attention mechanism. This head-level perspective complements NeMo by revealing the structural distribution and functional role of memorization-related attention responses.

\begin{table*}[h]
\centering
\caption{Mitigation results on different memorization types. AM denotes all memorized samples, VM denotes verbatim memorization, and TM denotes template memorization. We report SSCD and CLIP scores under SD v1.4 and SD v2.0, with four hyperparameter settings for each method. Lower SSCD indicates better memorization mitigation, while higher CLIP indicates better prompt-image alignment.}
\label{M}
\setlength{\tabcolsep}{3pt}
\begin{tabular}{l|cc|cc|cc|cc}
\toprule
\multirow{2}{*}{Method} & \multicolumn{2}{c}{SDv1.4 – AM} & \multicolumn{2}{c}{SDv1.4 – VM} & \multicolumn{2}{c}{SDv1.4 – TM} & \multicolumn{2}{c}{SDv2.0 – TM} \\

& SSCD ($\downarrow$) & CLIP ($\uparrow$) & SSCD ($\downarrow$) & CLIP ($\uparrow$) & SSCD ($\downarrow$) & CLIP ($\uparrow$) & SSCD ($\downarrow$) & CLIP ($\uparrow$) \\
\midrule

No Mitigation
& 0.5122 & 0.3009 & 0.7895 & 0.3371 & 0.4322 & 0.2904 & 0.6306 & 0.2947 \\

\midrule
\multirow{4}{*}{RTA}
& 0.1045 & 0.1959 & 0.0888 & 0.2053 & 0.1090 & 0.1932 & 0.1583 & 0.2093 \\
& 0.1906 & 0.2394 & 0.1849 & 0.2552 & 0.1923 & 0.2348 & 0.2152 & 0.2463 \\
& 0.2407 & 0.2487 & 0.2660 & 0.2727 & 0.2334 & 0.2418 & 0.2866 & 0.2633 \\
& 0.2873 & 0.2545 & 0.3173 & 0.2516 & 0.2786 & 0.2553 & 0.4547 & 0.2824 \\
\midrule

\multirow{4}{*}{PEA}
& 0.1003 & 0.2136 & 0.0751 & 0.2270 & 0.1075 & 0.2097 & 0.1617 & 0.2050 \\
& 0.1866 & 0.2532 & 0.1704 & 0.2761 & 0.1912 & 0.2466 & 0.2421 & 0.2426 \\
& 0.2864 & 0.2704 & 0.3023 & 0.2891 & 0.2818 & 0.2650 & 0.2852 & 0.2513 \\
& 0.3435 & 0.2765 & 0.3748 & 0.2972 & 0.3344 & 0.2705 & 0.3723 & 0.2731 \\
\midrule

\multirow{4}{*}{CALS}
& 0.0956 & 0.2207 & 0.0751 & 0.2358 & 0.1015 & 0.2163 & 0.1268 & 0.1992 \\
& 0.1790 & 0.2471 & 0.1953 & 0.2652 & 0.1742 & 0.2418 & 0.2371 & 0.2527 \\
& 0.2263 & 0.2489 & 0.2461 & 0.2671 & 0.2206 & 0.2436 & 0.4121 & 0.2683 \\
& 0.3469 & 0.2702 & 0.4013 & 0.3063 & 0.3311 & 0.2598 & 0.4582 & 0.2719 \\

\midrule

\multirow{4}{*}{DOG}
& 0.1469 & 0.2076 & 0.1631 & 0.2274 & 0.1422 & 0.2019 & 0.2387 & 0.1928 \\
& 0.1931 & 0.2322 & 0.2104 & 0.2500 & 0.1881 & 0.2270 & 0.2906 & 0.2109 \\
& 0.3265 & 0.2696 & 0.3842 & 0.2820 & 0.3098 & 0.2660 & 0.2095 & 0.1839 \\
& 0.3641 & 0.2769 & 0.4311 & 0.2903 & 0.3447 & 0.2730 & 0.4262 & 0.2539 \\

\midrule

\multirow{4}{*}{SALI}
& 0.1114 & 0.2130 & 0.1135 & 0.2339 & 0.1107 & 0.2070 & 0.1414 & 0.2130 \\
& 0.1714 & 0.2445 & 0.1760 & 0.2693 & 0.1701 & 0.2373 & 0.2917 & 0.2656 \\
& 0.2617 & 0.2656 & 0.2753 & 0.2884 & 0.2578 & 0.2590 & 0.4730 & 0.2918 \\
& 0.3430 & 0.2818 & 0.3646 & 0.3080 & 0.3367 & 0.2742 & 0.2014 & 0.2445 \\
\midrule

\multirow{4}{*}{IND}
& 0.1363 & 0.2346 & 0.1121 & 0.2521 & 0.1432 & 0.2295 & 0.1472 & 0.1701 \\
& 0.1688 & 0.2459 & 0.1531 & 0.2545 & 0.1733 & 0.2434 & 0.2443 & 0.2148 \\
& 0.2663 & 0.2747 & 0.2525 & 0.2794 & 0.2702 & 0.2733 & 0.2764 & 0.2300 \\
& 0.3309 & 0.2849 & 0.3522 & 0.2967 & 0.3247 & 0.2815 & 0.4740 & 0.2676 \\

\midrule

\multirow{4}{*}{Ours}
& 0.1114 & 0.2433 & 0.1164 & 0.2567 & 0.1099 & 0.2394 & 0.1090 & 0.2041 \\
& 0.1578 & 0.2590 & 0.1983 & 0.2897 & 0.1461 & 0.2501 & 0.1641 & 0.2360 \\
& 0.2654 & 0.2788 & 0.3054 & 0.3049 & 0.2538 & 0.2712 & 0.2662 & 0.2730 \\
& 0.3225 & 0.2882 & 0.3793 & 0.3179 & 0.3061 & 0.2796 & 0.3848 & 0.2892 \\
\bottomrule
\end{tabular}
\end{table*}

\subsection{Different Memorization Types}
Since verbatim memorization (VM) and template memorization (TM) reflect different reproduction behaviors, we further evaluate our method on these two memorization types, following studies~\cite{ren2024unveiling, chen2024exploring}. As shown in Tab.~\ref{M}, our method achieves strong mitigation performance on both VM and TM, demonstrating its effectiveness in suppressing not only near-verbatim training-image reproduction but also template-level copying. Meanwhile, the competitive CLIP scores indicate that our method preserves prompt alignment while substantially reducing memorization.

\section{Discussions}
\paragraph{Why Do Shallow and Deep Attention Heads Behave Differently?} Our layer-wise analysis shows that memorized samples exhibit a substantially larger number of high semantic-control heads in shallow
cross-attention layers, whereas the distributions of memorized and non-memorized samples become increasingly similar in deeper layers. We hypothesize that shallow cross-attention layers act as semantic routing bottlenecks, where textual conditions first establish the global subject, layout, and generation trajectory. For memorized prompts, this early semantic injection may become highly concentrated in a small number of attention heads, effectively steering the generation toward a specific training example or visual template. Once this trajectory has been established, deeper layers may mainly integrate and refine the resulting visual representation. Through
successive residual, self-attention, and cross-attention operations, the initially concentrated conditional signal may also become distributed across multiple features, making individual deep heads
less distinguishable between memorized and non-memorized samples. Furthermore, perturbations introduced in shallow layers propagate through a longer downstream computation path and may therefore exert
a larger influence on the final output. 

\paragraph{Why Does Projection-Based Suppression Work?} Our analysis reveals that memorization is associated with abnormally strong conditional control concentrated in a small number of shallow attention heads. Therefore, it is sufficient to project out and suppress the corresponding conditional activation directions, thereby weakening their excessive control over generation. This targeted intervention mitigates memorization without completely removing the attention outputs, allowing the model to preserve semantic information necessary for faithful generation.

\paragraph{Why Detect on Attention Maps but Suppress Hidden Representations?}
For the $h$-th cross-attention head, the attention map and its output are:
\begin{equation}
A_h=\operatorname{softmax}\left(\frac{Q_hK_h^\top}{\sqrt{d}}\right),
\qquad
H_h=A_hV_h.
\end{equation}
The attention map $A_h$ explicitly characterizes how textual conditions route information to spatial queries, making the conditional--unconditional discrepancy in $A_h$ a direct signal for identifying heads with abnormal condition-dependent routing. In contrast, $H_h$ is the resulting representation that is actually propagated to subsequent layers, and is therefore a more appropriate target for intervention. The connection between the two can be seen from:
\begin{equation}
\begin{aligned}
H_h^{c}-H_h^{u}
&=A_h^{c}V_h^{c}-A_h^{u}V_h^{u}\\
&=(A_h^{c}-A_h^{u})V_h^{u}
+A_h^{c}(V_h^{c}-V_h^{u}).
\end{aligned}
\end{equation}
Thus, the routing discrepancy in the attention map constitutes one source of the downstream representation discrepancy, while the latter also captures condition-dependent changes in value content. Directly modifying $A_h$ changes attention routing, which may affect both memorization-related and normal semantic interactions. We therefore retain the original attention routing and intervene on $H_h$, where the representation can be selectively adjusted along the condition-specific direction.

\section{Memorized Prompts}

The following memorized prompts are used in our qualitative comparisons of different memorization mitigation methods. These prompts correspond to the examples presented in Sec.~\ref{Visualization}:
\begin{itemize}
    \item Kitchen painted in SHIMMERING GREEN.
    \item Mothers influence on her young hippo.
    \item Foyer painted in WHITE.
    \item Donna Tartt's The Goldfinch Scores Film Adaptation.
\end{itemize}


\section{Details for Metrics}

We use the \texttt{sscd\_disc\_large} model from the official SSCD copy-detection repository\footnote{\url{https://github.com/facebookresearch/sscd-copy-detection}} to compute image-level memorization similarity. Prompt-image alignment is evaluated using the CLIP score with the ViT-B-32 backbone CLIP model\footnote{\url{https://huggingface.co/laion/CLIP-ViT-B-32-laion2B-s34B-b79K}}. To assess distributional fidelity, we compute FID by extracting image features with an Inception-v3 network. Finally, we measure perceptual diversity using LPIPS with an AlexNet backbone.

\section{Details for Baseline Methods}
\label{sec:baselines}

\paragraph{Random Token Addition.}
\citet{somepalli2023understanding} showed that memorization is closely
related to image-caption associations. They mitigate memorization
by randomizing the text condition. During inference, the prompt is perturbed by inserting a random token $r$:
\begin{equation}
p' = \operatorname{Insert}(p,r),
\end{equation}
where $r$ is randomly sampled from the vocabulary. After multiple random-token insertions, the final perturbed prompt $p'$ is used for generation.

\paragraph{Prompt Embedding Adjustment.}
\citet{wen2024detecting} observed that memorized prompts often induce a large discrepancy between conditional and unconditional noise predictions. Their inference-time mitigation optimizes the prompt embedding by minimizing this discrepancy:
\begin{equation}
\mathcal{L}_{\text{emb}}(\mathbf{e}^{*})
=
\left\|
\epsilon_{\theta}(\mathbf{x}_t,t,\mathbf{e}^{*})
-
\epsilon_{\theta}(\mathbf{x}_t,t,\mathbf{e}_{\emptyset})
\right\|_2 .
\label{eq:wen_loss}
\end{equation}
Here, $\mathbf{e}^{*}$ denotes the learnable prompt embedding initialized from the original prompt, $\mathbf{e}_{\emptyset}$ denotes the empty-prompt embedding. The optimized embedding $\mathbf{e}^{*}$ is then used for generation instead of the original prompt embedding.

\paragraph{Cross-Attention Scaling.}
\citet{ren2024unveiling} found that memorized samples often place excessive cross-attention on the trigger tokens. Let $\mathbf{s}=(s_1,\ldots,s_N)$ be the attention logits over $N$ text tokens, where $s_1$ is the beginning-token logit. The adjusted logits are:
\begin{equation}
\mathbf{s}' = (Cs_1,\ s_2,\ldots, s_N, \ -\infty,\ldots,-\infty),
\end{equation}
where $C>1$ enlarges the beginning-token logit. The cross-attention map is then computed by:
\begin{equation}
\mathbf{A}'= \operatorname{softmax}(\mathbf{s}').
\label{eq:ren_attn}
\end{equation}

\paragraph{Dynamic Opposite Guidance.}
\citet{jain2025classifier} interpret memorization as the effect of applying the CFG inside an attraction basin. They delay CFG until the denoising trajectory reaches a transition point. Let $\mathbf{e}_p$ denote the text embedding, the transition point is estimated by tracking:
\begin{equation}
d_t=\|\Delta\epsilon_t\| =
\|\epsilon_{\theta}(\mathbf{x}_t,t,\mathbf{e}_p)-\epsilon_{\theta}(\mathbf{x}_t,t,\mathbf{e}_{\emptyset})\|, 
\label{eq:jain_transition_metric}
\end{equation}
and selecting the first local minimum or a timestep after a sharp drop in $d_t$. The noise prediction is then modified as:
\begin{equation}
\hat{\epsilon}_{\theta}
=
\begin{cases}
\epsilon_{\theta}(\mathbf{x}_t,t,\mathbf{e}_{\emptyset}),
& t>\tau, \\
\epsilon_{\theta}(\mathbf{x}_t,t,\mathbf{e}_{\emptyset})
+
w\Delta\epsilon_t,
& t\leq\tau,
\end{cases}
\label{eq:transition_cfg}
\end{equation}
where $\tau$ is the transition point. To escape the attraction basin earlier, they further propose opposite guidance:
\begin{equation}
\hat{\epsilon}_{\theta}^{\text{OG}}
=
\epsilon_{\theta}(\mathbf{x}_t,t,\mathbf{e}_{\emptyset})
-
w_{\text{og}}\Delta\epsilon_t,
\label{eq:opposite_guidance}
\end{equation}
where $w_{\text{og}}$ denotes the strength of opposite guidance. After the trajectory escapes the basin, the sampler switches back to standard positive CFG.

\paragraph{Initial Noise Adjustment.}
Based on Dynamic Opposite Guidance \cite{jain2025classifier}, \citet{han2025adjusting} also modify the initial noise to directly reduce the conditional guidance magnitude at the initial timestep. They define:
\begin{equation}
\tilde{\epsilon}_{\theta}(\mathbf{x}_T,T,\mathbf{e}_p)
=
\epsilon_{\theta}(\mathbf{x}_T,T,\mathbf{e}_p)
-
\epsilon_{\theta}(\mathbf{x}_T,T,\mathbf{e}_{\emptyset}).
\label{eq:han_delta}
\end{equation}
A sharpness-inspired perturbation is computed as:
\begin{equation}
\hat{\delta}(\mathbf{x}_T)
=
\rho
\frac{
\tilde{\epsilon}_{\theta}(\mathbf{x}_T,T,\mathbf{e}_p)
}{
\left\|
\tilde{\epsilon}_{\theta}(\mathbf{x}_T,T,\mathbf{e}_p)
\right\|_2
},
\end{equation}
where $\rho$ controls the radius of the local perturbation used to estimate the sharpness direction.
The initial noise is then updated by:
\begin{equation}
\begin{split}
\tilde{\mathbf{x}}_T
=
\mathbf{x}_T
-
\tilde{\gamma}
\Big[
&
\tilde{\epsilon}_{\theta}
\bigl(
\mathbf{x}_T+\hat{\delta}(\mathbf{x}_T),
T,\mathbf{e}_p
\bigr) \\
&-
\tilde{\epsilon}_{\theta}
\bigl(
\mathbf{x}_T,T,\mathbf{e}_p
\bigr)
\Big],
\end{split}
\end{equation}
where $\tilde{\gamma}$ is the step size for updating the initial noise. This encourages the trajectory to leave the attraction basin earlier, allowing CFG to be applied with better prompt alignment.

\paragraph{Sharpness-Aware Latent Initialization.}
\citet{jeon2025understanding} interpreted memorization as sampling from
sharp regions of the learned probability landscape. Their method optimizes
the initial noise $\mathbf{x}_T$ toward a smoother region. They first define
the conditional score difference and the corresponding Hessian difference:
\begin{equation}
\begin{aligned}
\Delta s_{\theta}(\mathbf{x}_T)
&=
s_{\theta}(\mathbf{x}_T,\mathbf{e}_p)
-
s_{\theta}(\mathbf{x}_T,\mathbf{e}_{\emptyset}), \\
\Delta H_{\theta}(\mathbf{x}_T)
&=
H_{\theta}(\mathbf{x}_T,\mathbf{e}_p)
-
H_{\theta}(\mathbf{x}_T,\mathbf{e}_{\emptyset}).
\end{aligned}
\label{eq:sail_def}
\end{equation}
Here, $s_{\theta}(\cdot)$ denotes the score function, $H_{\theta}(\cdot)$
denotes the Hessian of the log probability. The ideal sharpness-aware objective is:
\begin{equation}
\begin{split}
\min_{\mathbf{x}_T}\quad
\left\|
\Delta H_{\theta}(\mathbf{x}_T)
\Delta s_{\theta}(\mathbf{x}_T)
\right\|_2^2 -\alpha \log p_G(\mathbf{x}_T),
\end{split}
\label{eq:sail_ideal}
\end{equation}
where $p_G(\mathbf{x}_T)$ is the Gaussian prior density of the initial noise,
and $\alpha$ controls the strength of the prior regularization. For efficiency, the Hessian-related term is approximated by finite differences:
\begin{equation}
\begin{aligned}
&\mathcal{L}_{\text{SAIL}}(\mathbf{x}_T) = \alpha \|\mathbf{x}_T\|_2^2+ \\
&\left\|
\Delta s_{\theta}
\bigl(
\mathbf{x}_T
+
\delta \Delta s_{\theta}(\mathbf{x}_T)
\bigr)
-
\Delta s_{\theta}(\mathbf{x}_T)
\right\|_2^2,
\end{aligned}
\label{eq:sail_loss}
\end{equation}
where $\delta$ is a small finite-difference step size. The optimized initial noise is
then used for generation.

\section{Limitations}
This work focuses on understanding and mitigating memorization in text-to-image diffusion models from the perspective of cross-attention. Our experiments are conducted on models and memorization benchmarks with publicly available memorized prompts and corresponding training images. As newer diffusion models continue to emerge, systematic studies of their memorization behavior and the construction of matched image-prompt benchmarks remain important research directions. Future work could focus on developing memorization benchmarks for newer diffusion models to facilitate further investigation of memorization mechanisms and mitigation strategies.

\section{Ethics Statement}
This work aims to improve the privacy of text-to-image diffusion models by analyzing and mitigating memorization, which may lead to unintended training data reproduction and raise privacy and copyright concerns. Our method mitigates these risks at inference time without accessing or modifying training data. All experiments use publicly available models, datasets, and established benchmarks for research purposes. By investigating the internal mechanisms of memorization, our work contributes to developing safer and more privacy-aware generative models.